\documentclass[final]{cvpr}

\usepackage{times}
\usepackage{epsfig}
\usepackage{graphicx}
\usepackage{amsmath}
\usepackage{amssymb}
\usepackage{tabularx}
\usepackage{verbatim}
\usepackage{float}

\usepackage[pagebackref=true,breaklinks=true,colorlinks,bookmarks=false]{hyperref}

\def\cvprPaperID{4686} \def\confYear{CVPR 2021}

\begin{document}

\title{Explaining Classifiers using Adversarial Perturbations on the Perceptual Ball}

\author{
Andrew Elliott\thanks{Equal contribution}\\
Uni. of Glasgow/Turing Institute\\
{\tt\small andrew.elliott@glasgow.ac.uk}\\
\and 
Stephen Law$^{*}$\\
UCL/Turing Institute \\
{\tt\small stephen.law@ucl.ac.uk}\\
\and
Chris Russell\thanks{Much of this work was done prior to joining Amazon.}\\
Amazon T\"ubingen\\
{\tt\small cmruss@amazon.com}
}

\maketitle

\begin{abstract}
We present a simple regularization of adversarial perturbations based upon the
perceptual loss.
While the resulting perturbations remain imperceptible to the human eye,
they differ from existing adversarial perturbations in that they are
semi-sparse alterations that highlight objects and regions of interest while
leaving the background unaltered. 
As a semantically meaningful adverse perturbations, it forms a bridge between
counterfactual explanations and adversarial perturbations in the space of
images.

We evaluate our approach on several
standard explainability benchmarks, namely, weak localization, insertion-deletion, and the pointing game demonstrating that perceptually regularized
counterfactuals are an effective explanation for image-based
classifiers.

\end{abstract}

\section{Introduction}
\label{sec:introduction}
\begin{figure}
\centering
 \includegraphics[width=2.5cm]{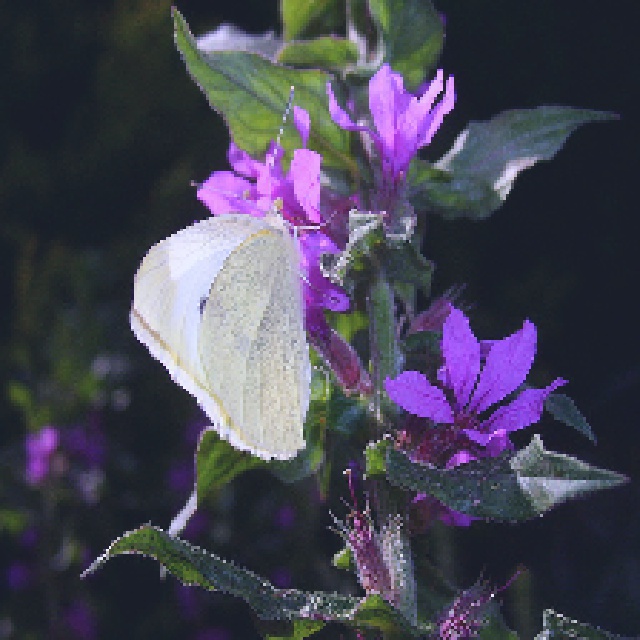}
\includegraphics[width=2.5cm]{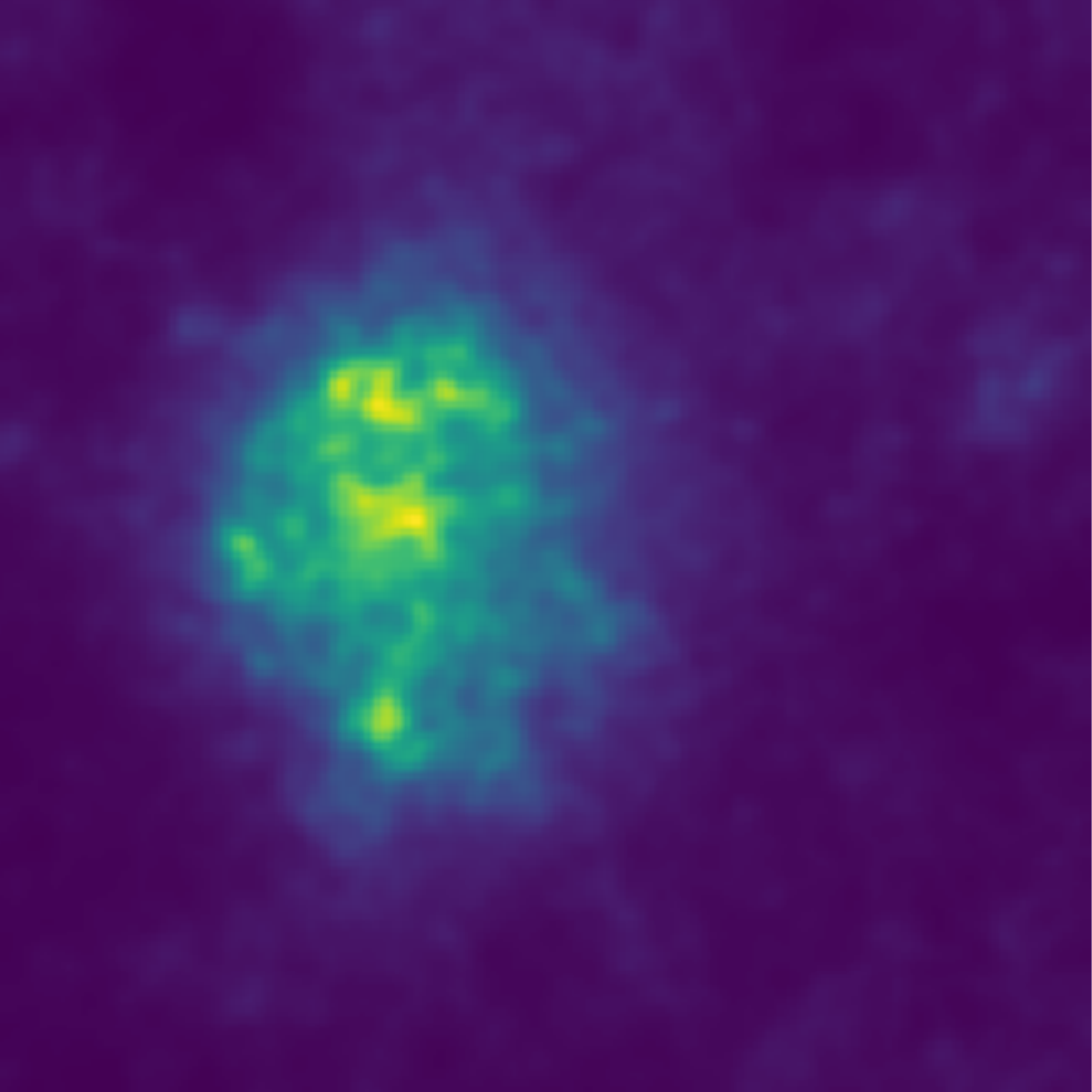}
 \includegraphics[width=2.5cm]{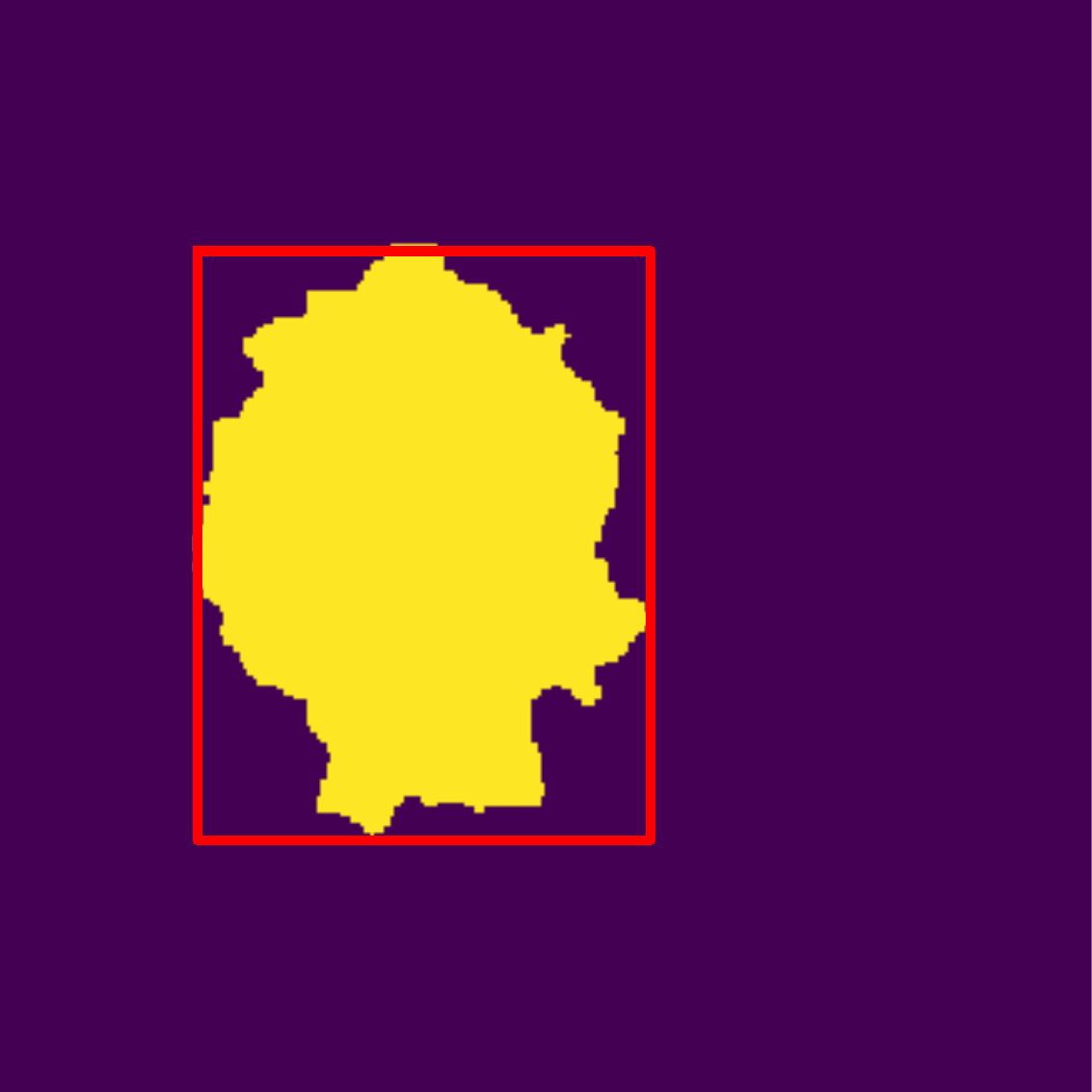}
\caption{Object localization. From left to right: Original image;  Magnitude of the perceptual perturbations; Dominant connected component and the  bounding box from automatic object detection.  Despite the flowers our method highlights the  butterfly as salient. 
\label{fig:teaser}}
 \end{figure}

\begin{figure*}[!t]
\centering
\begin{tabular}{c 
@{\hspace{1mm}}c 
@{\hspace{1mm}}c 
@{\hspace{1mm}}c 
@{\hspace{1mm}}c
@{\hspace{1mm}}c}
Source& noPerceptual& Layers 0-2 & Layers 0-4&Layers 0-6&Layers0-12 \\
\includegraphics[width=0.16\linewidth]{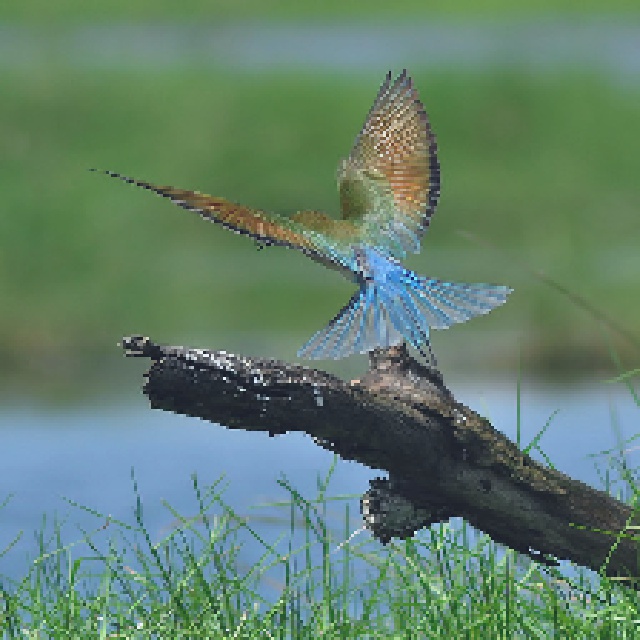}&
\includegraphics[width=0.16\linewidth]{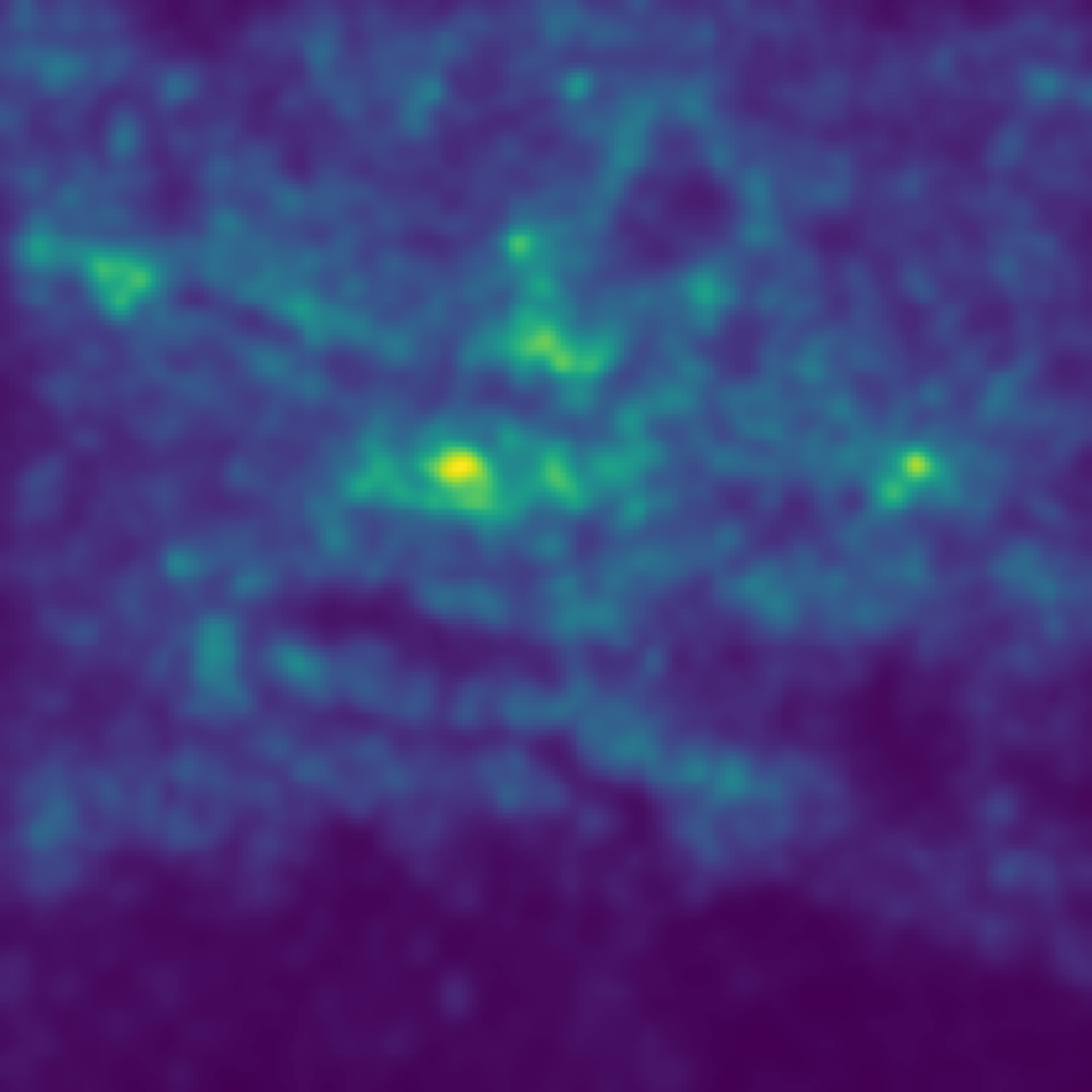} &
\includegraphics[width=0.16\linewidth]{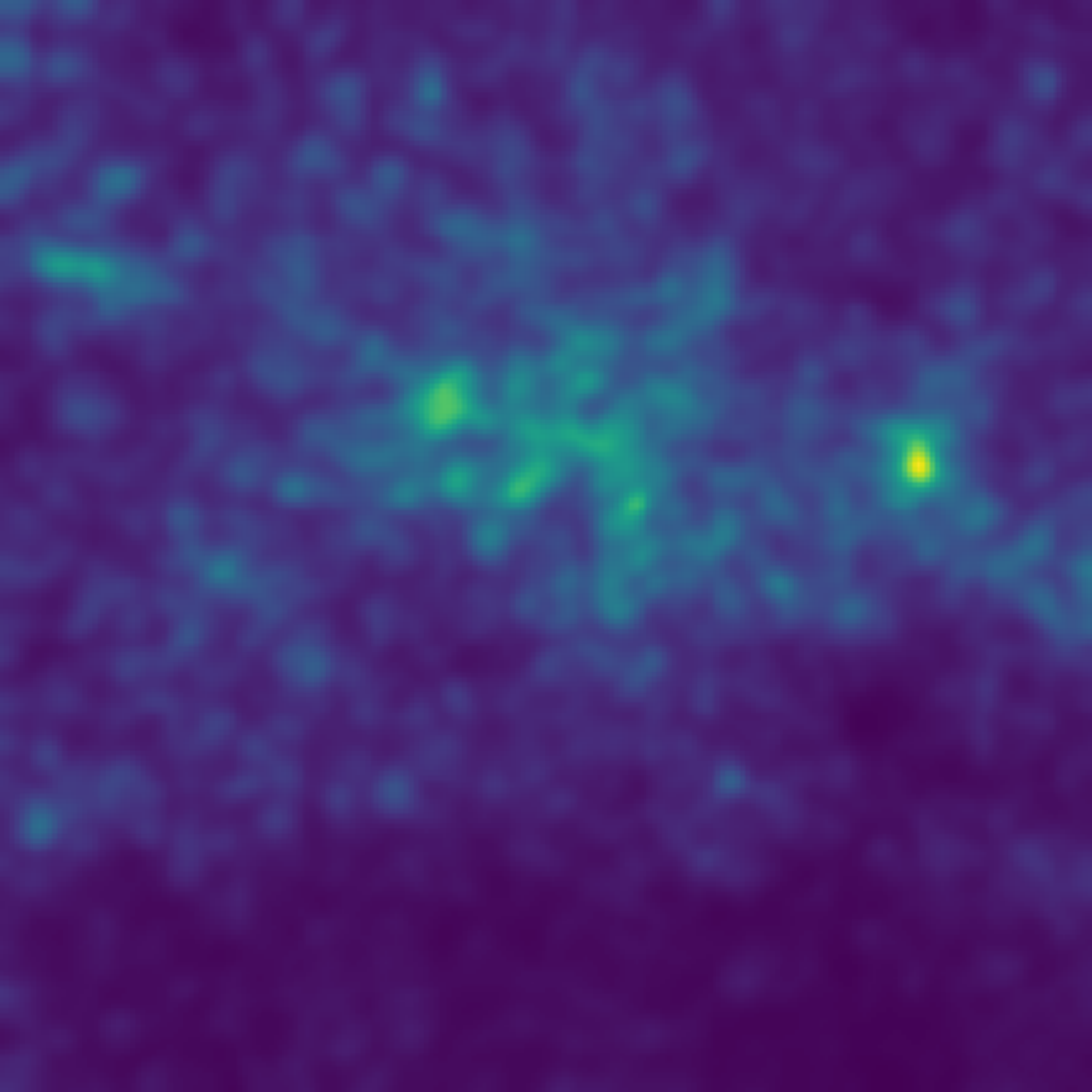} &
\includegraphics[width=0.16\linewidth]{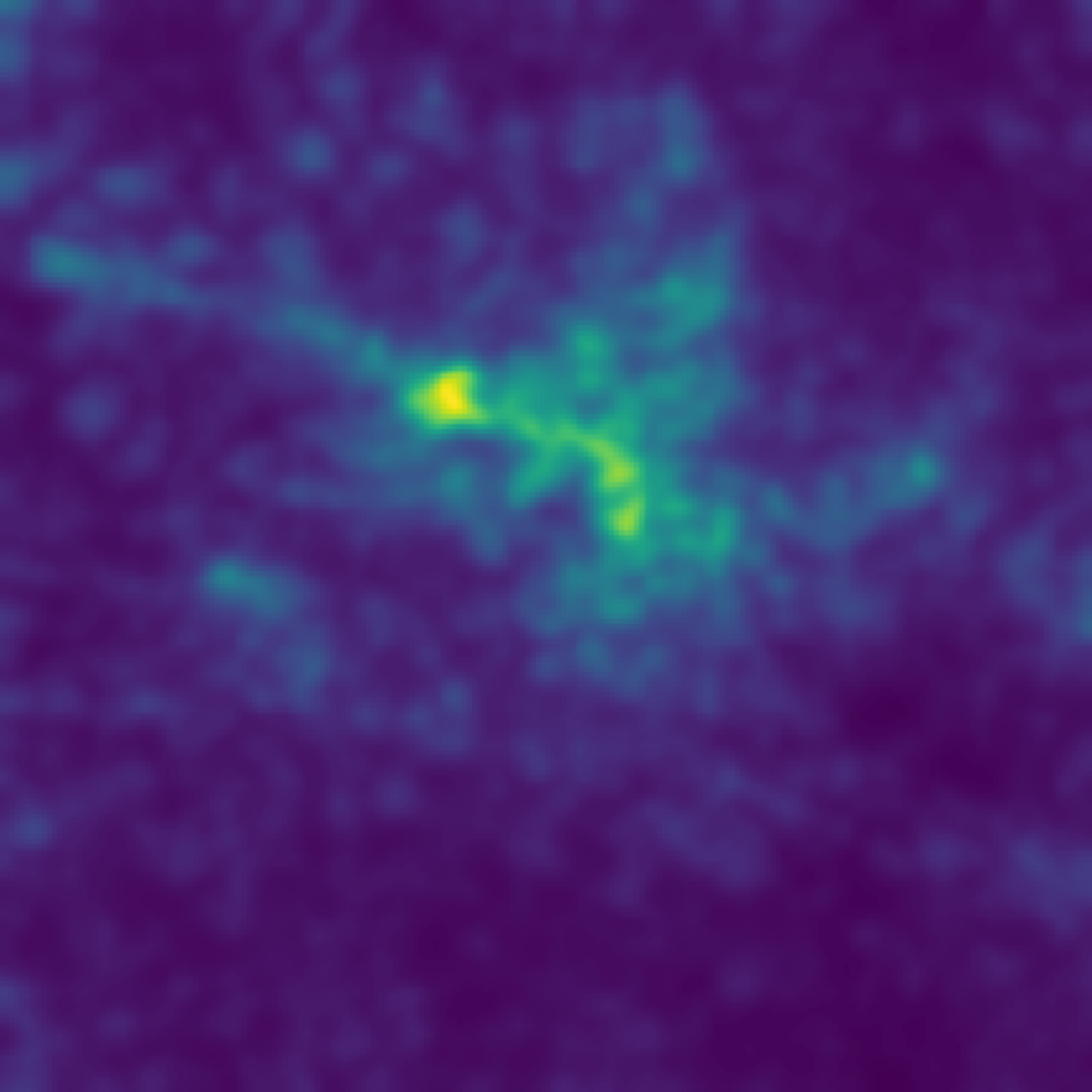} &
\includegraphics[width=0.16\linewidth]{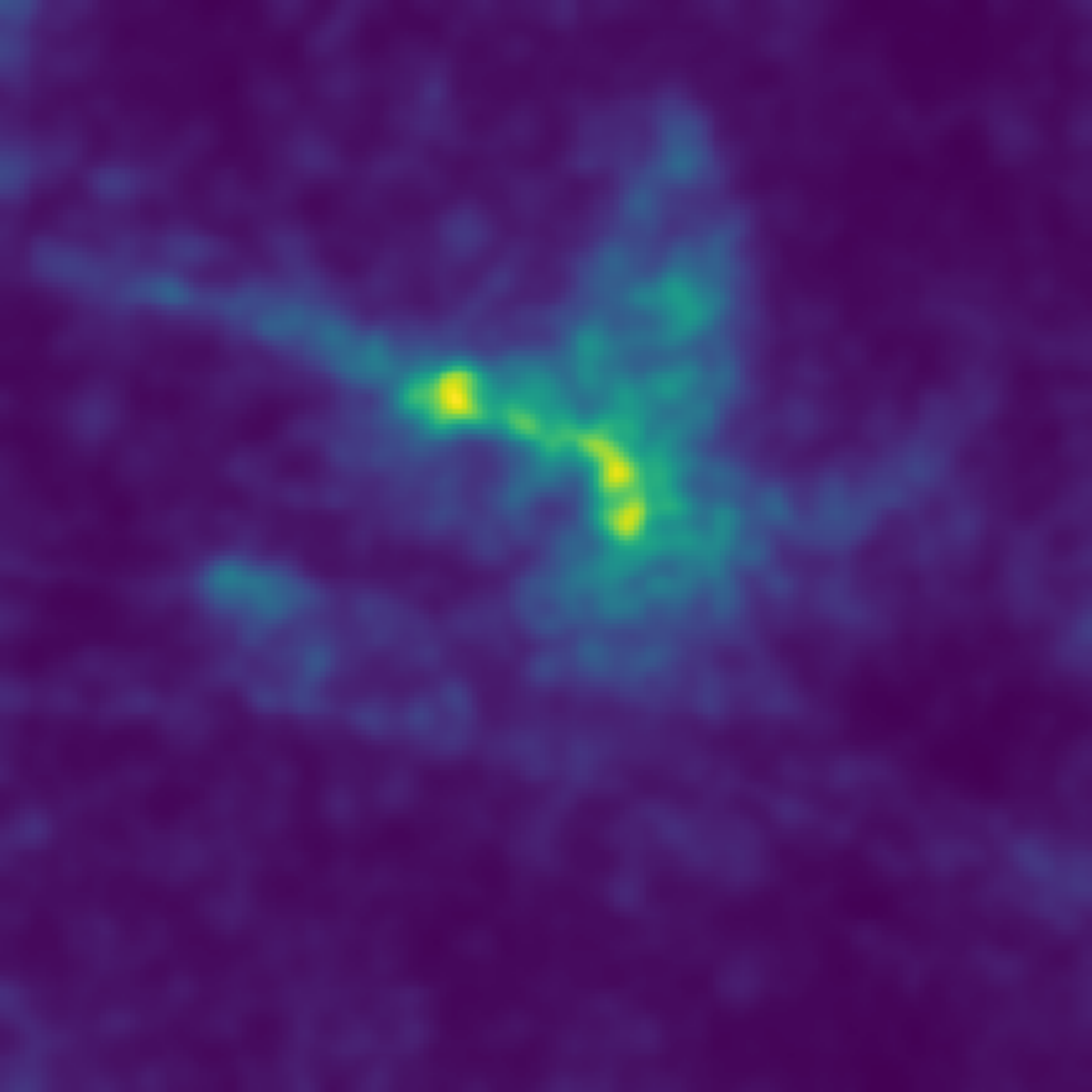} &
\includegraphics[width=0.16\linewidth]{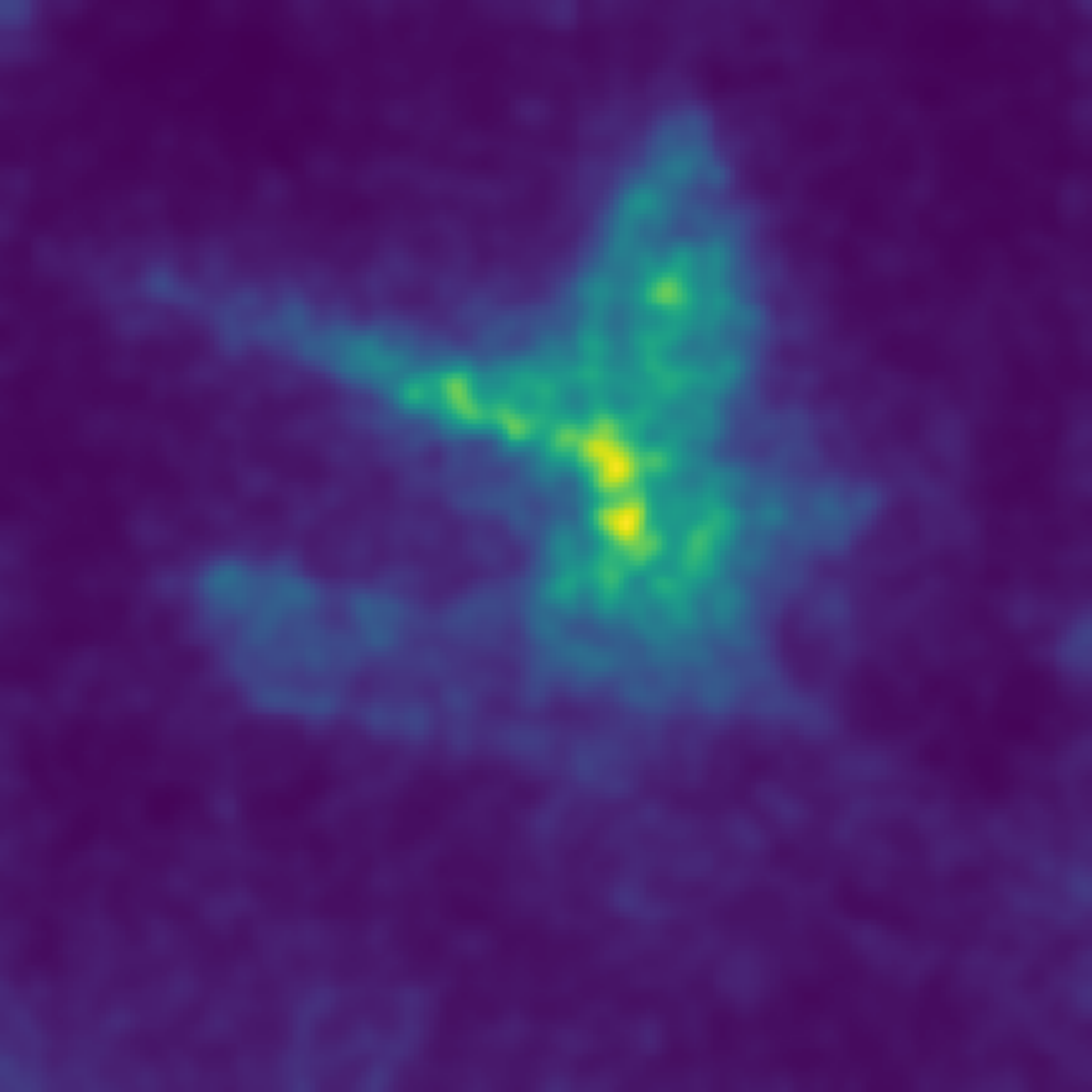} \end{tabular}
\caption{A representative image showing the effects of regularization over different VGG19bn layers ($\sigma=2.0$). As we extend the regularization to cover higher layers we find the perturbation becomes more compact and better localized upon the object.
\label{diff_layers}
}
\end{figure*}

We address the gap between counterfactual explanations
\cite{wachter2017counterfactual} and adversarial perturbations
\cite{szegedy2013intriguing}, and show why  minimal changes in
image data that results in a change in classifier response does not result in semantically meaningful alteration.
One might  hope that the smallest
edit to alter classifier response of an image labeled as bird should 
alter the
bird pixels, but in practice adversarial perturbations make
non-local changes that break the classifier. We show how penalizing changes in
the mid-level classifier response with a perceptual loss stops this breakage
and instead results in semantically meaningful changes that highlight the extent
of objects in images (see Figs.~\ref{fig:teaser},\ref{diff_layers}).

Outside of computer vision~\cite{wachter2017counterfactual}, counterfactual 
explanations are a popular method in explainable AI. They find
the smallest change needed to alter the decision 
of a classifier, and on 
tabular data, can give explanations such  as:

{\it 
    ``The loan was denied as your income was £30,000. If
it
had been £45,000, you would have been offered a loan.''
}
The close relationship between adversarial perturbations and counterfactual explanations follows from 
the definitions in philosophy and folk psychology of a counterfactual
explanation as answering the question ``What is a minimal change that would result in a different outcome?'' When applied to classifiers, these counterfactual explanations are simply adversarial perturbations, by a different name. As such, it is interesting to ask, why adversarial perturbations don't work as explanations: Why are they imperceptible? and: Why don't they localize on objects?

Two compelling arguments for the existence of imperceptible adversarial perturbations in
images have been offered. The first due to
\cite{goodfellow2014explaining} remarks that 
they are simply an artifact of a high-dimensional space, and thus, it is entirely expected that a small perturbation of
every pixel 
can add up to a large change in the classifier response, and that in fact the same behavior is found in linear classifiers.

A second argument attempts to understand why sparse 
(potentially 
single
pixel) attacks exist and
attributes the effectiveness of adversarial perturbations to exploding gradients. `Exploding gradients' refers to the 
phenomenon where  changes in functional response grow
exponentially with the depth of the network, relative to a change of input of fixed magnitude. These exploding gradients are an issue known to afflict the learning of Recurrent Neural
Networks \cite{pascanu2012understanding}, and the deep networks common to
computer vision.
This phenomenon occurs because, by construction, neural networks form a product
of 
(convolutional) 
matrix operations interlaced with nonlinearities; and for
directions/locations in which these nonlinearities act approximately linearly,
the eigenvalues of the Jacobian can grow exponentially with depth\footnote{
See
\cite{pascanu2012understanding} for a formal derivation.
}.
 While this  phenomenon
is well-studied in the context of training networks 
with remedies such as
normalization \cite{ioffe2015batch} and 
gradient
clipping \cite{pascanu2012understanding},
the same phenomenon occurs when
generating adversarial perturbations. 
As such,
a carefully chosen small
perturbation to have an extremely large effect on the response of 
a deep or recurrent classifier.

To explore how these arguments fit together, and which explanation
accounts for the familiar behavior of adversarial perturbations, we propose a
simple novel regularization that bounds the exponential growth of the classifier
response by regularizing the perceptual distance~\cite{johnson2016perceptual}
between the image and its adversarial perturbation.

One common criticism of adversarial perturbation is that the generated images
lie outside the manifold of natural images, and 
if we could
sample
from the manifold, 
our adversarial perturbations would be both larger
and more representative of the real world. Restricting adversarial perturbations
to this manifold 
should limit the impact of exploding gradients
-- if samples are drawn from this space then a well-trained classifier should
implicitly reflect the smoothness of the true labels of the underlying data
distribution.

While it is believed that the manifold of natural images is low dimensional~\cite{leeNonLinear2003}, characterizing 
this manifold outside of handwritten
digits has proven extremely challenging\footnote{See discussion in the
  experimental section of~\cite{stutz2019disentangling}.}. Our approach provides
a complementary lightweight alternative. Rather than attempting to characterize the manifold,
we penalize search directions that exploit exploding gradients as these encourage movement off the data manifold when searching for minimal adversarial perturbations.

We propose a novel regularization for adversarial perturbations based around the
perceptual loss. Our new perturbations tend to
highlight objects and regions of interest within the image (see Fig.~\ref{fig:teaser})\footnote{
 Our implementation can be found at \url{www.github.com/alan-turing-institute/perceptualBall}.
 }.
 We evaluate on several standard explainability
challenges for image classifiers and further validate using the sanity checks of~\cite{adebayo2018sanity}.

\begin{figure*}
\begingroup
\setlength{\tabcolsep}{0pt} \renewcommand{\arraystretch}{0.0} \centering
\begin{tabular}{@{}c@{}c@{}c@{}c@{}c@{}c@{}c@{}c@{}c@{}}
Orig 
& 
Grad-CAM
& 
SmoothGrad
& 
Excitation
& 
IntGrad
& 
G-Backprop
& 
Extremal
& 
RISE
& 
Us

\vspace{-0.05cm}
\\
 \includegraphics[width=1.9500cm]{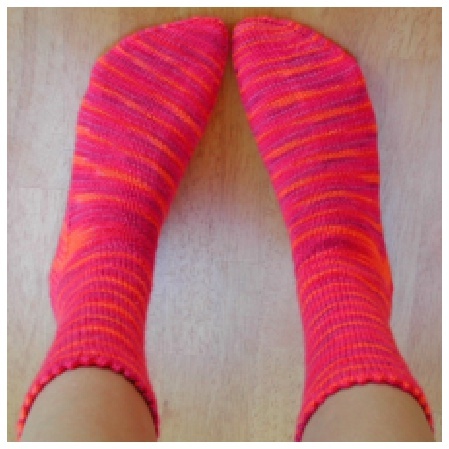}
& 
 \includegraphics[width=1.9500cm]{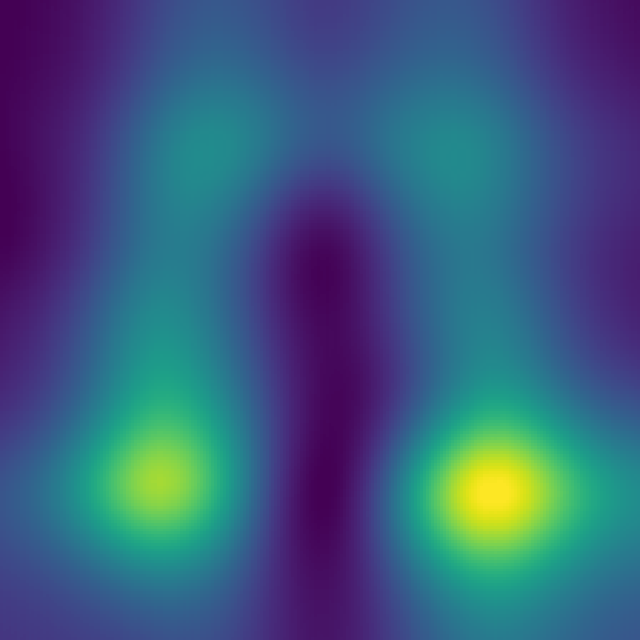}
& 
 \includegraphics[width=1.9500cm]{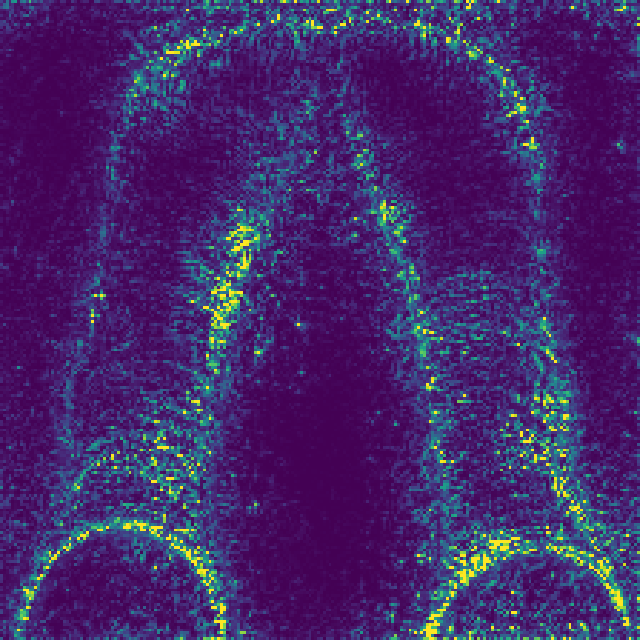}
& 
 \includegraphics[width=1.9500cm]{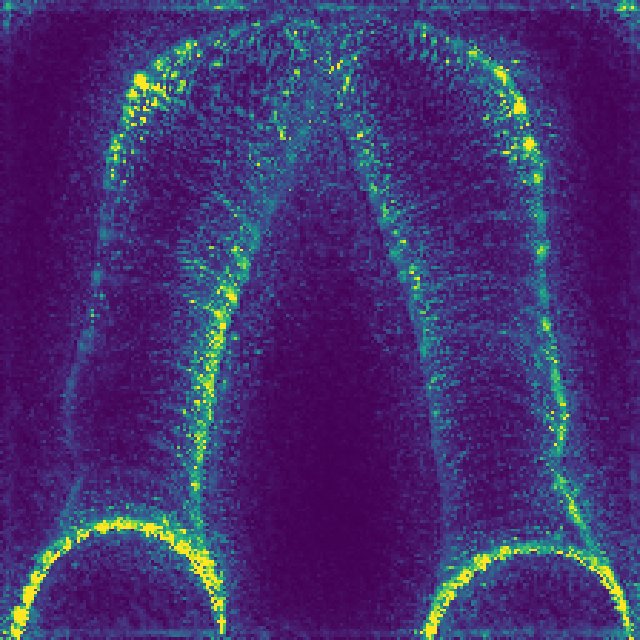}
& 
 \includegraphics[width=1.9500cm]{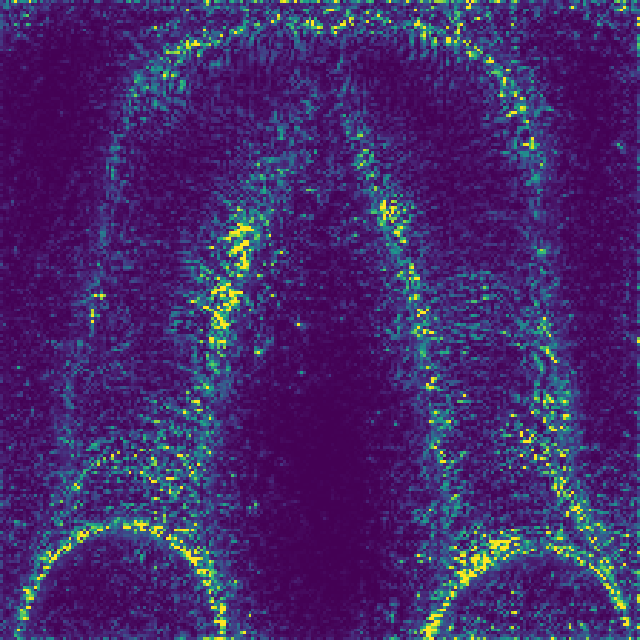}
& 
 \includegraphics[width=1.9500cm]{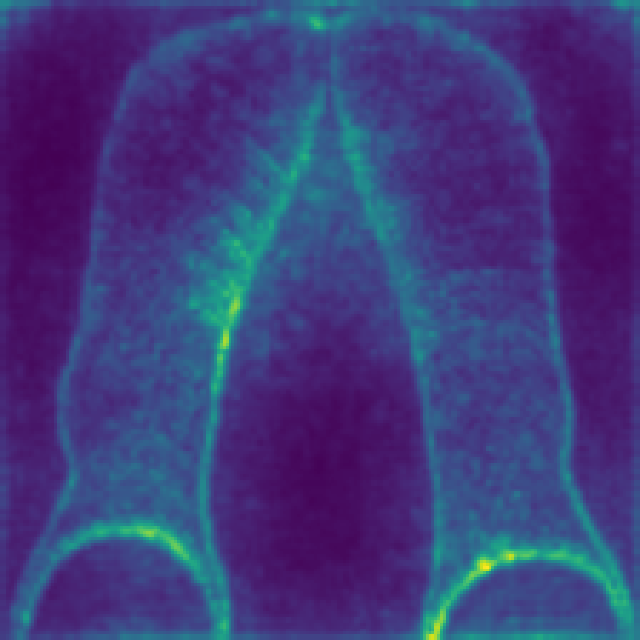}
& 
\includegraphics[width=1.9500cm]{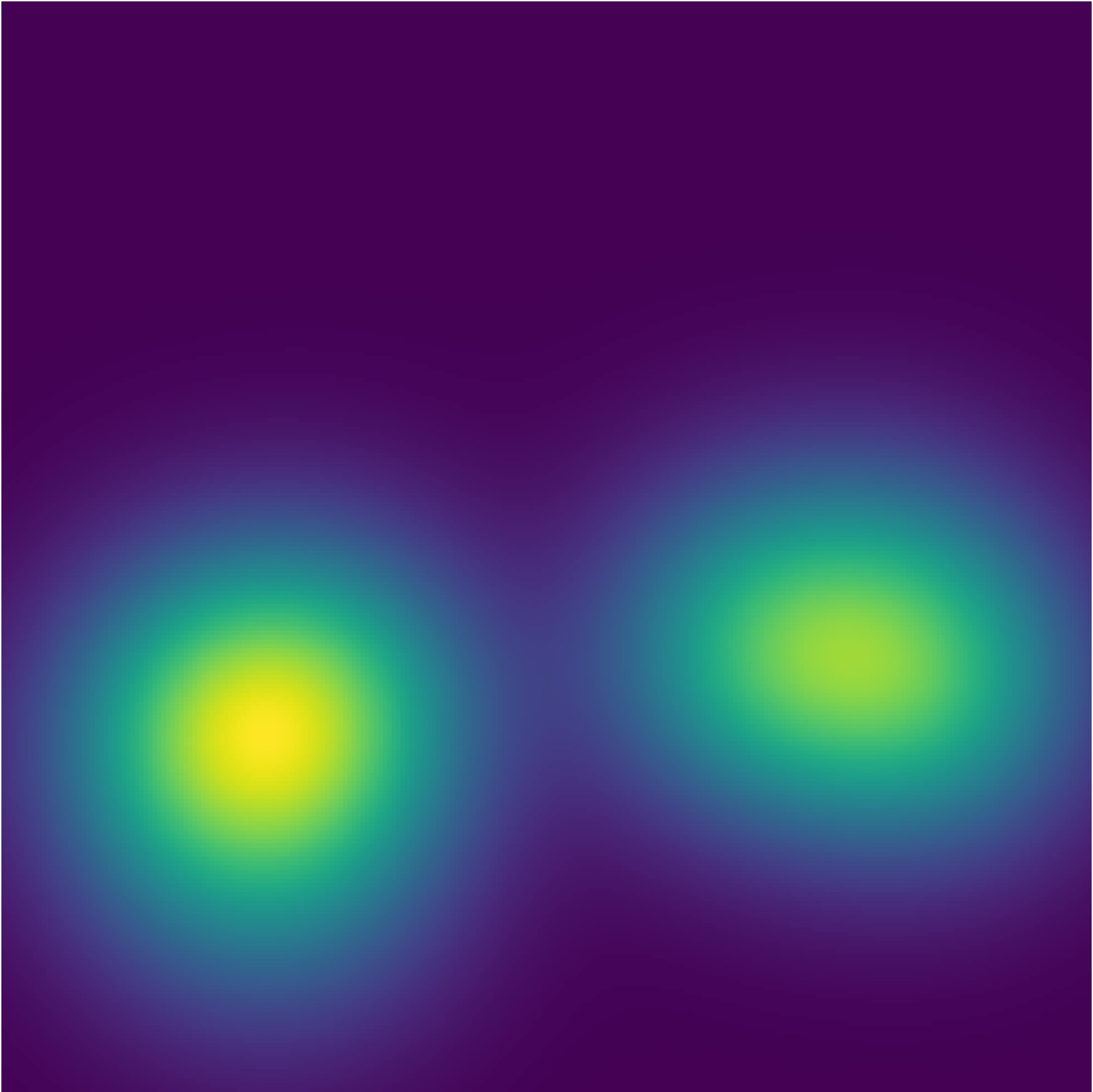}
& 
  \includegraphics[width=1.9500cm]{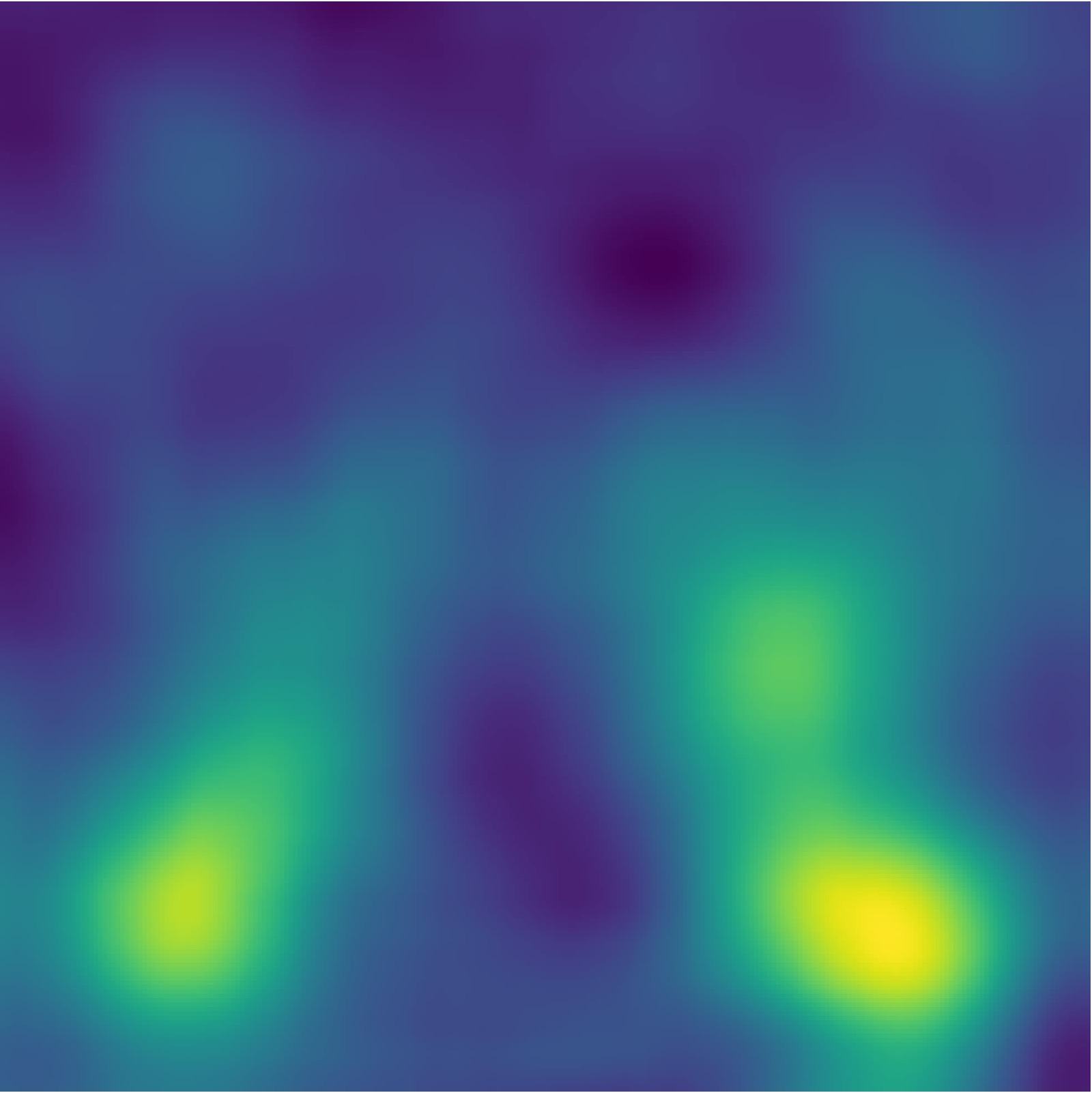}
& 
 \includegraphics[width=1.9500cm]{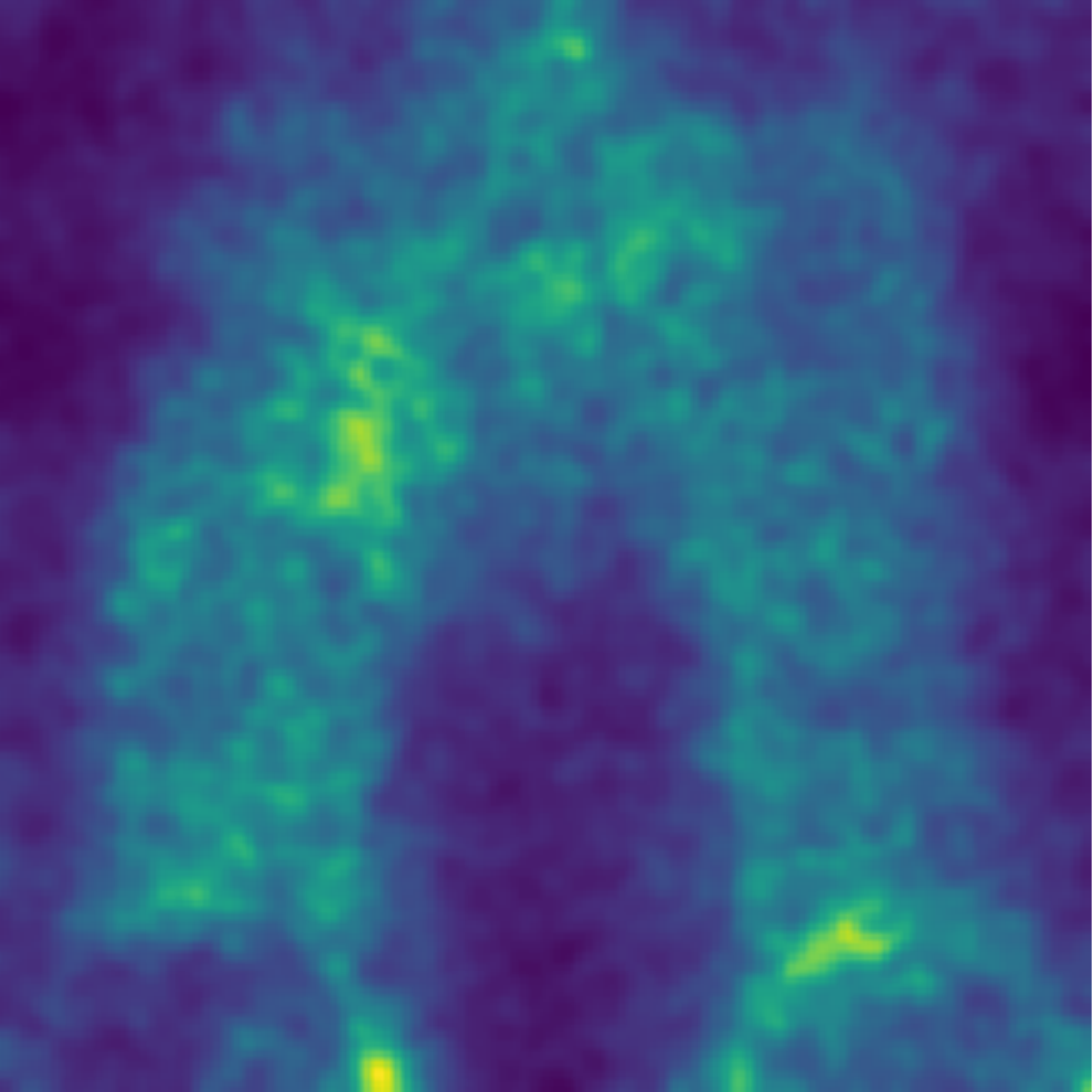}
\\ \includegraphics[width=1.9500cm]{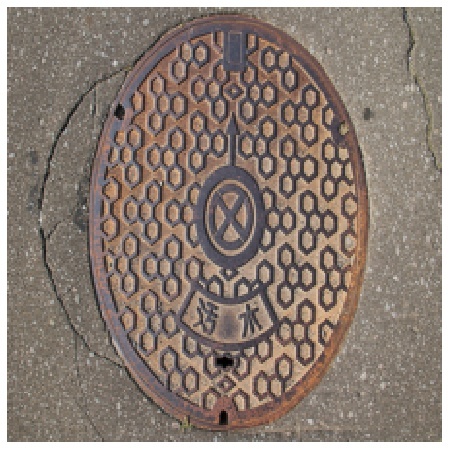}
& 
 \includegraphics[width=1.9500cm]{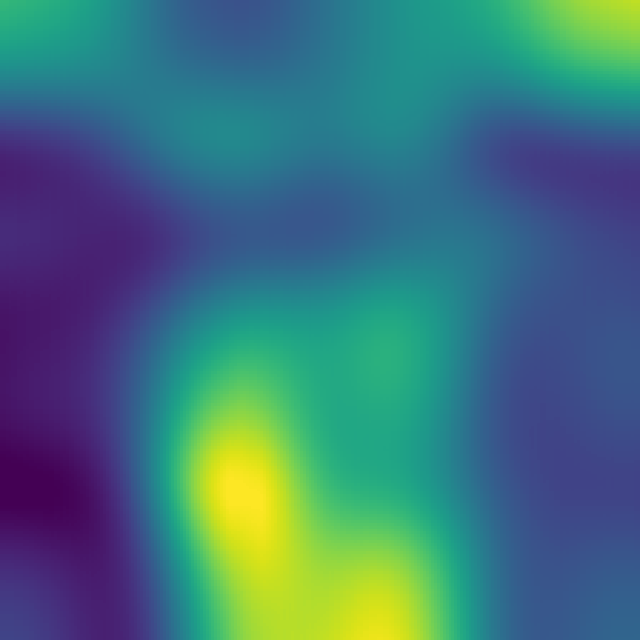}
& 
 \includegraphics[width=1.9500cm]{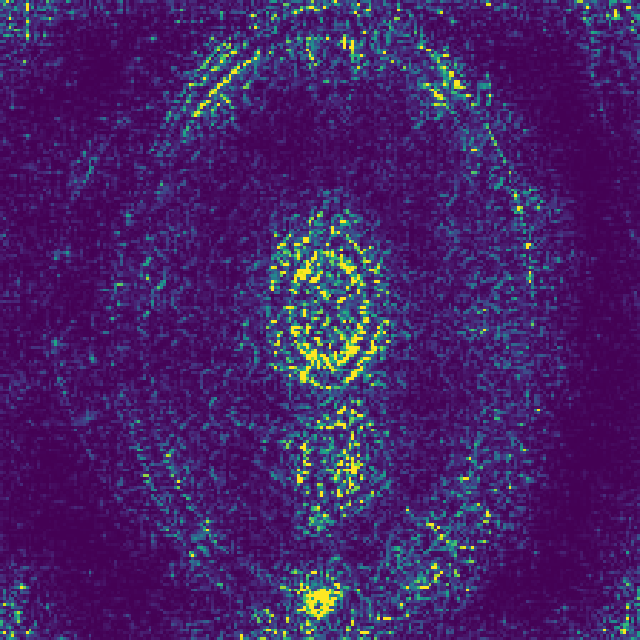}
& 
 \includegraphics[width=1.9500cm]{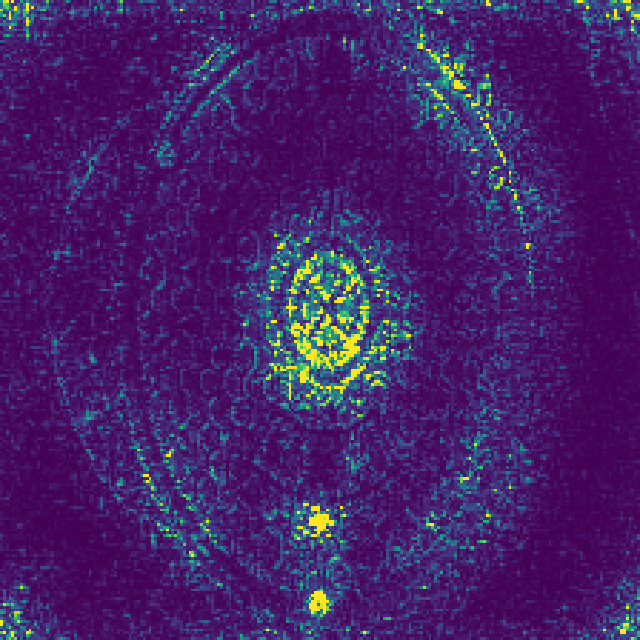}
& 
 \includegraphics[width=1.9500cm]{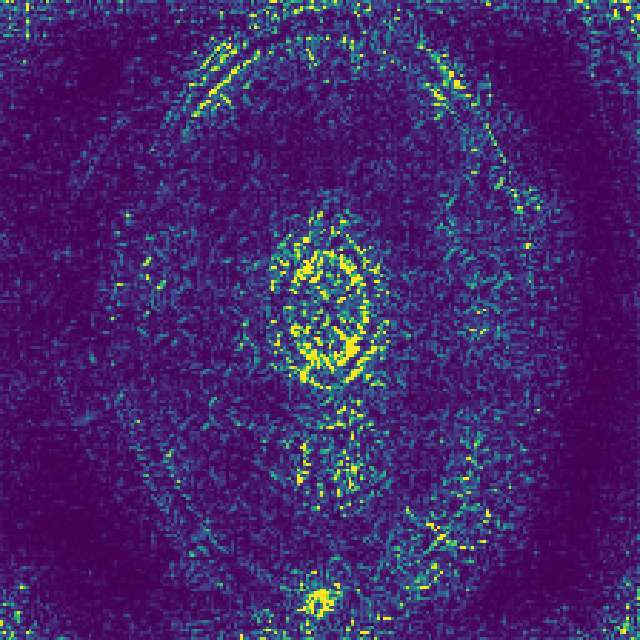}
& 
 \includegraphics[width=1.9500cm]{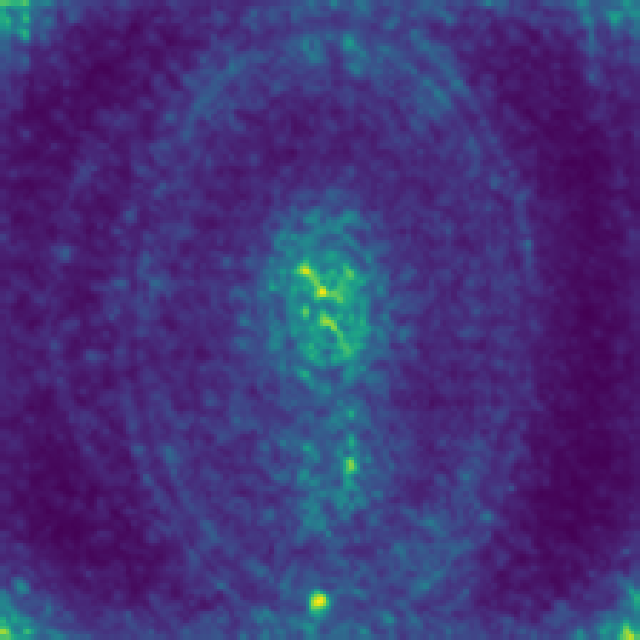}
& 
\includegraphics[width=1.9500cm]{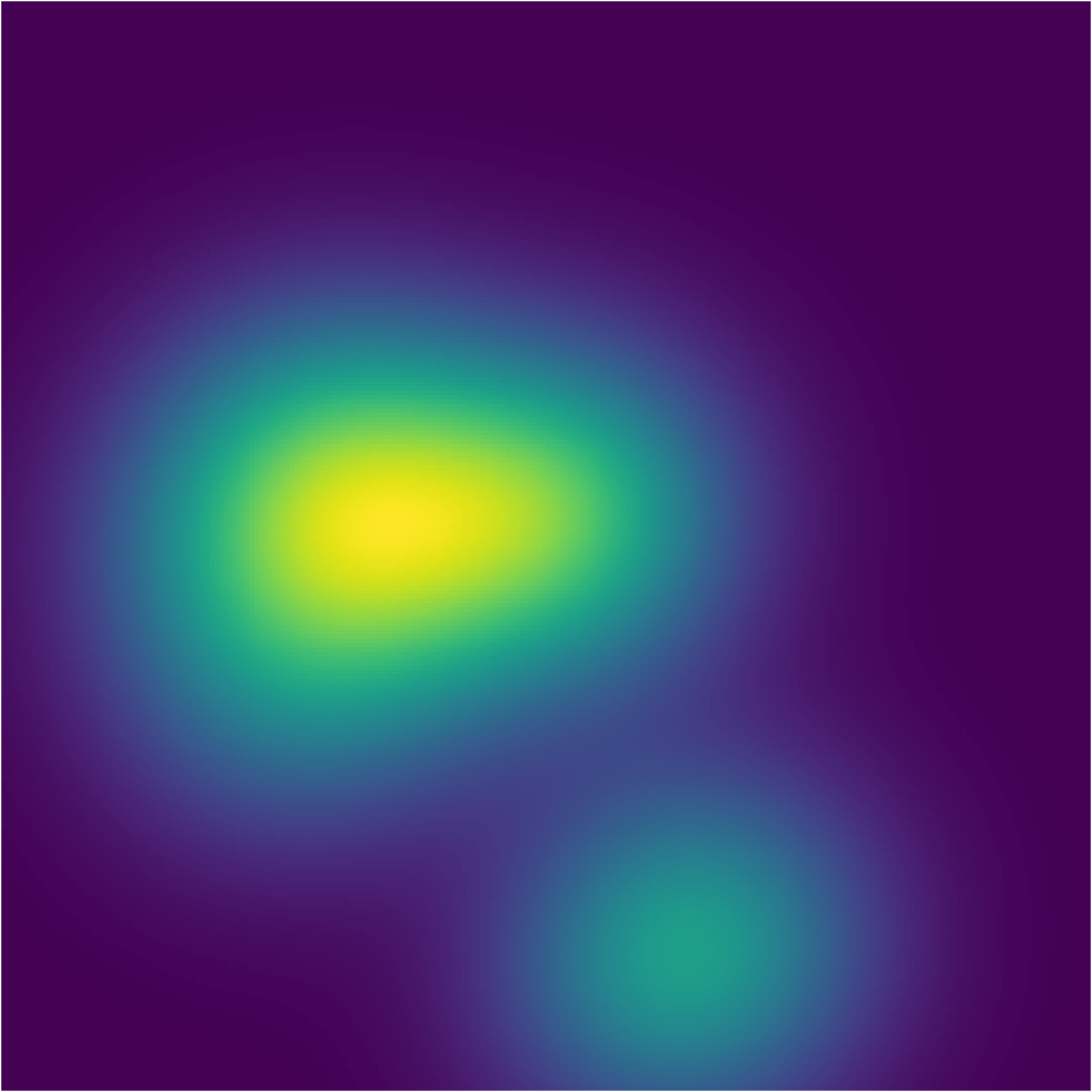}
& 
  \includegraphics[width=1.9500cm]{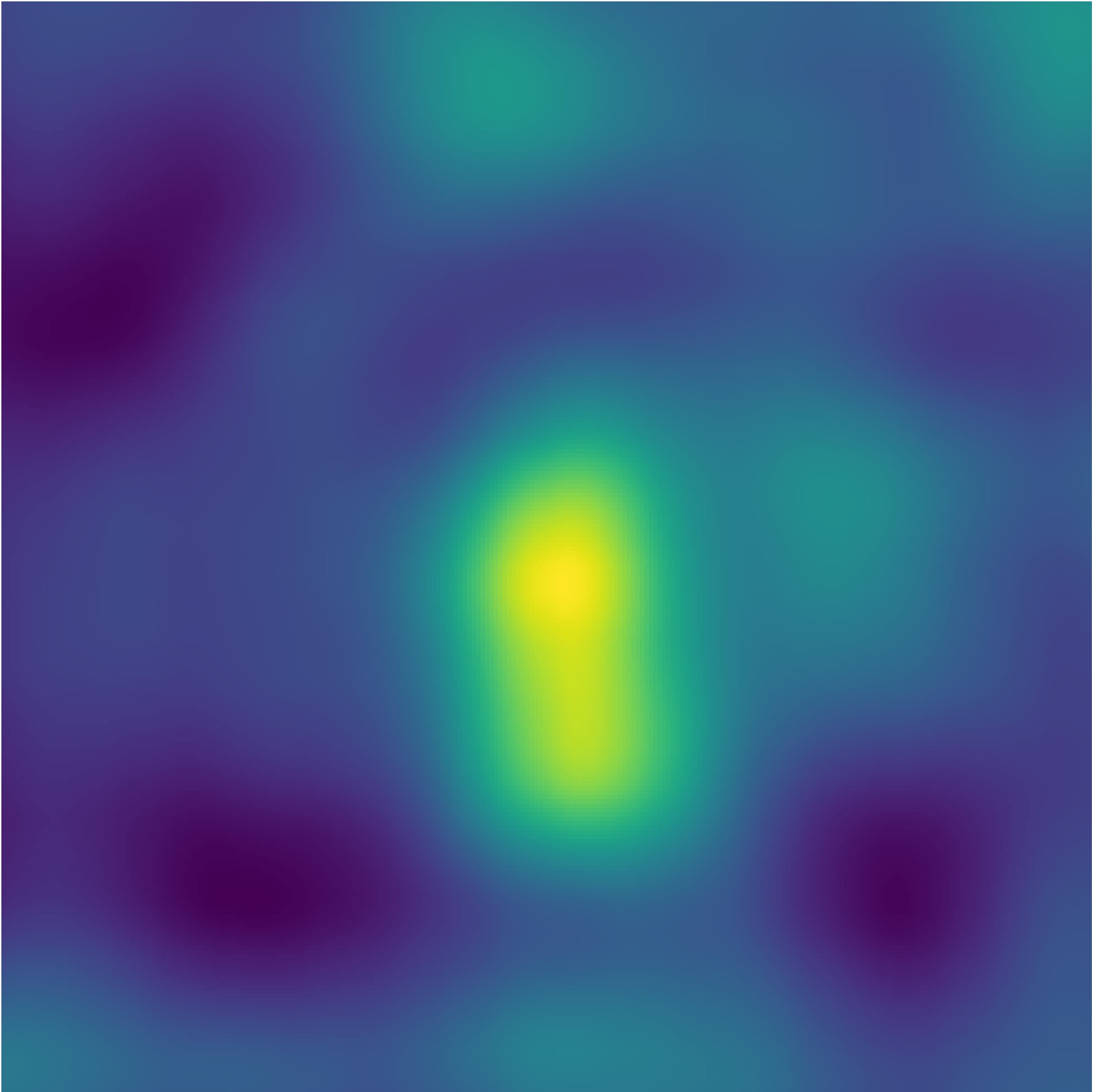}
& 
 \includegraphics[width=1.9500cm]{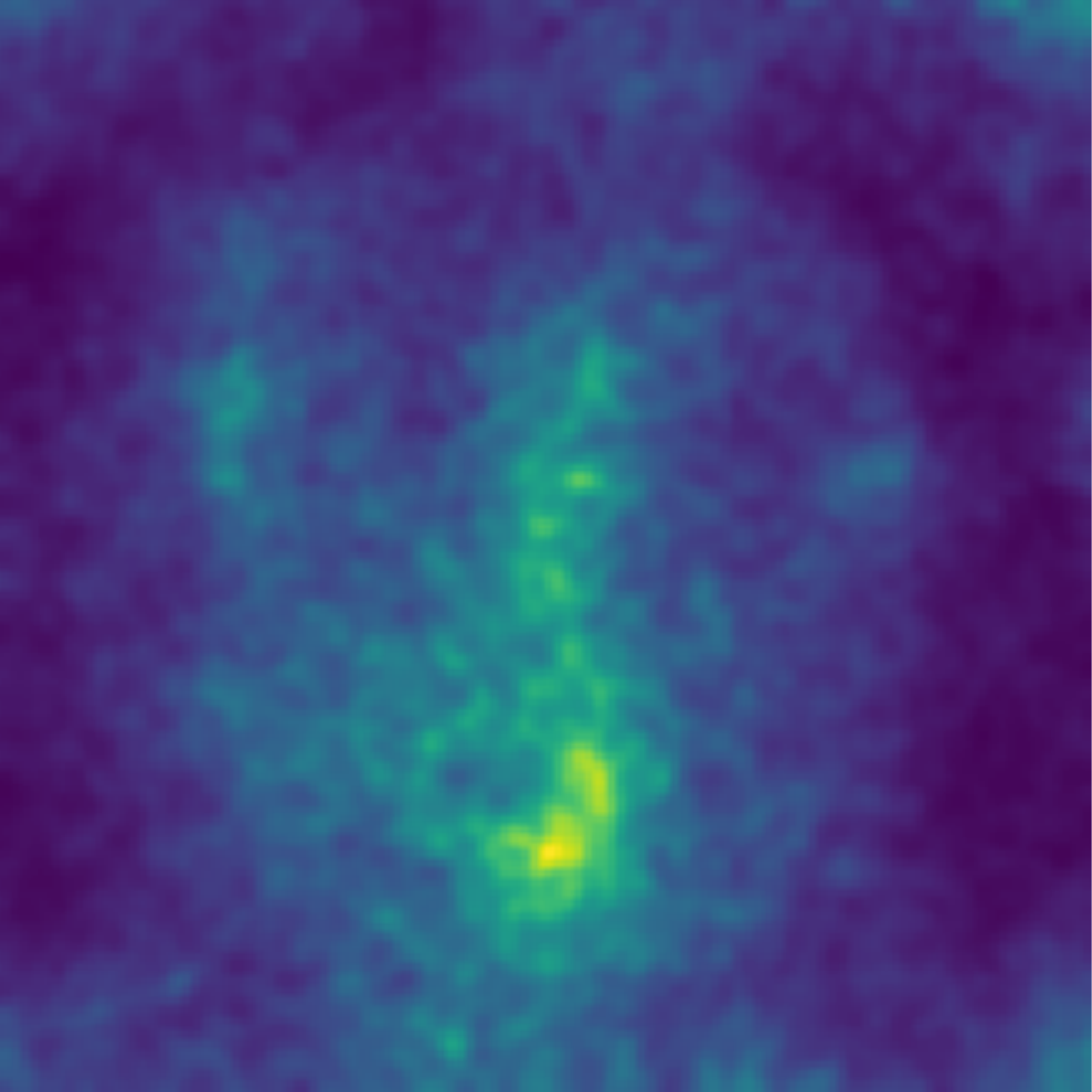}
\end{tabular}
\endgroup
\caption{Comparison of explainability methods on ImageNet validation images.  In contrast to many other approaches that highlight edges, ours highlights the textured 
  interior of the object. This is consistent with the known  
  textual 
  bias of ImageNet trained CNNs \cite{geirhos2019imagenettrained}. 
\label{qualitative}}
\end{figure*}

\section{Prior work}
\label{sec:prior-work}
Numerous approaches to adversarial perturbations have been proposed previously.
These can 
loosely 
be divided into
white-box~\cite{carlini2017towards,chen2018ead,modas2019sparsefool,Moosavi_Dezfooli_2016}
approaches that assume access to the underlying nature of the model and
black-box methods which do not~\cite{liu2016delving,papernot2016distillation}.
The search for an adversarial perturbation is often formulated as trying to find
the closest point to a particular image, under the $\ell_\infty$, $\ell_1$ or
$\ell_2$ norm that takes a different class label.
Numerous defenses have been proposed~\cite{hiddenSpace,roth2019odds} but they can often be circumvented~\cite{carlini2020}.

Other works that add additional constraints to
the perturbation to try to make the generated images more plausible. Such works
restrict the space of perturbations considered by trying to find an
adversarial perturbation that confounds many classifiers at
once~\cite{elsayed2018adversarial}, or is robust to image
warps~\cite{athalye2017synthesizing}. Other approaches considered only a single image and single classifier, but
restricted adversarial perturbations to lie on the manifold of plausible
images~\cite{gilmer2018adversarial,simon2018adversarial,song2017pixeldefend,stutz2019disentangling}.
The principal limitation of these approaches is that they
require a plausible generator of natural images, something that is achievable
with small simple datasets such as MNIST but currently out of reach for even the
224 by 224 thumbnails used by typical ImageNet~\cite{ILSVRC15} classifiers.

\paragraph{Adversarial Perturbations and Counterfactuals}
There are substantial works ~\cite{verma2020counterfactual,wachter2017counterfactual} 
relating adversarial perturbations and counterfactual explanations. This 
relationship follows from the definitions in philosophy and folk psychology 
of a counterfactual explanation as answering the question ``What could have 
 been different in order for outcome A to have occurred instead of B?''. 
With full causal models of images being outside our grasp, such questions 
are commonly answered using Lewis's Closest Possible World semantics~\cite{lewis2013counterfactuals}, rather than Pearl's Structured 
Causal Models~\cite{pearl2000causality}. Under Lewis's framework, an explanation 
for why an image is classified as `dog' rather than `cat' can be found by 
searching for the most similar possible world (i.e. image) which is assigned 
the label `cat' by the classifier.

Conceptually, this is no different to searching for an adversarial perturbation sampled from the space of
possible images. Several approaches have been proposed that either bypass the
requirement that the counterfactual is an image, and return text descriptions~\cite{hendricks2018generating}, na\"{i}vely
ignoring the requirement that the world is
plausible~\cite{wachter2017counterfactual}, used
prototypes~\cite{van2019interpretable}, or  auto-encoders~\cite{dhurandhar2018explanations}, or Gaussian kernels~\cite{gardner2015deep} to
characterize the manifold of plausible images, or require large edits that
replace regions of the image, either with the output of GANs~\cite{chang2018explaining} or with patches
from other images~\cite{goyal2019counterfactual}.

Recent works \cite{kaur2019perceptuallyaligned,tsipras2018robustness} suggested that robust classifiers naturally encourage perceptually aligned gradients. Our insight can be seen as complementary, rather than requiring that the classifier is robust, we prohibit perturbations from exploiting the fragility arising from exploding gradients.

\paragraph{ Adversarial Perturbations and Gradient Methods}
The majority of methods in the explainability of computer vision tend to be
gradient or importance-based methods that assign an importance weight to every pixel in the image; every superpixel; or to mid-level neurons. These gradient methods and adversarial perturbations are strongly related. In fact, with most modern networks being piecewise linear, if the found adversarial
perturbation and the original image lie on the same linear piece, the difference
between the original image and closest adversarial perturbations under the
$\ell_2$ norm is equivalent to the direction of steepest descent, up to scaling.
As such, $\ell_2$ adversarial perturbations can be thought of as a slightly
robustified method of estimating the gradient, that takes into account some
local non-linearities.

Of the pure gradient-based approaches, \cite{Simonyan2013} calculated the output
gradient with respect to the input image to create a saliency map giving
fine-grained, but potentially less interpretable results. Other gradient
approaches include SmoothGrad \cite{smilkov2017smoothgrad} which stabilizes the
saliency maps by averaging over multiple noisy copies, and Integrated Gradients
\cite{sundararajan2017axiomatic} which accumulates gradients seen when perturbing
an empty image to the input
image.

CAM based approaches~\cite{GradCAM2016,Zhou_2016CAM} sum the activation maps
in the final convolutional layer of the network. These small activation maps are
up-sampled to obtain a heatmap that highlights particularly salient
regions. Grad-CAM is a generalized variant which finds similar regions of interest
to the perturbation based approaches~\cite{GradCAM2016}.
Recently, \cite{rebuffi2020revisiting} introduced a framework that tries to unify these various gradient approach by proposing NormGrad,
which
aggregates the spatial gradient contributions of individual layers.

Perturbation methods estimate the local sensitivity over a larger range than gradient methods. For example,~\cite{Zeiler2014} applied constant occlusion masks to different input patches repeatedly to find sensitive regions. LIME~\cite{Ribiero2016} constructed a linear model using
the responses obtained from perturbing super-pixels. The recent work on
Extremal Perturbation~\cite{fong2019understanding} estimates an optimal
mask of the image to occlude which gives a maximal effect on the network’s
output.

Various experiments have been proposed to test explanations including the pointing game~\cite{petsiuk2018rise,GradCAM2016,Zhang_2017}, the weakly 
supervised object localization task~\cite{Chattopadhay_2018,fong_2017} and 
the insertion and deletion game~\cite{petsiuk2018rise,wagner2019interpretable}. 
In particular,~\cite{adebayo2018sanity} developed experiments to test the 
suitability of saliency methods.  A number 
of existing saliency techniques have been evaluated using these experiments, including: 
NormGrad~\cite{rebuffi2020revisiting},
Extremal Perturbation~\cite{fong2019understanding},
Gradient~\cite{Simonyan2013},
RISE~\cite{petsiuk2018rise},
Grad-CAM~\cite{GradCAM2016},
SmoothGrad~\cite{smilkov2017smoothgrad},
GuidedBackprop~\cite{springenberg2014striving}, 
Integrated Gradients~\cite{sundararajan2017axiomatic},
Deconvolution~\cite{Zeiler2014},
and 
Excitation Backpropagation~\cite{Zhang_2017}. We evaluate our 
approach on all these tests and compare against standard methods.

\begin{figure*}
\centering
\begingroup
\setlength{\tabcolsep}{0pt} \renewcommand{\arraystretch}{0.0} \begin{tabular}{p{0.0985\linewidth}p{0.0985\linewidth}p{0.0985\linewidth}p{0.0985\linewidth}p{0.0985\linewidth}p{0.0985\linewidth}p{0.0985\linewidth}p{0.0985\linewidth}p{0.0985\linewidth}p{0.0985\linewidth}}
   \includegraphics[width=\linewidth]{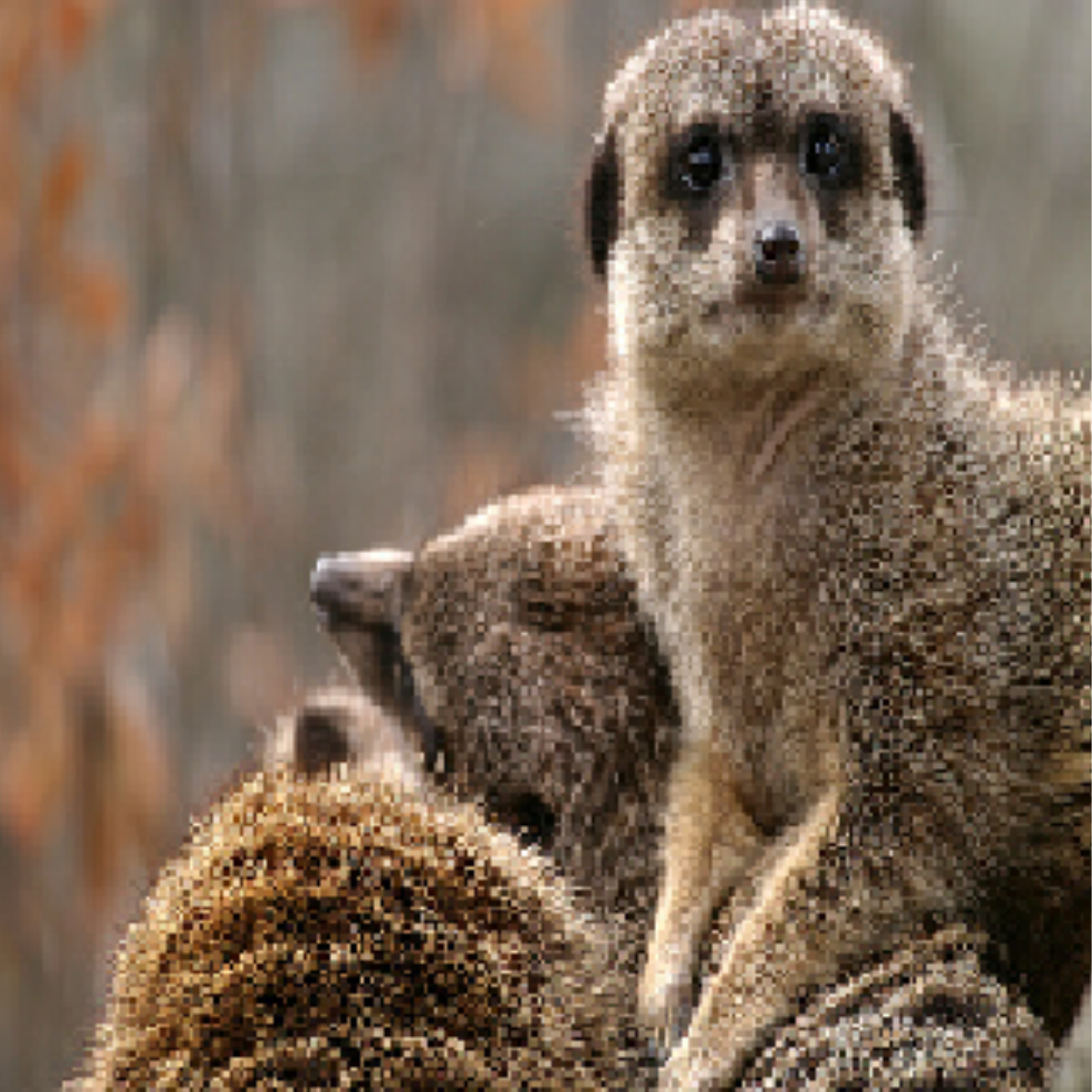}& \includegraphics[width=\linewidth]{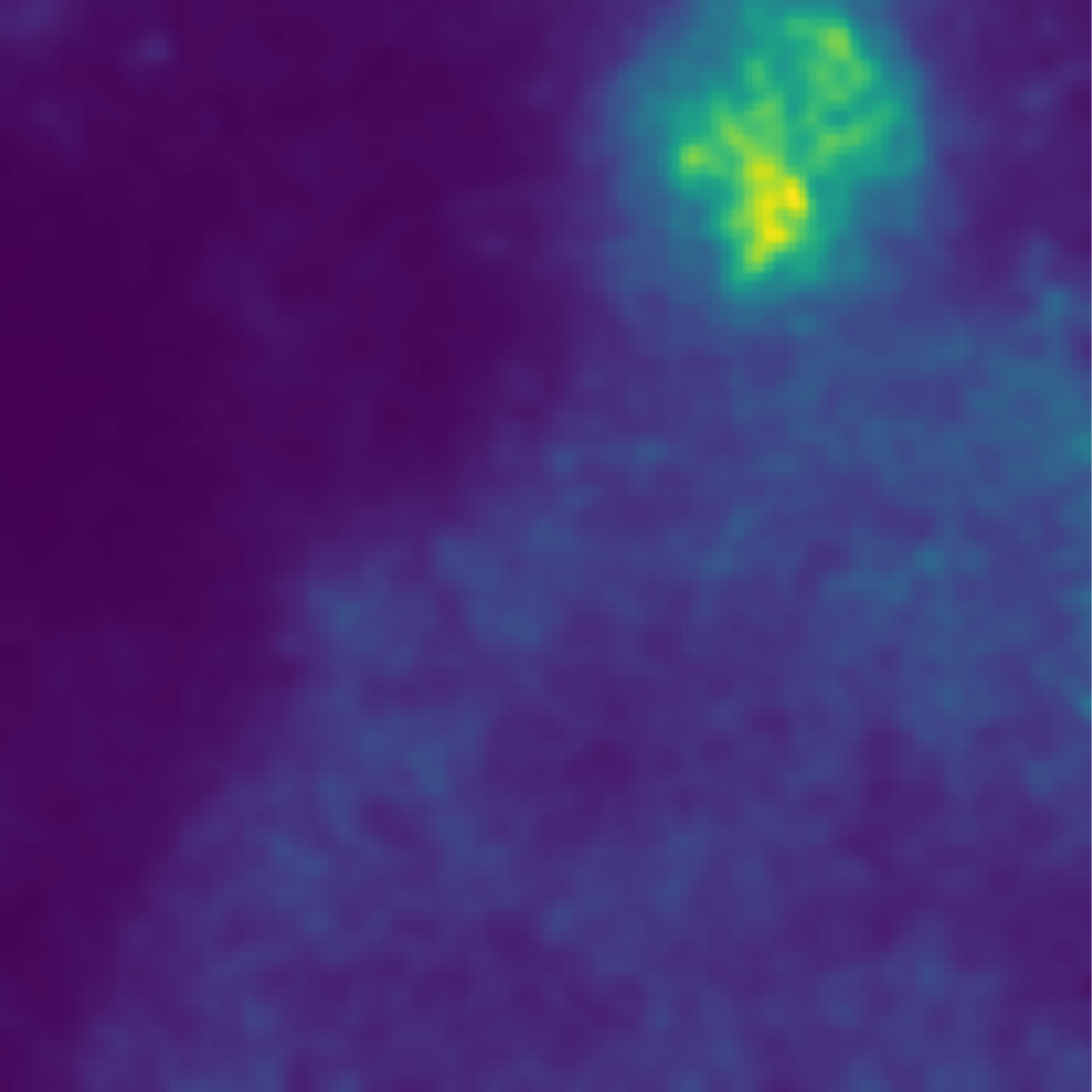}&
\includegraphics[width=\linewidth]{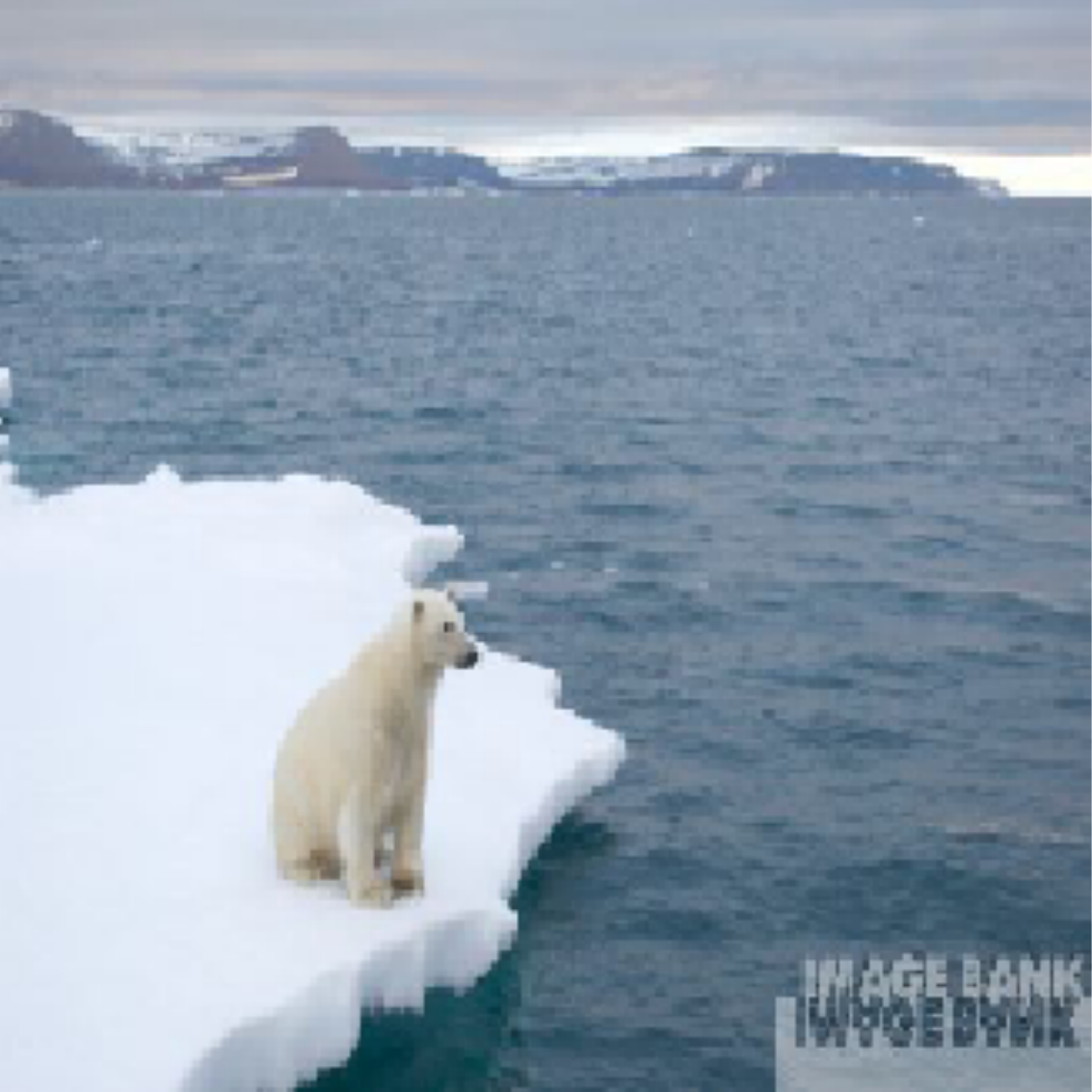}& \includegraphics[width=\linewidth]{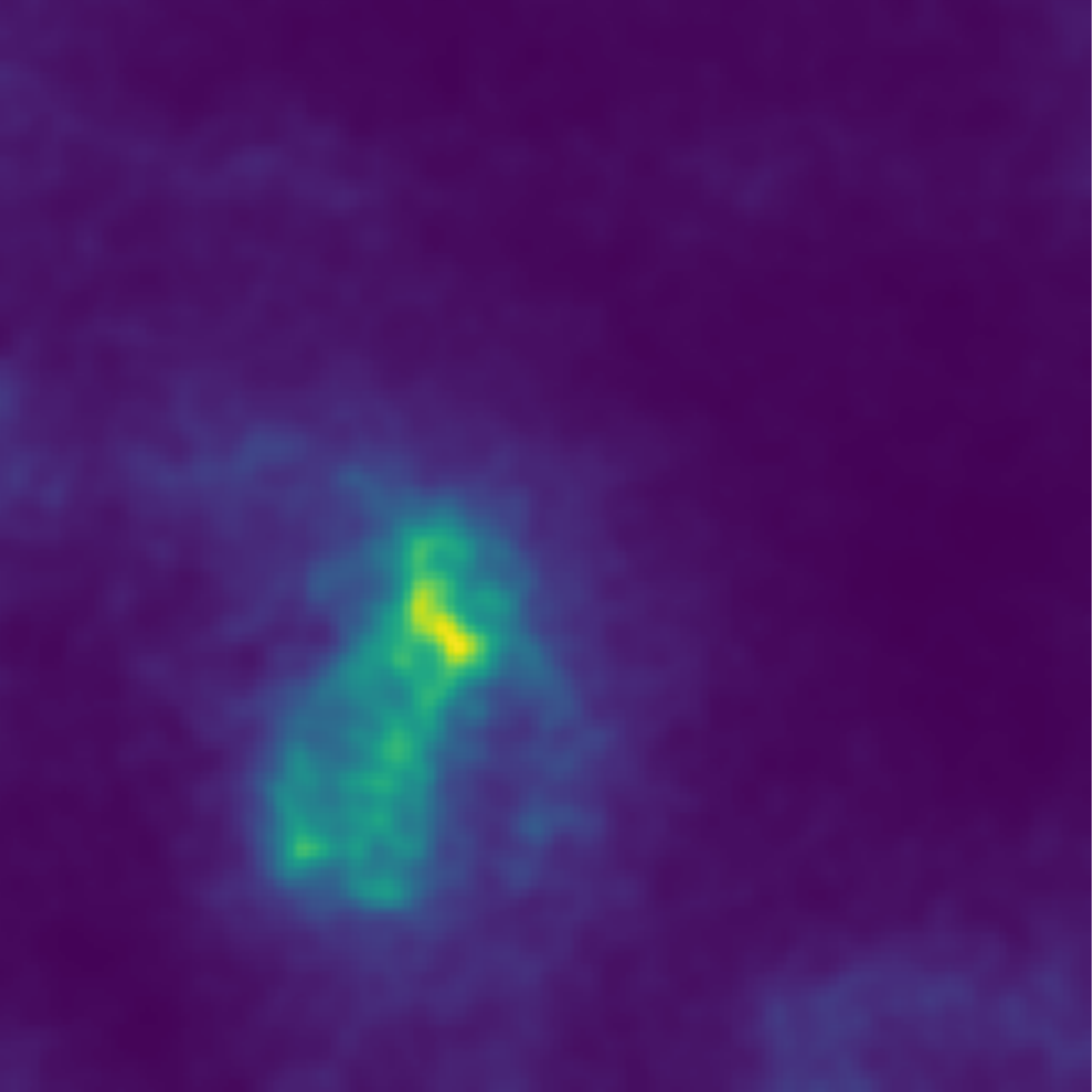} &
\includegraphics[width=\linewidth]{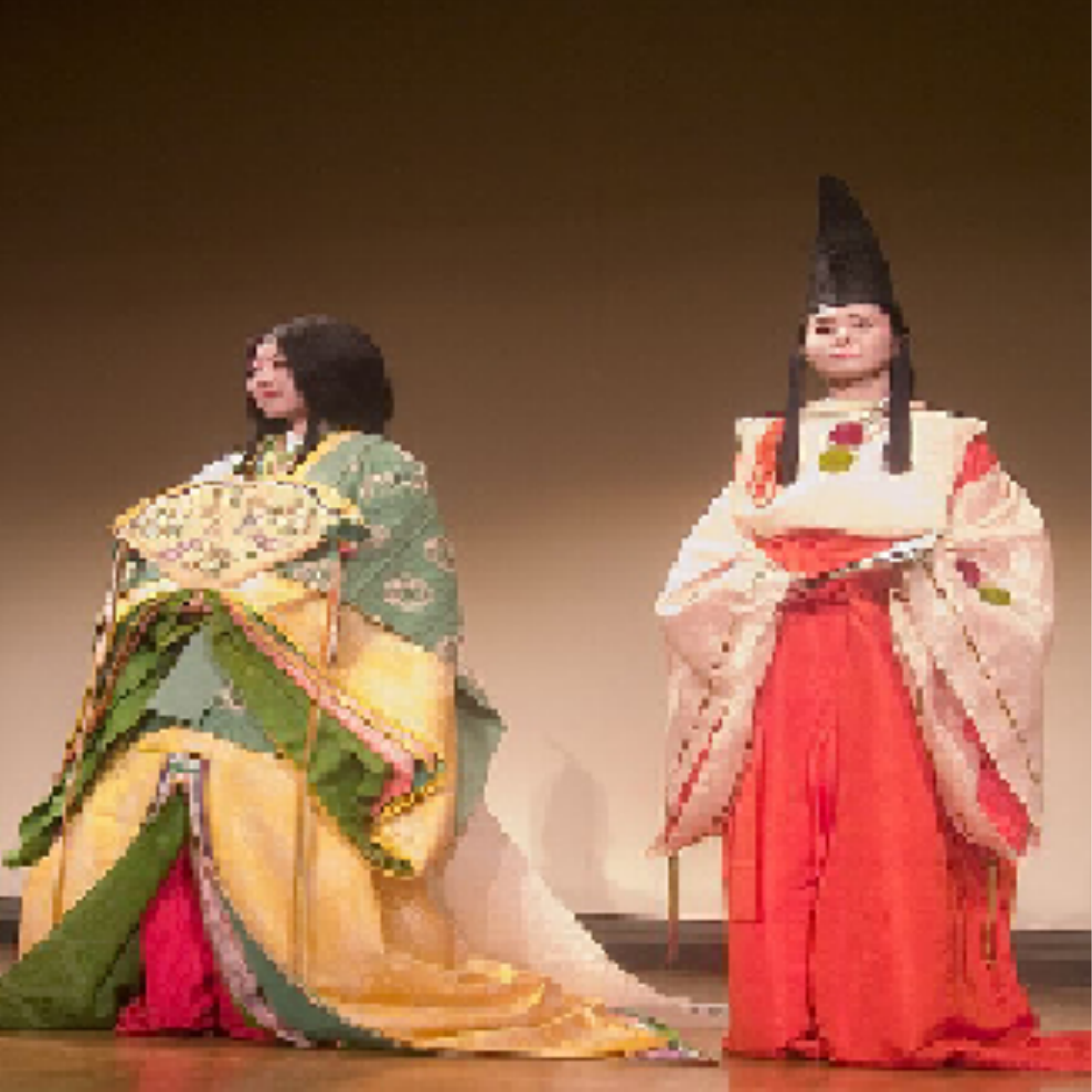}& \includegraphics[width=\linewidth]{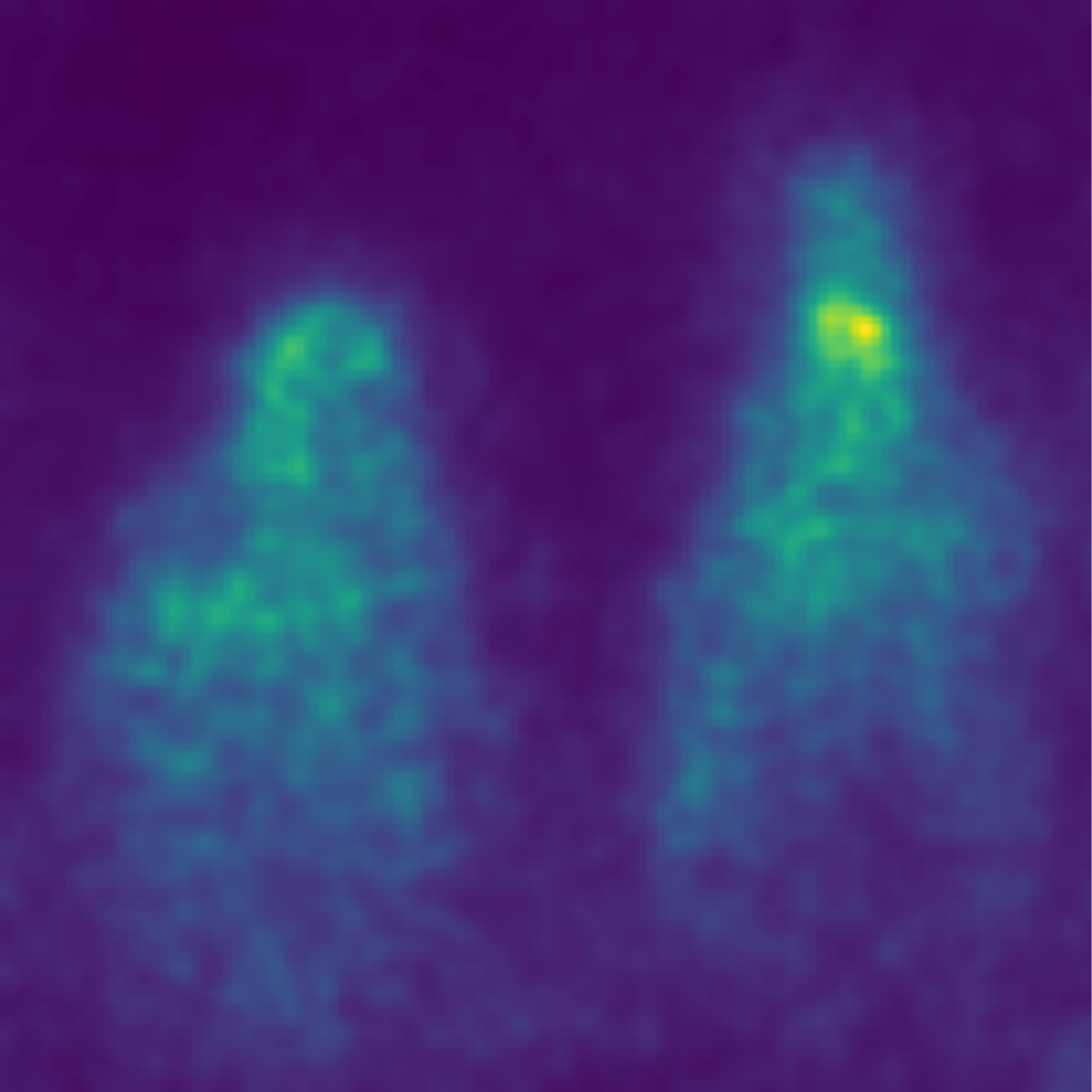}&
 \includegraphics[width=\linewidth]{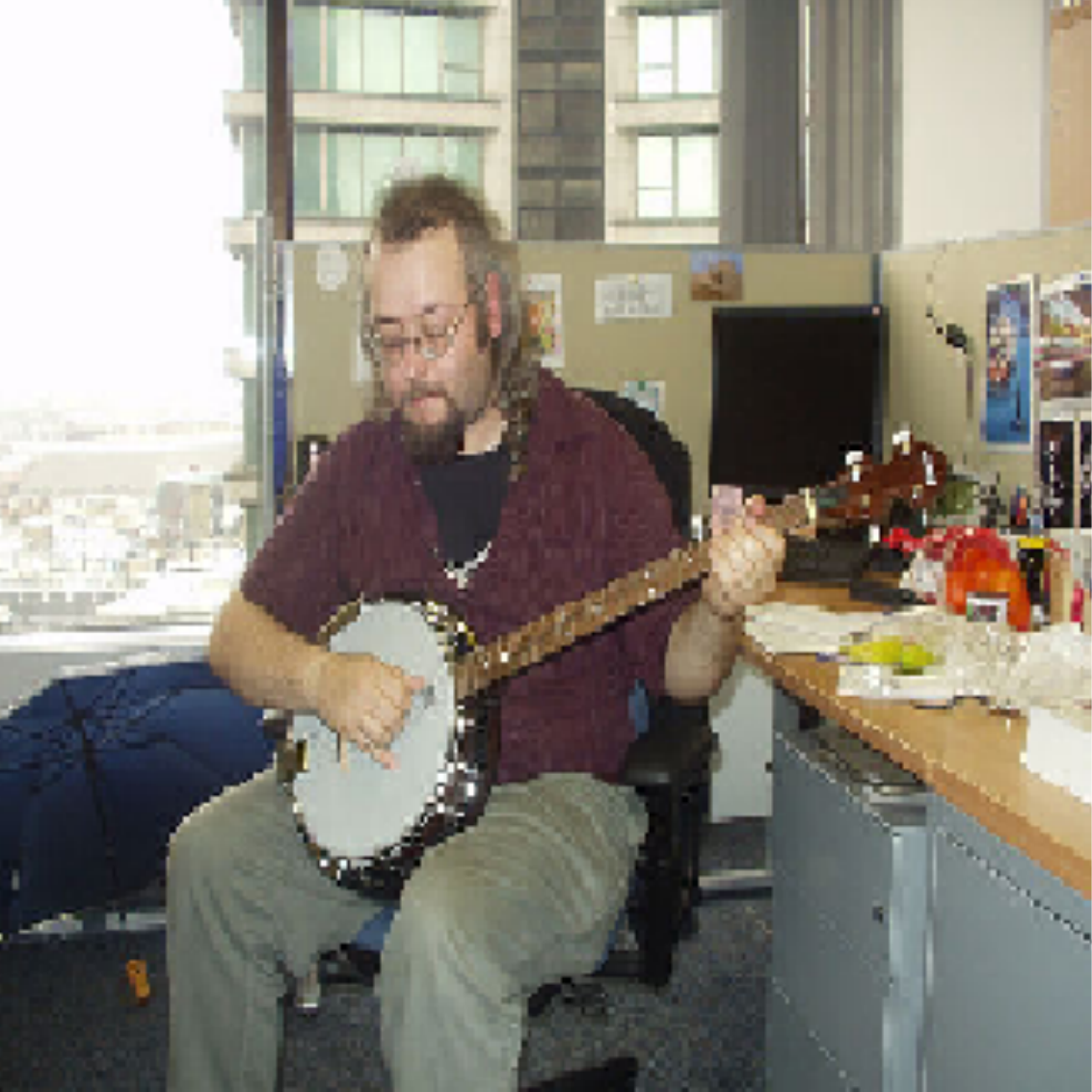}& \includegraphics[width=\linewidth]{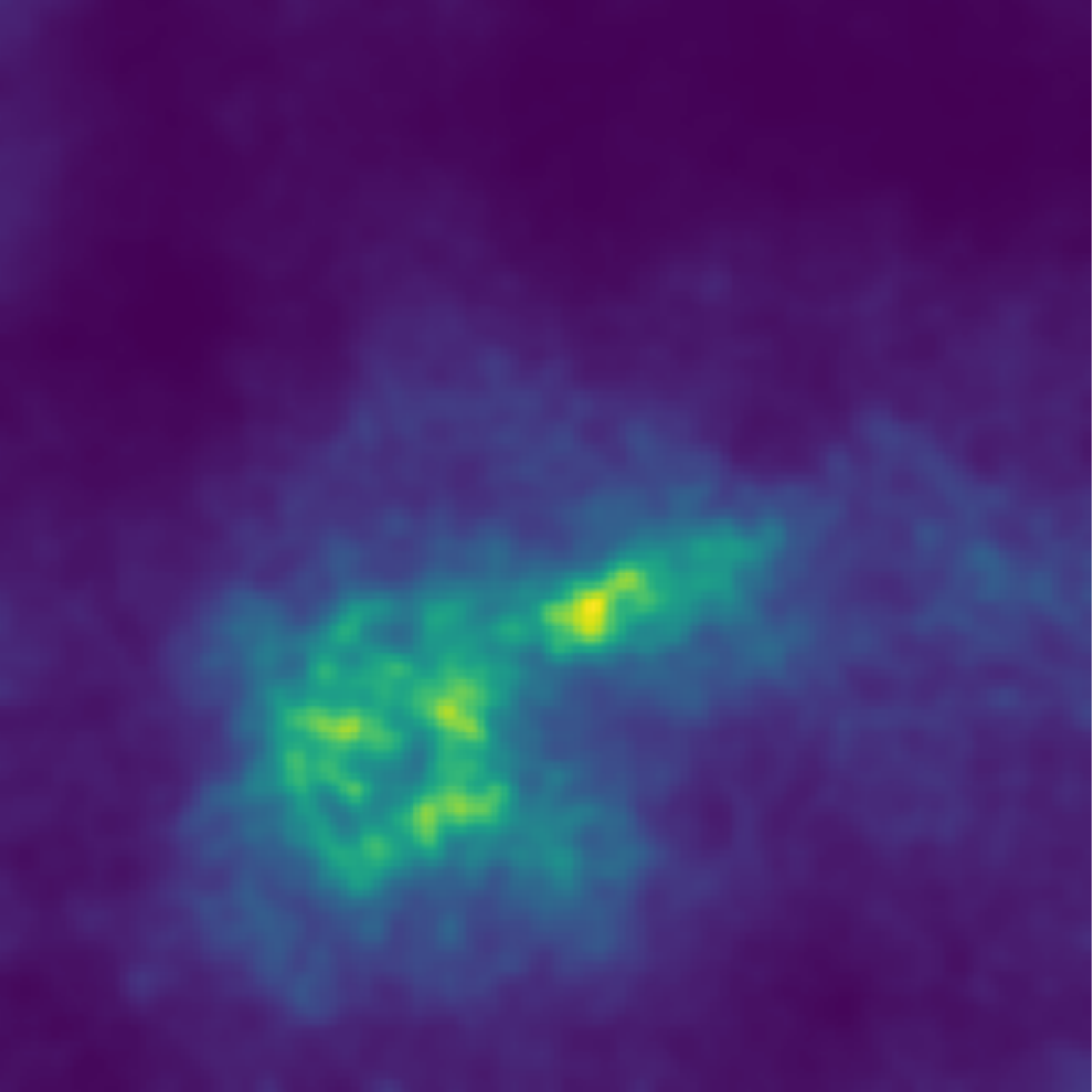} & 
\includegraphics[width=\linewidth]{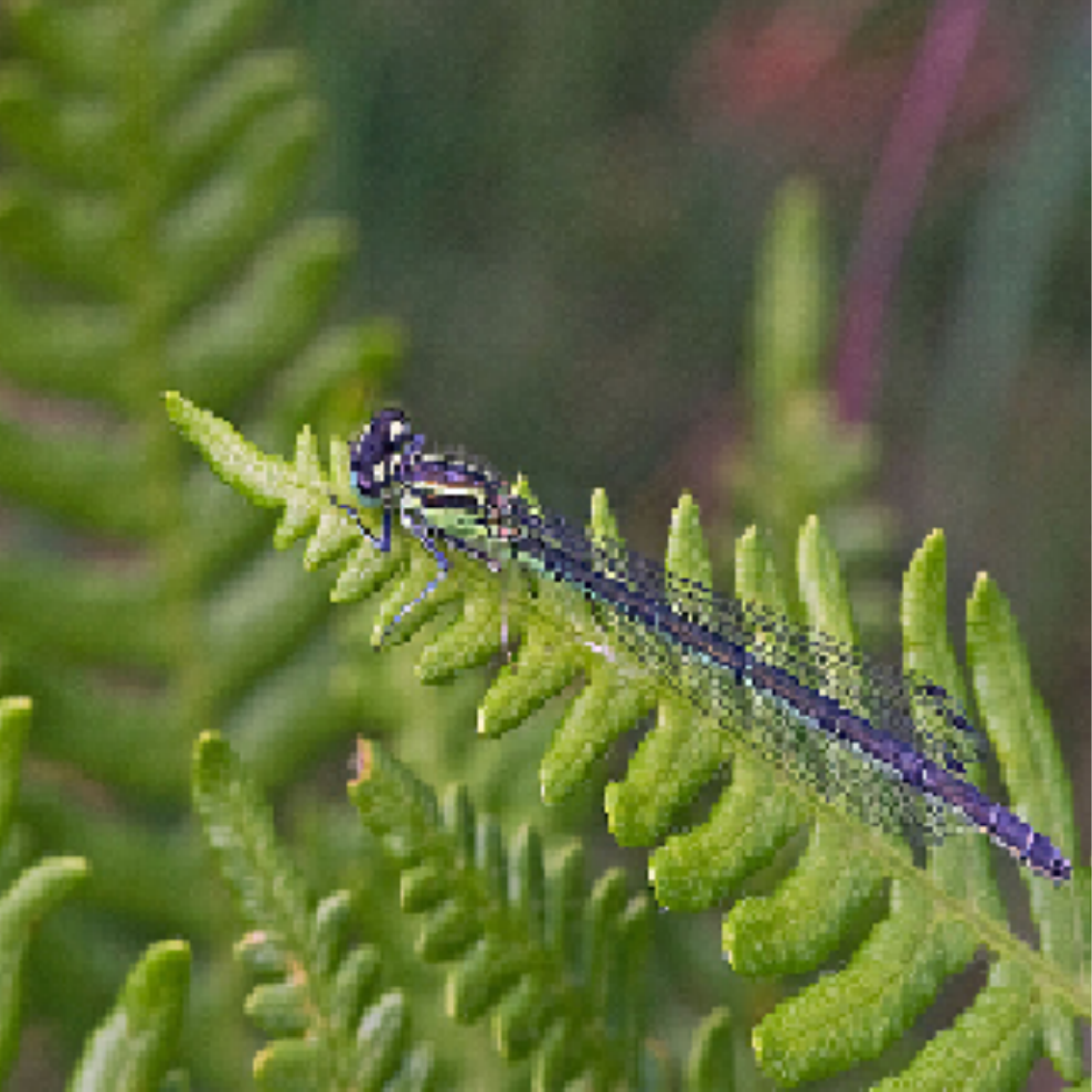}& \includegraphics[width=\linewidth]{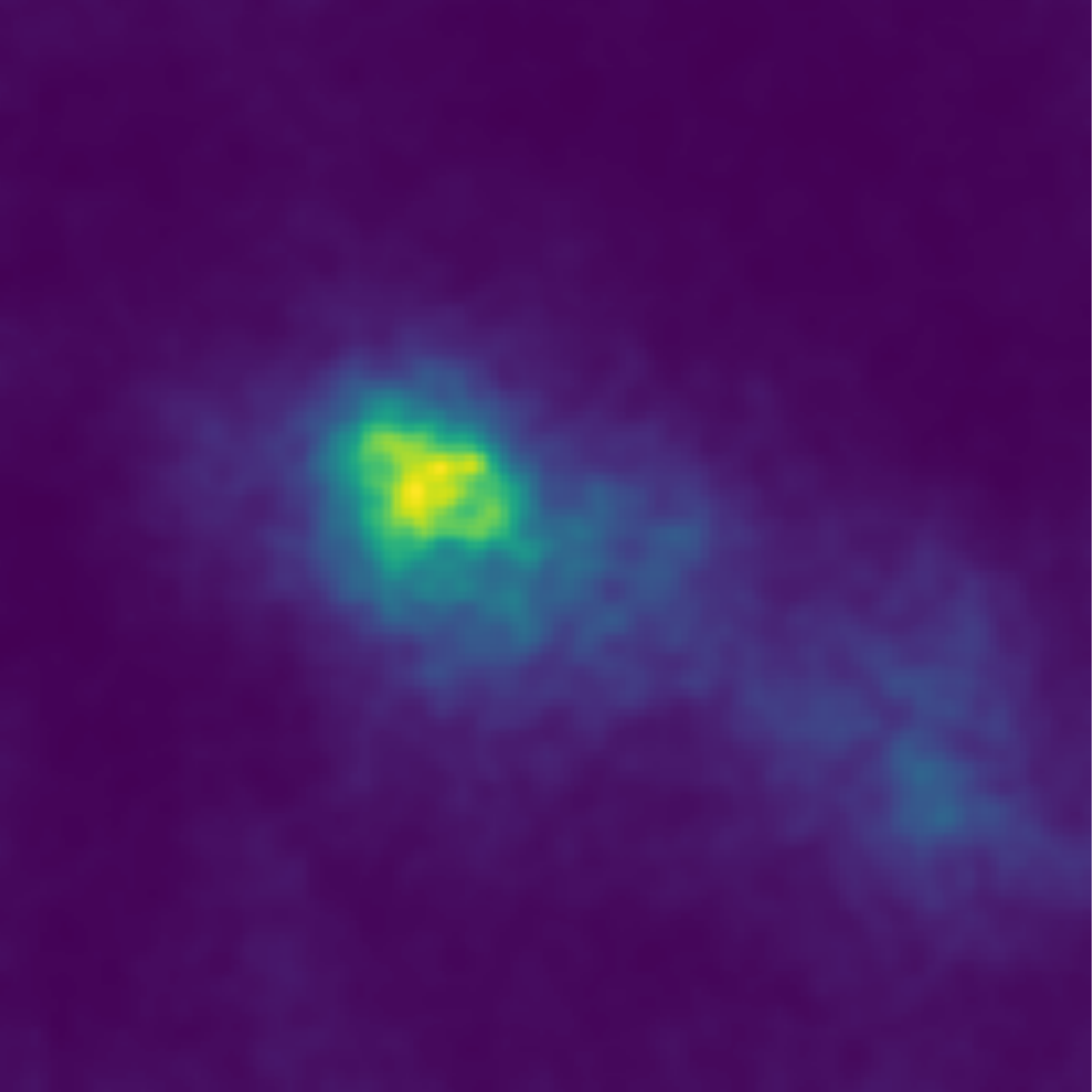} \\
\includegraphics[width=\linewidth]{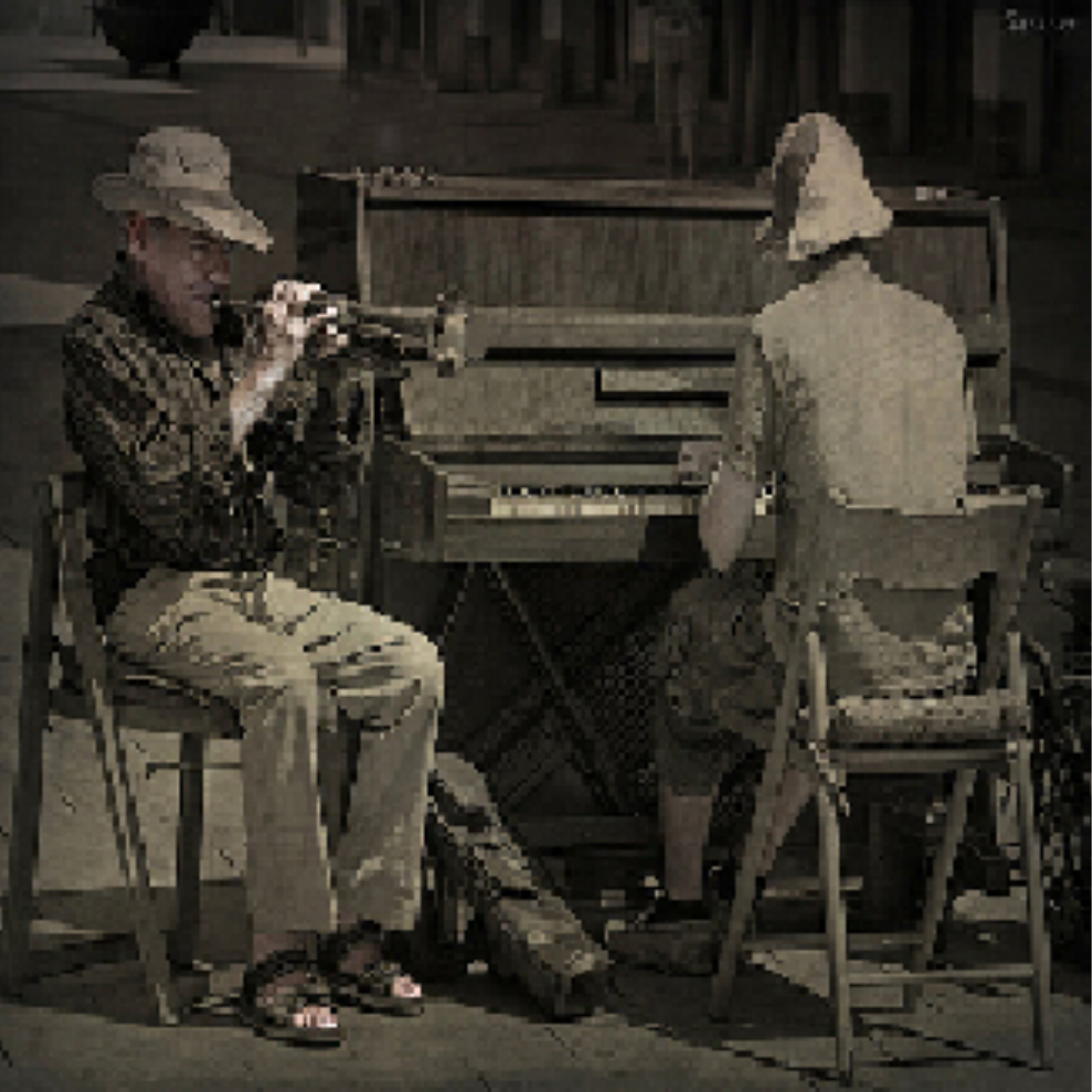}& \includegraphics[width=\linewidth]{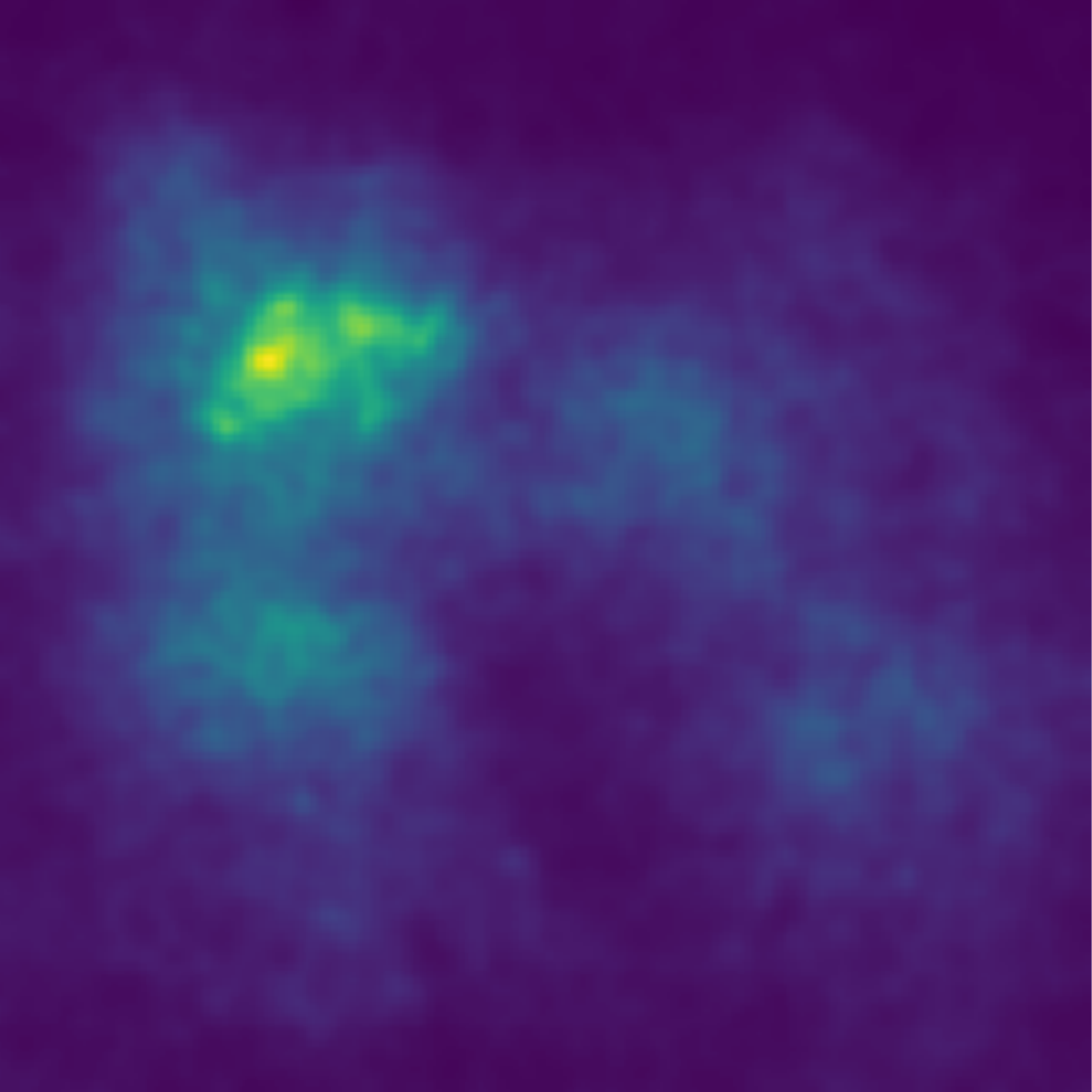} & 
\includegraphics[width=\linewidth]{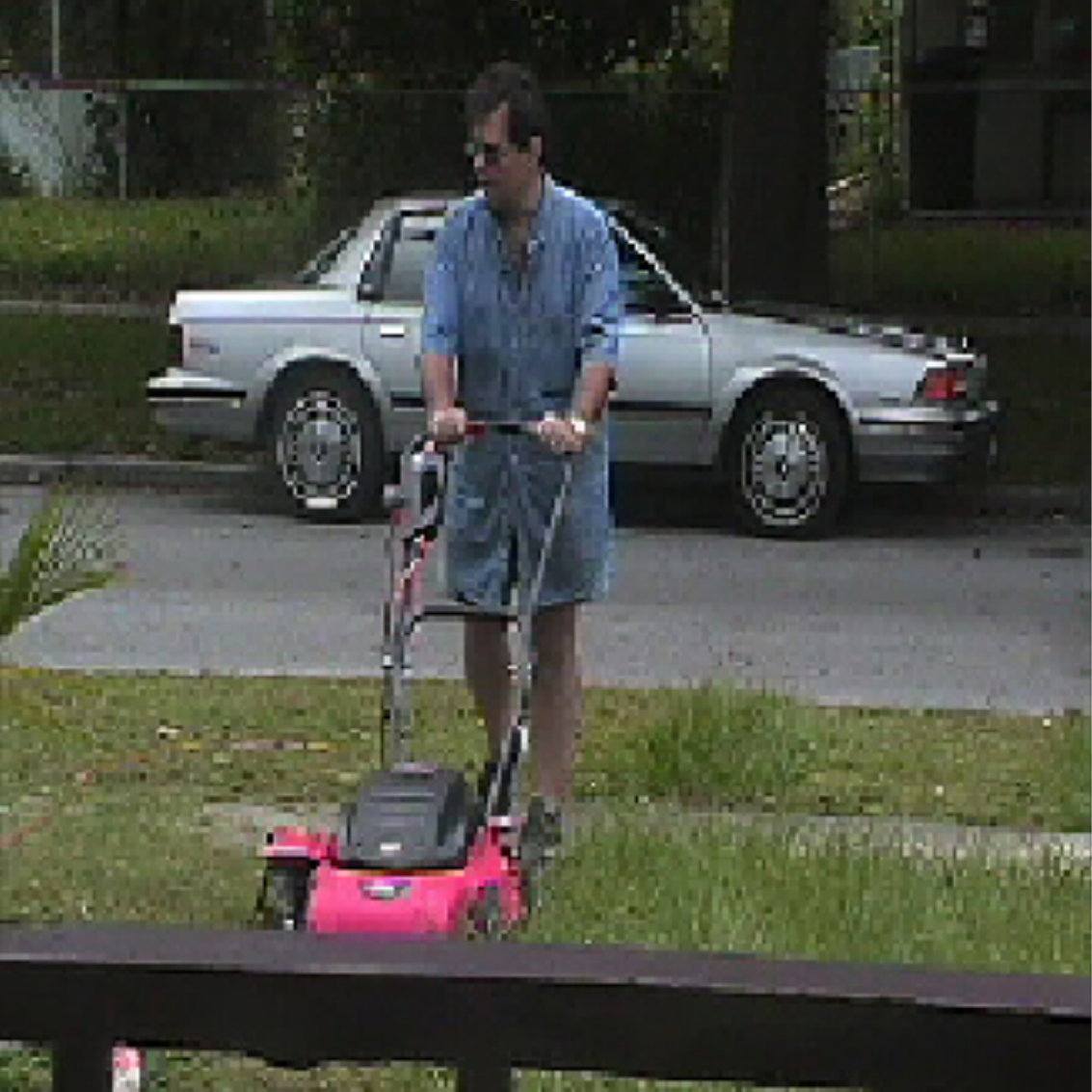}& \includegraphics[width=\linewidth]{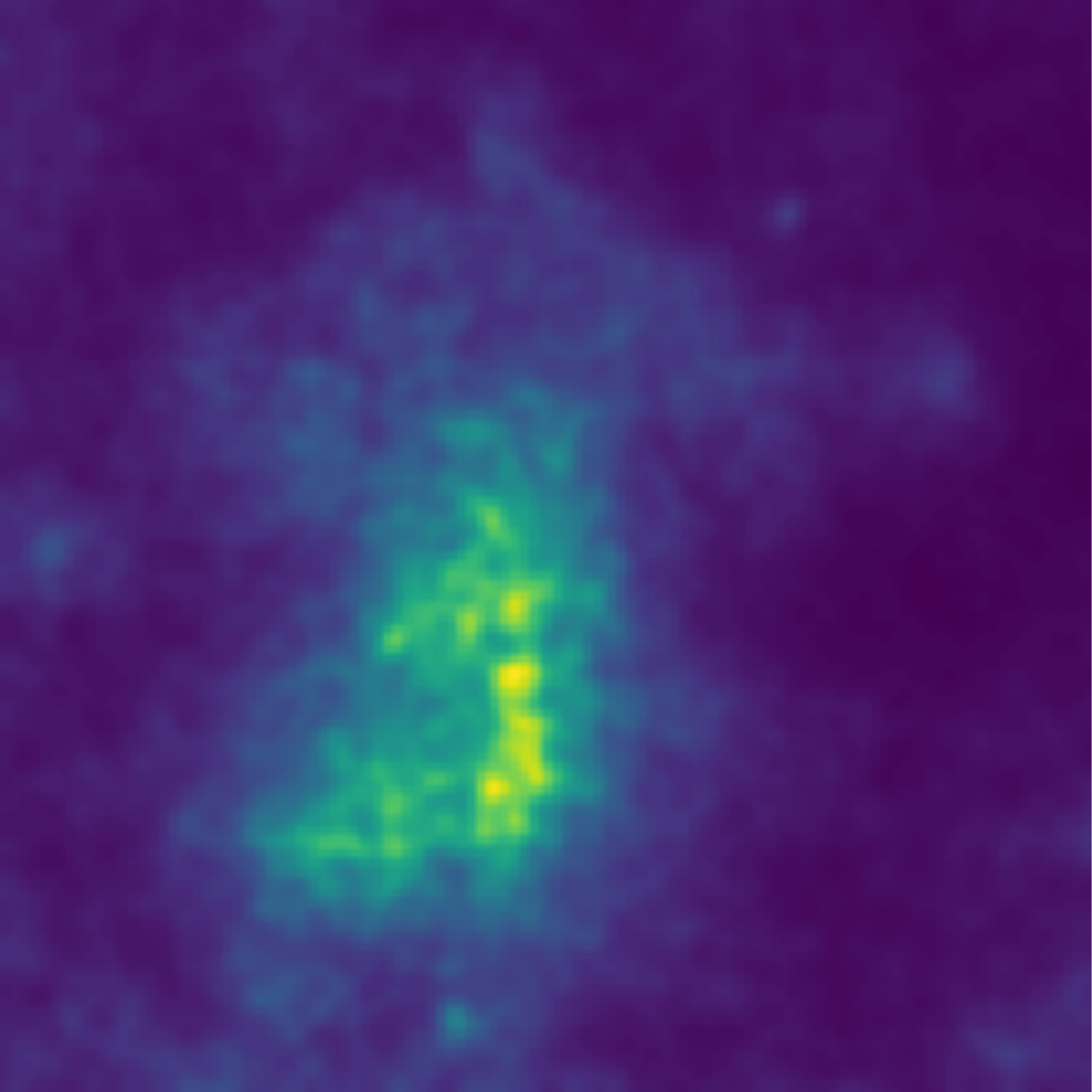}&
\includegraphics[width=\linewidth]{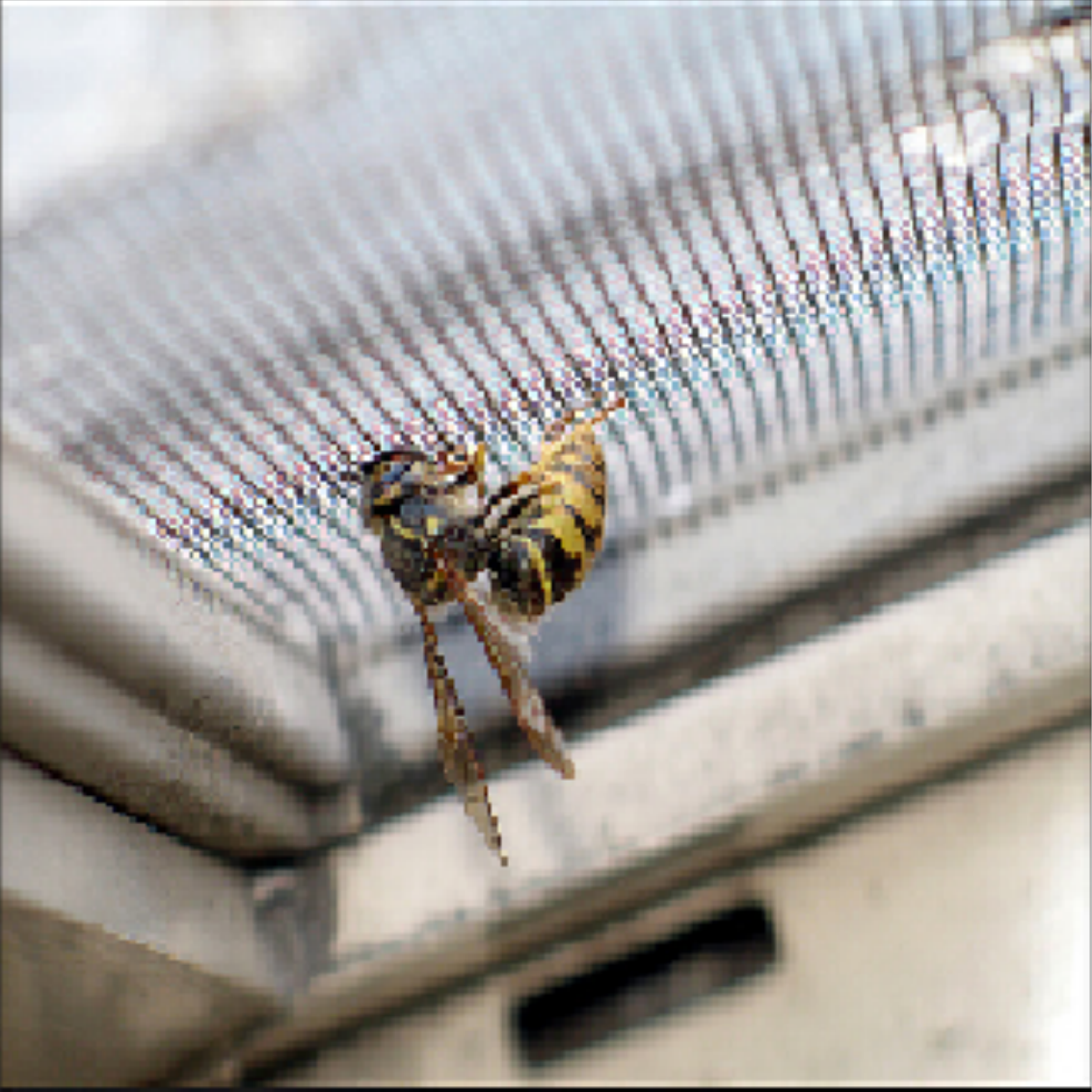}& \includegraphics[width=\linewidth]{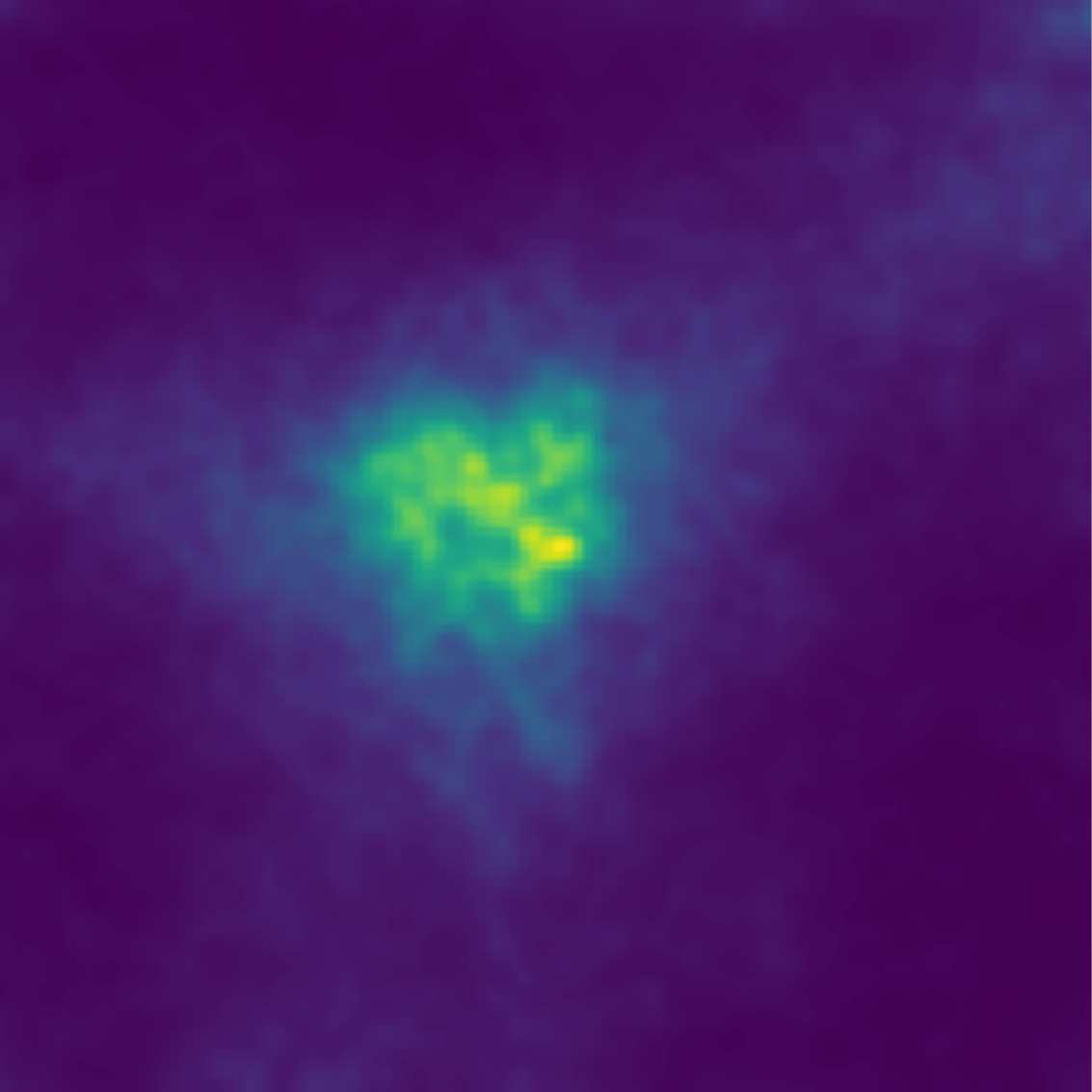} & 
\includegraphics[width=\linewidth]{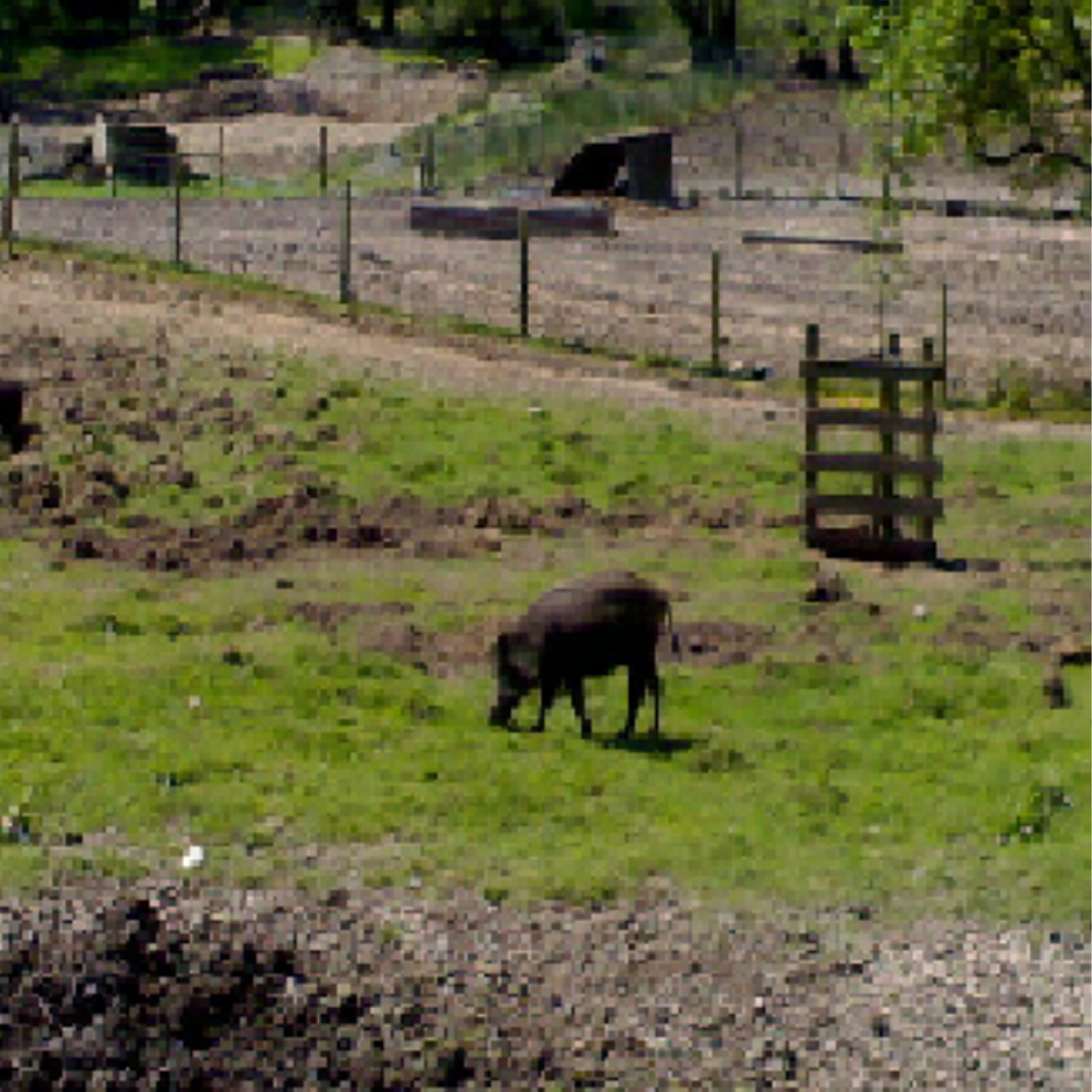}& \includegraphics[width=\linewidth]{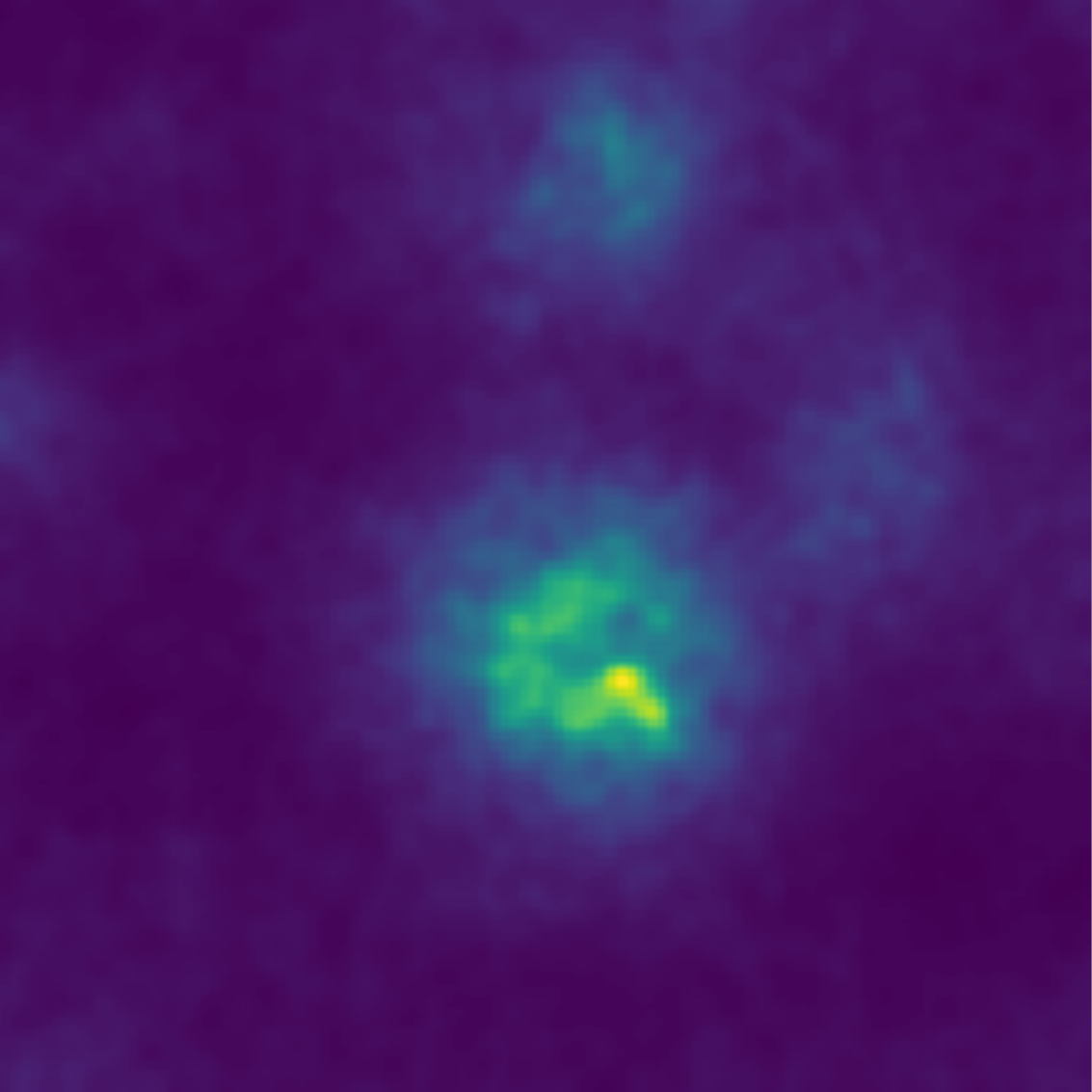}&
\includegraphics[width=\linewidth]{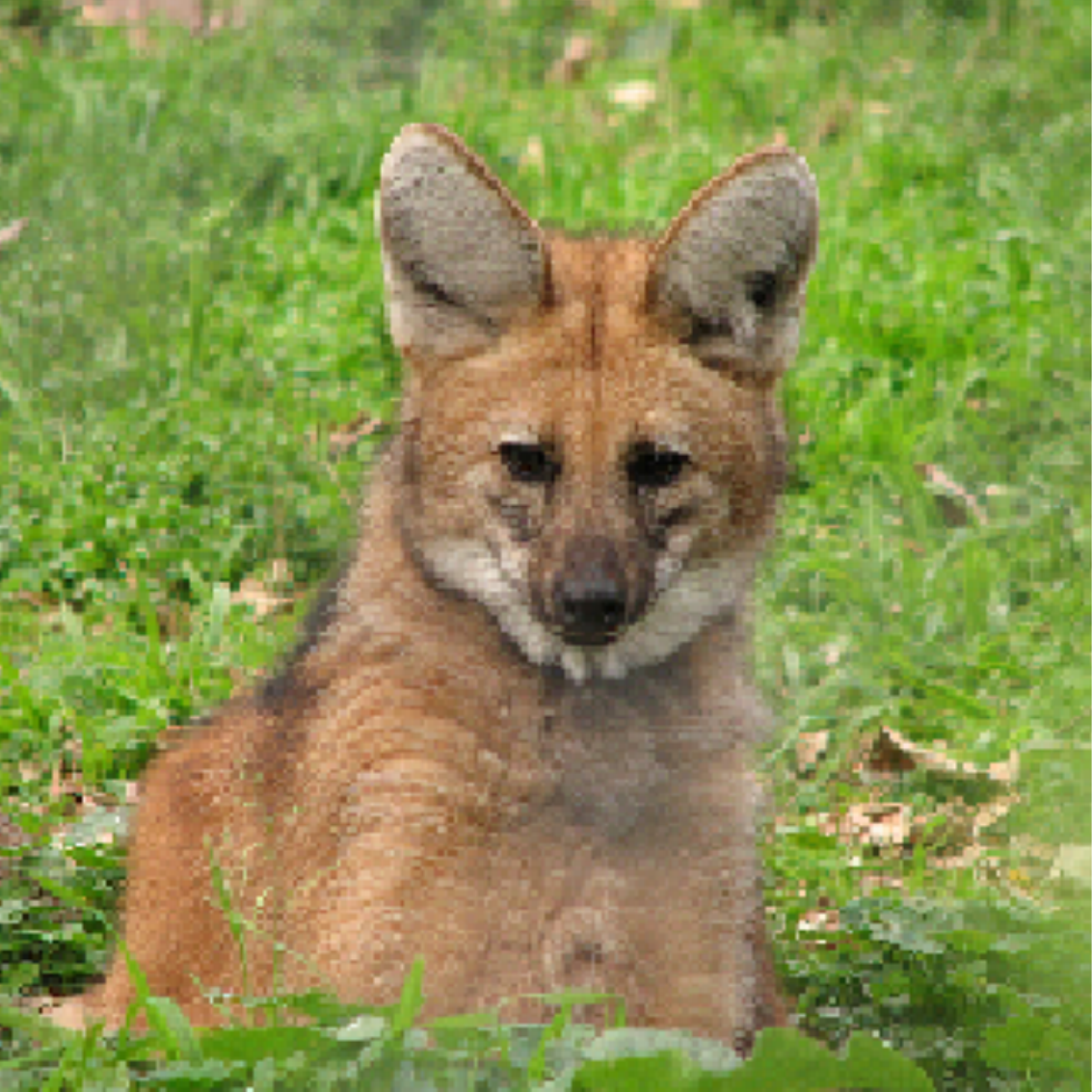}& \includegraphics[width=\linewidth]{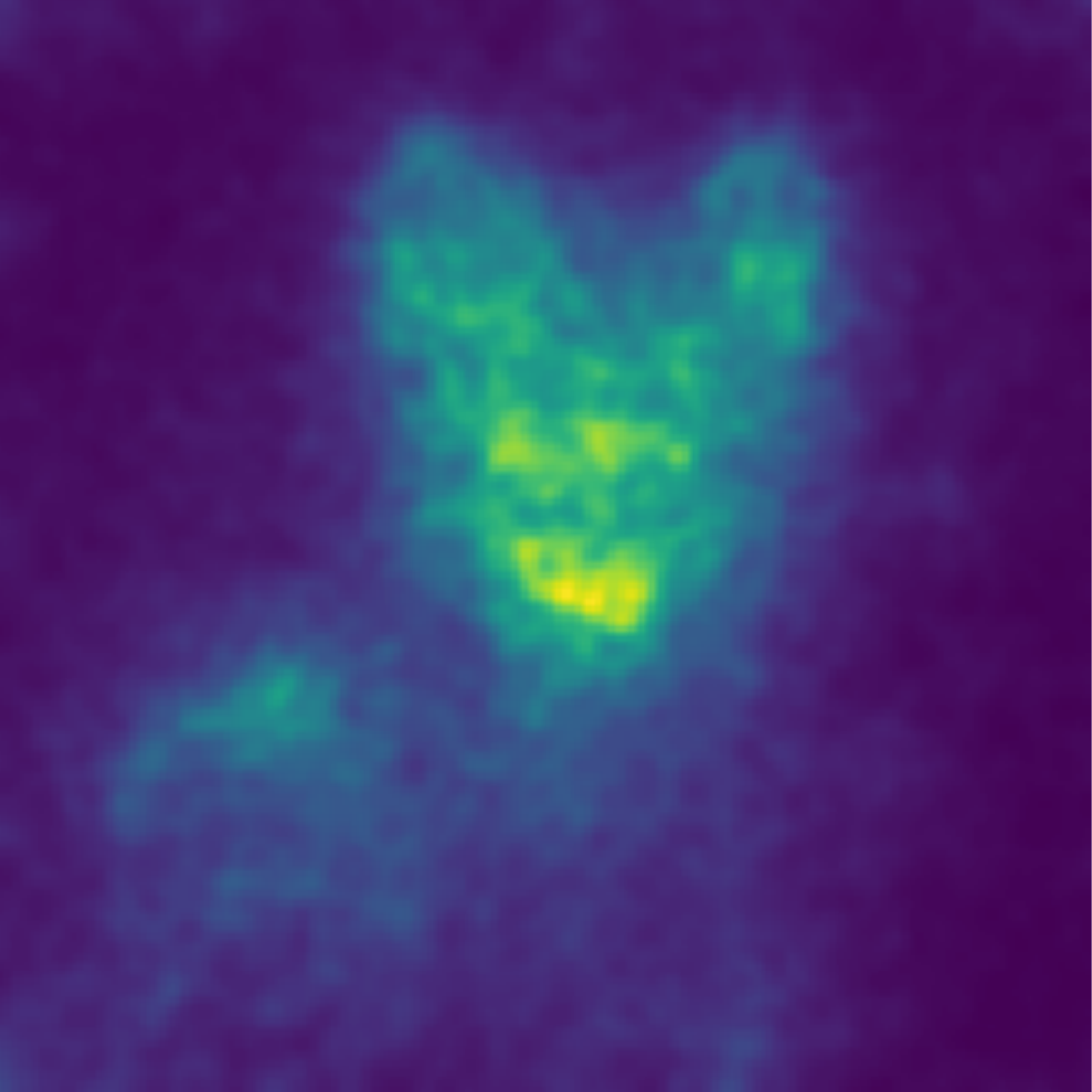} \\ 
\includegraphics[width=\linewidth]{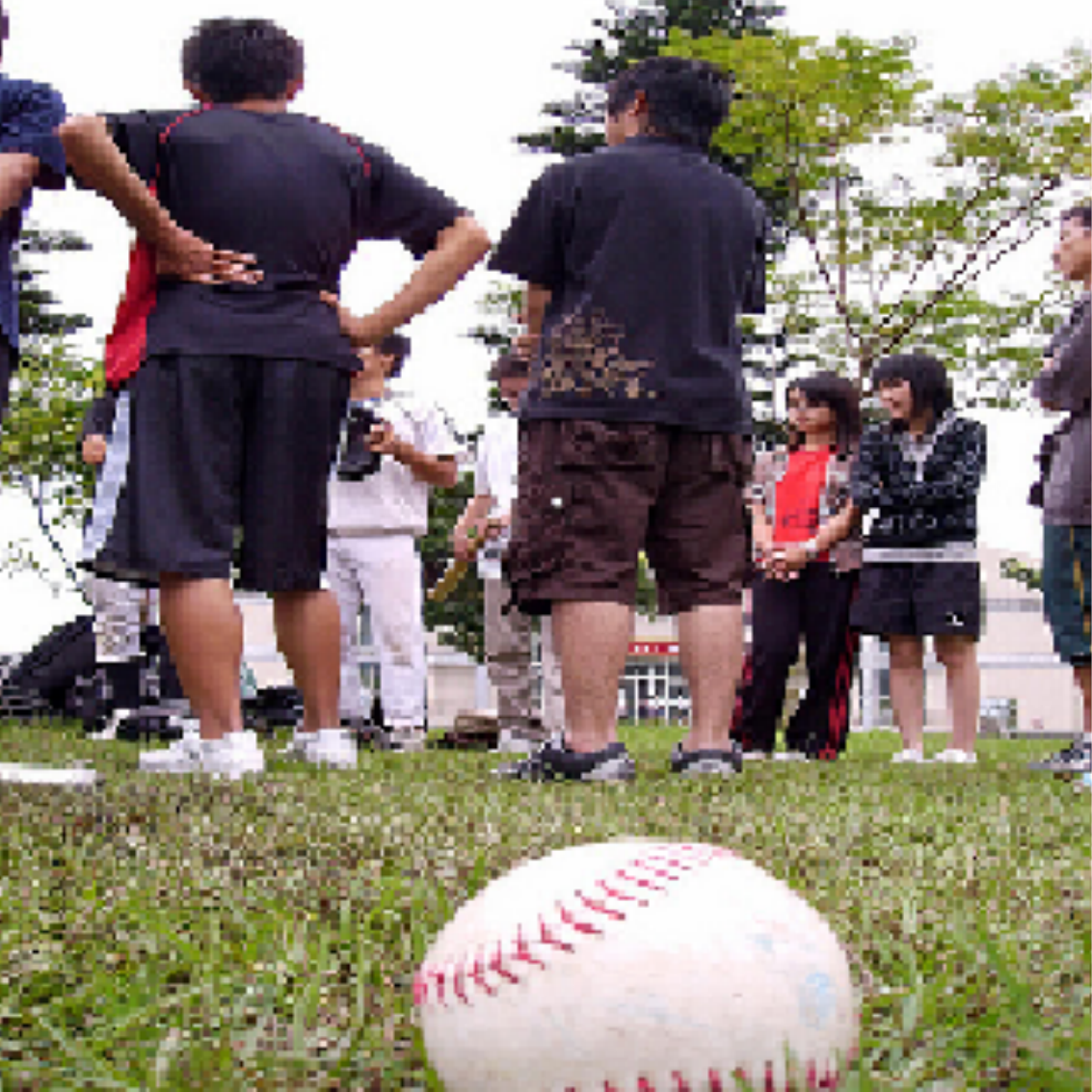}& \includegraphics[width=\linewidth]{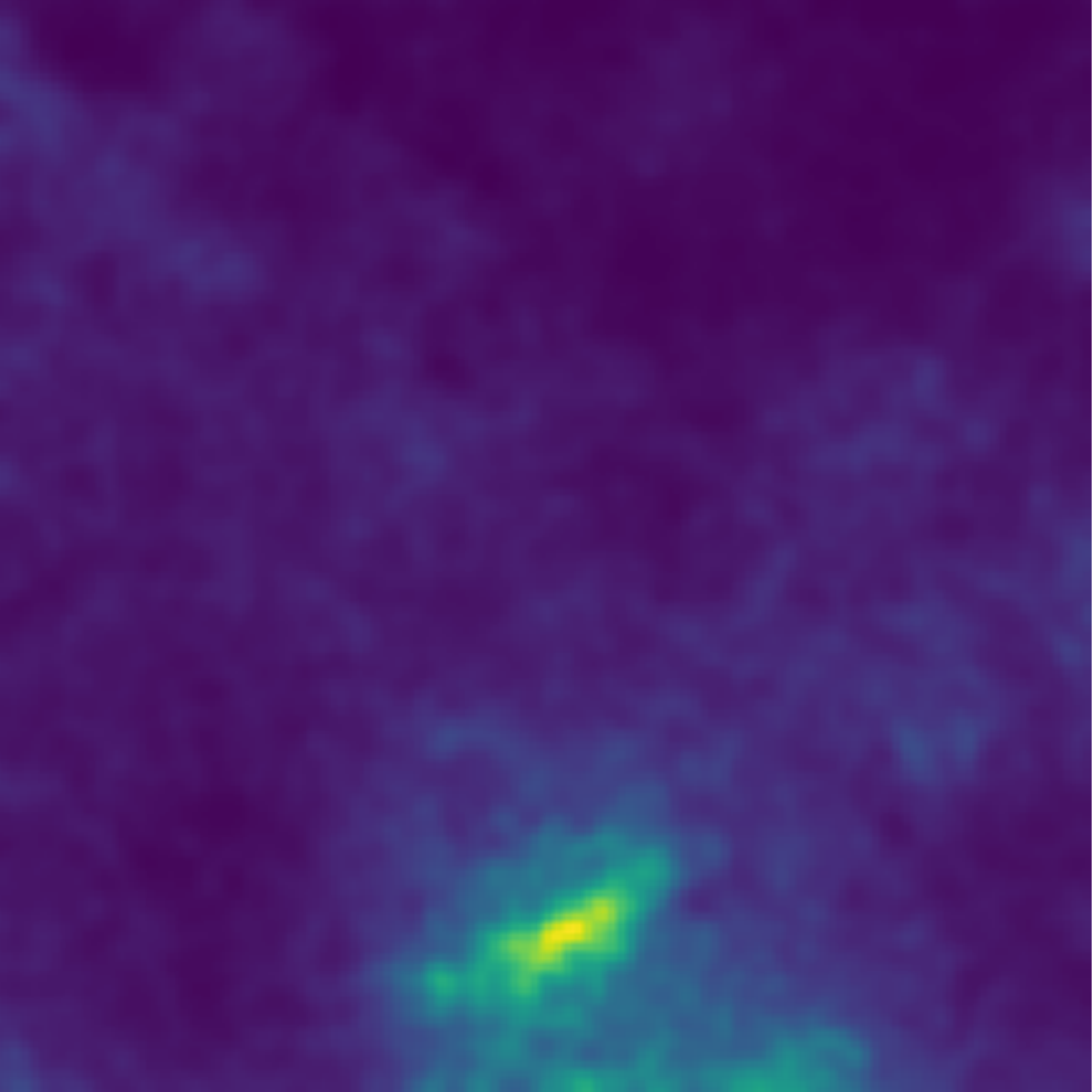}&
\includegraphics[width=\linewidth]{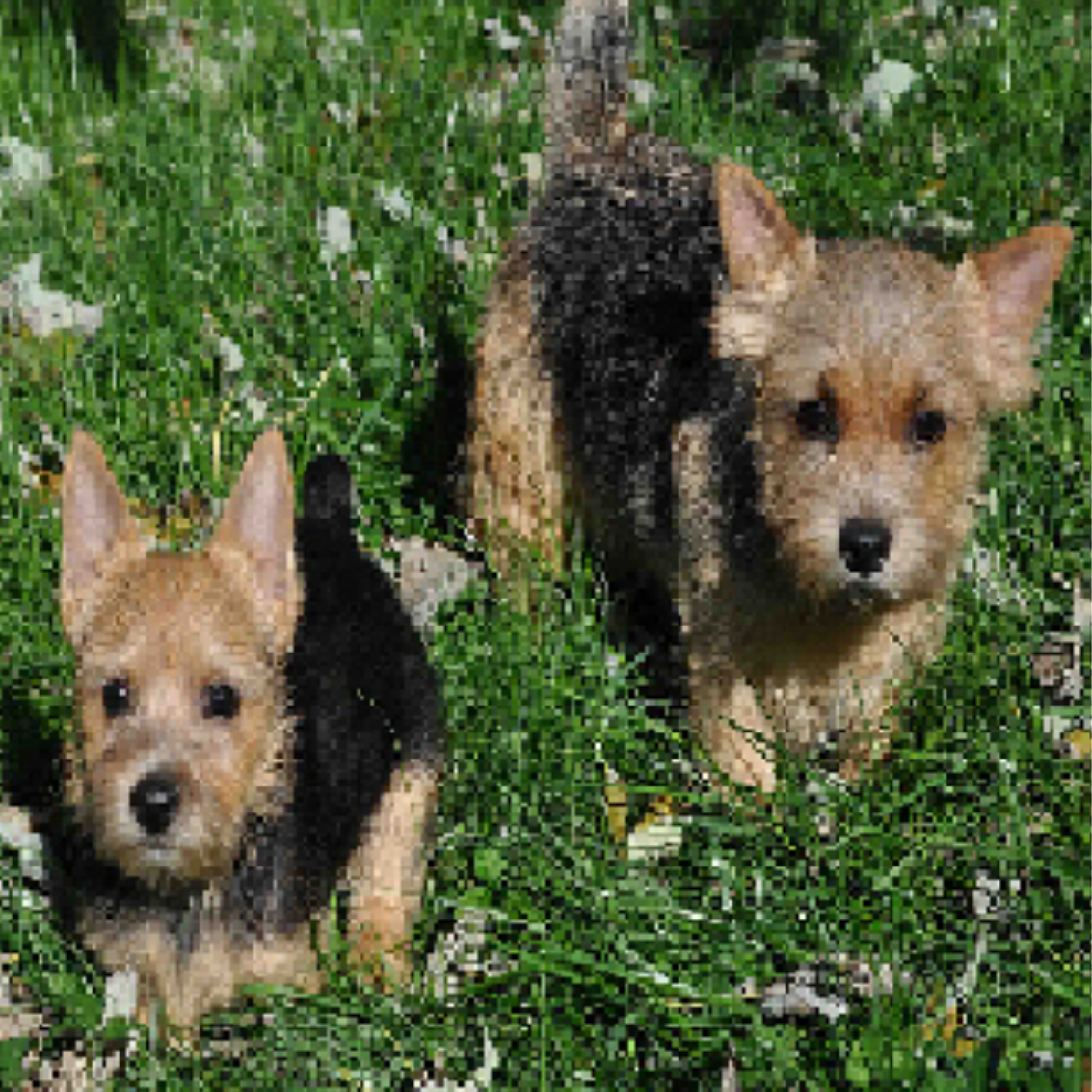}& \includegraphics[width=\linewidth]{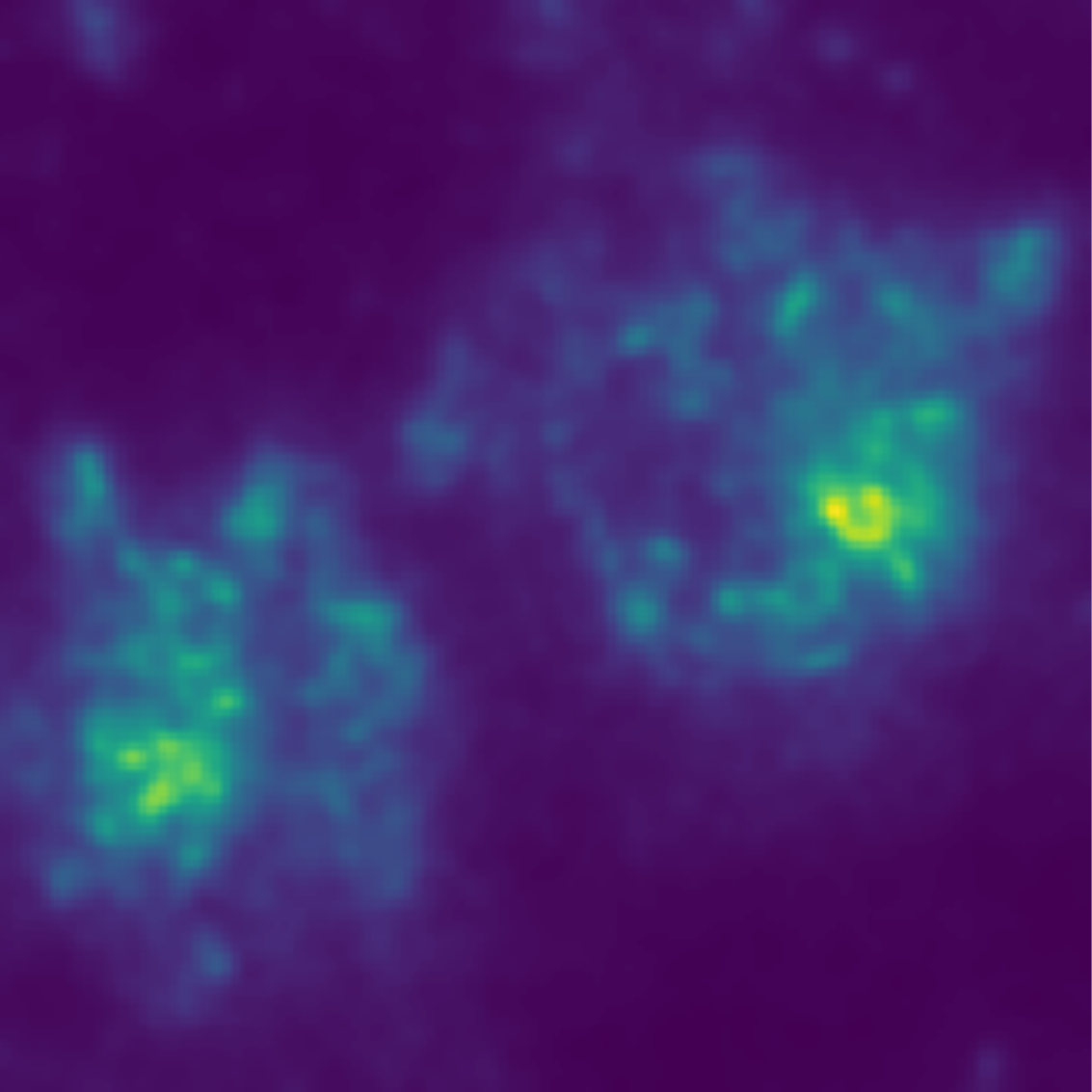} &
\includegraphics[width=\linewidth]{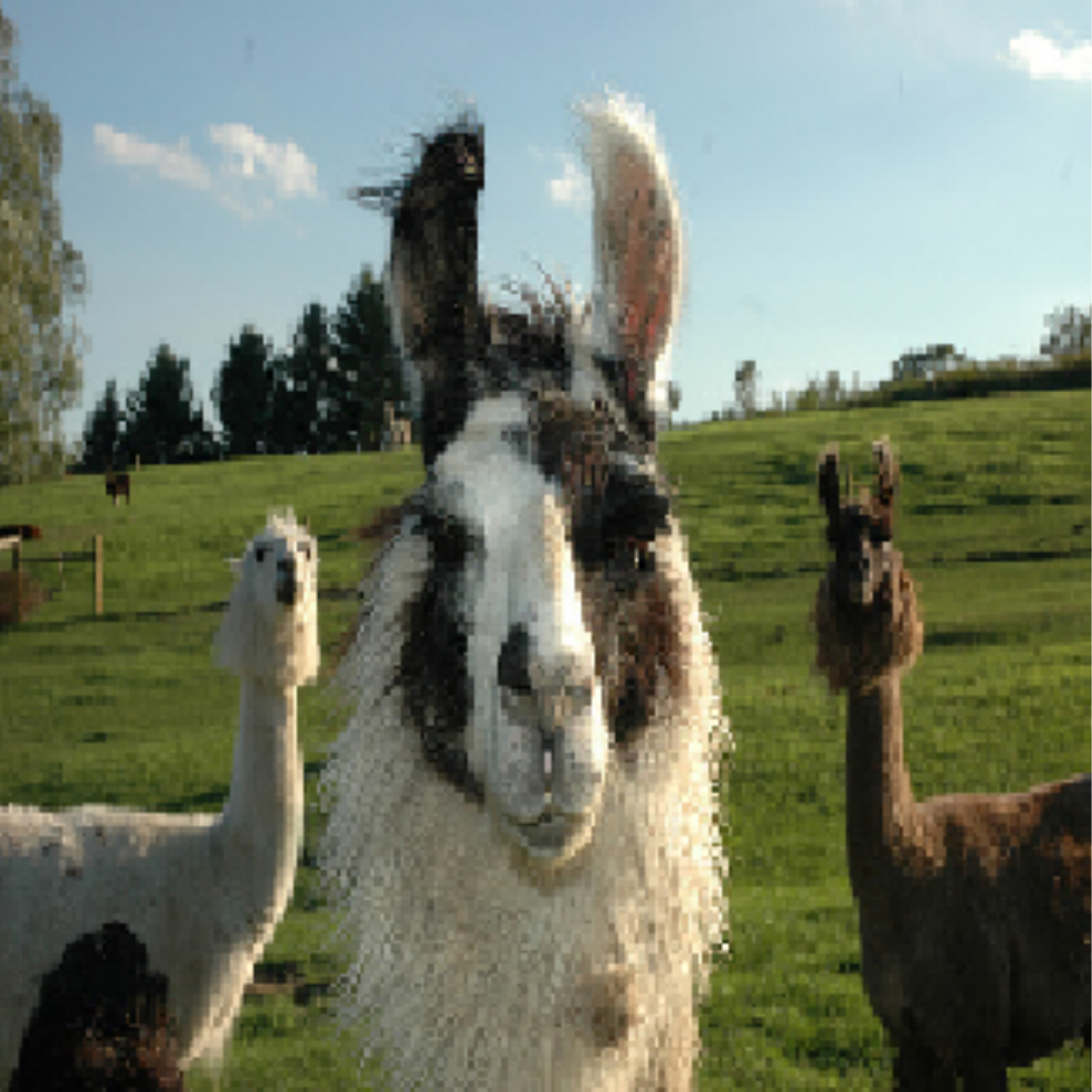}& \includegraphics[width=\linewidth]{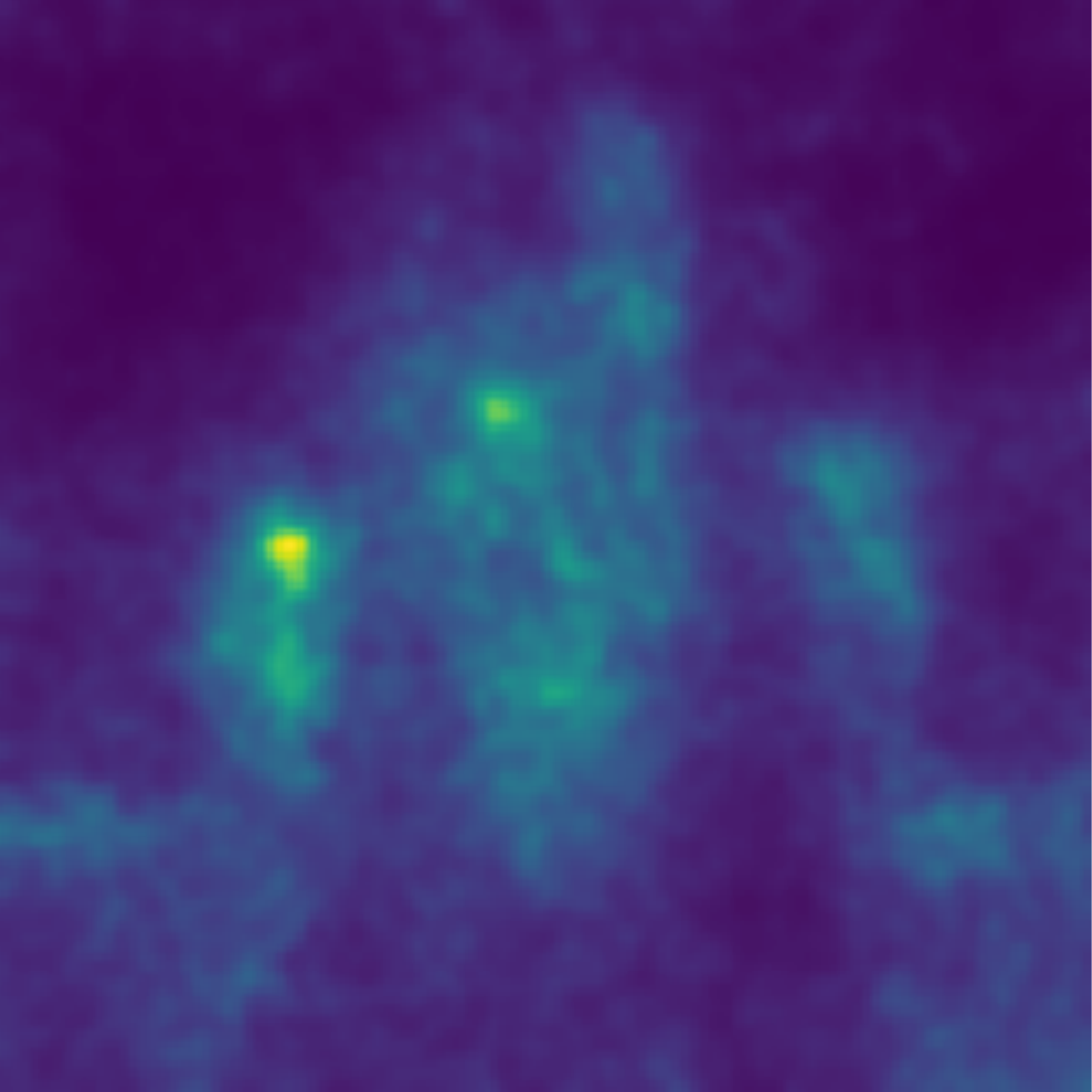}&
\includegraphics[width=\linewidth]{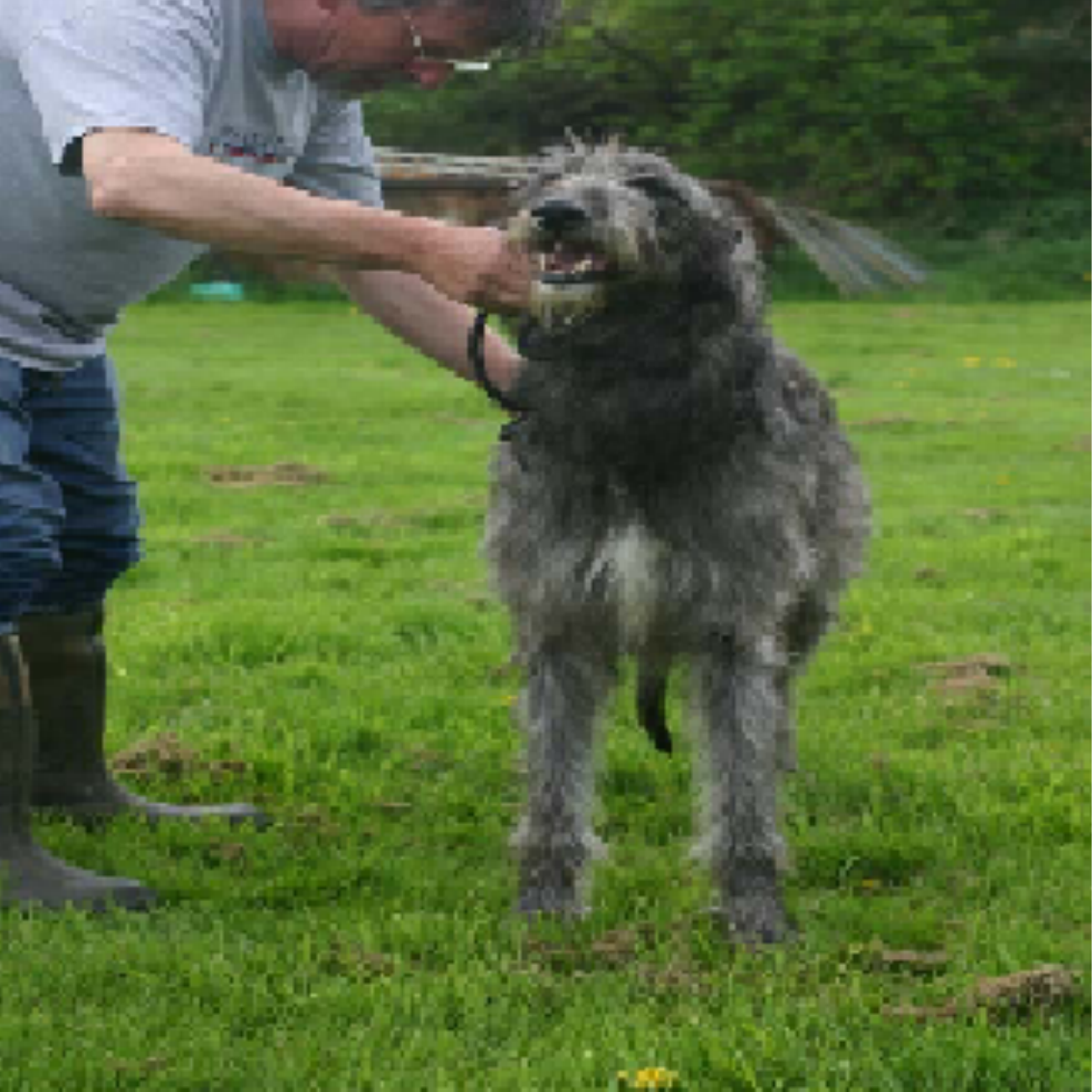}& \includegraphics[width=\linewidth]{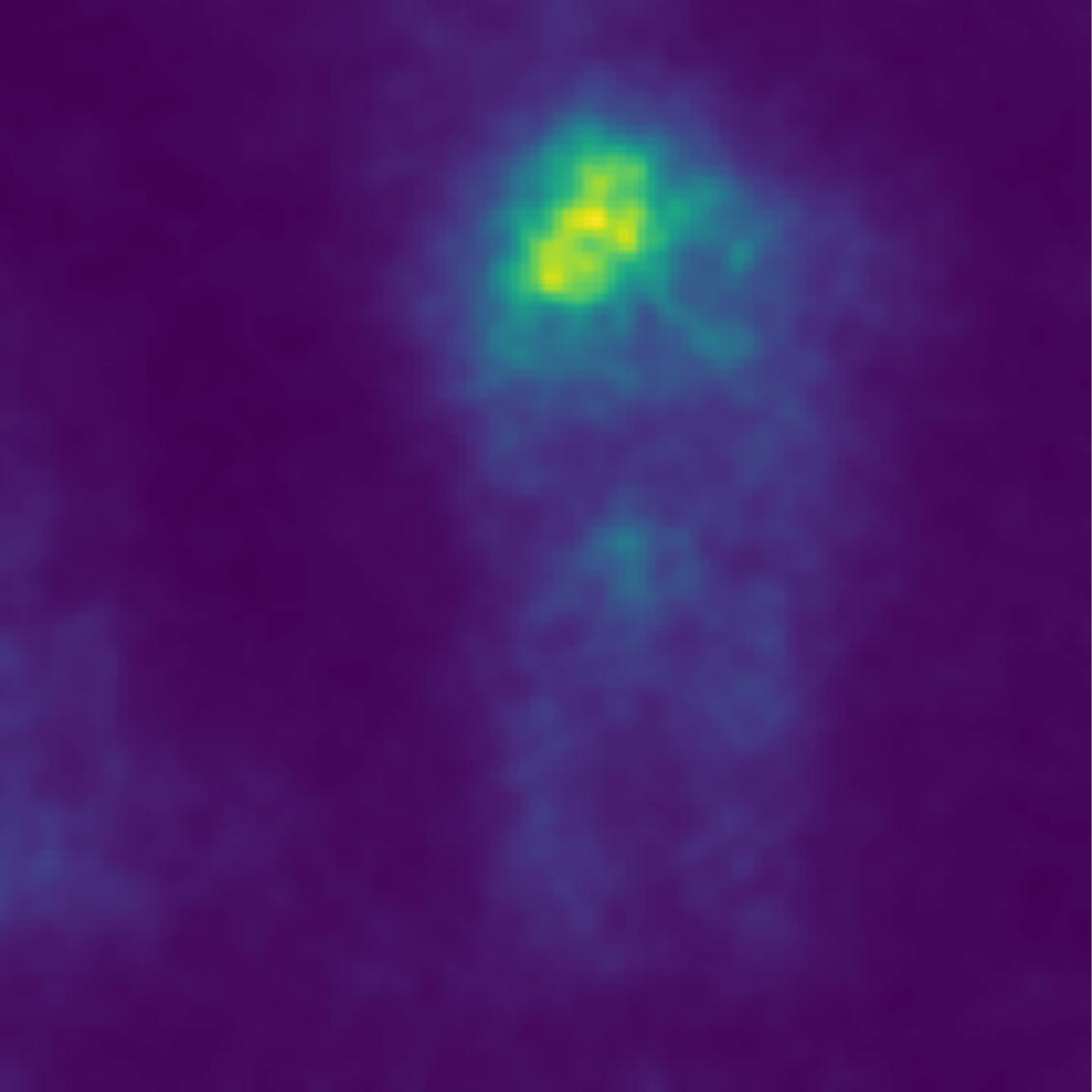} & 
\includegraphics[width=\linewidth]{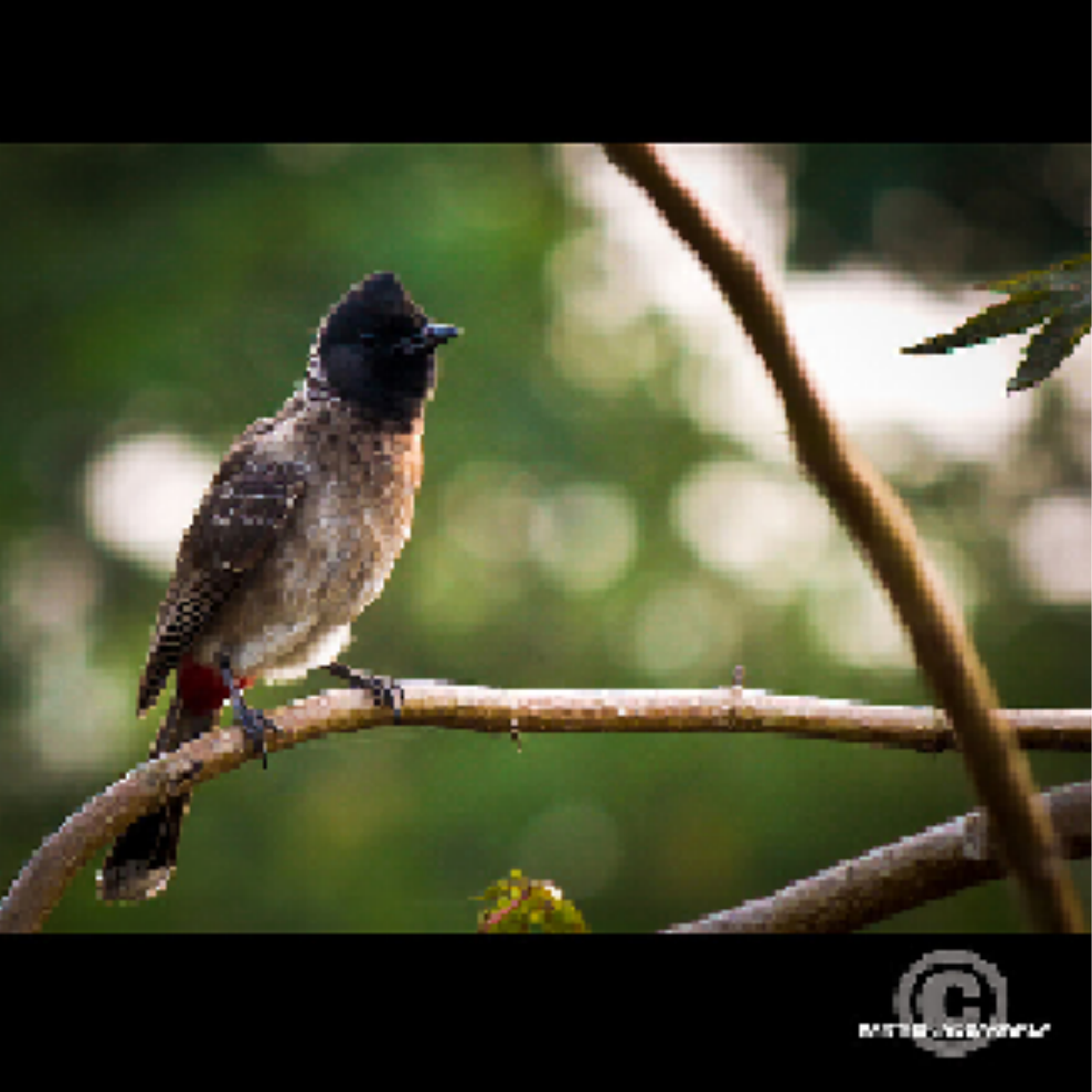}& \includegraphics[width=\linewidth]{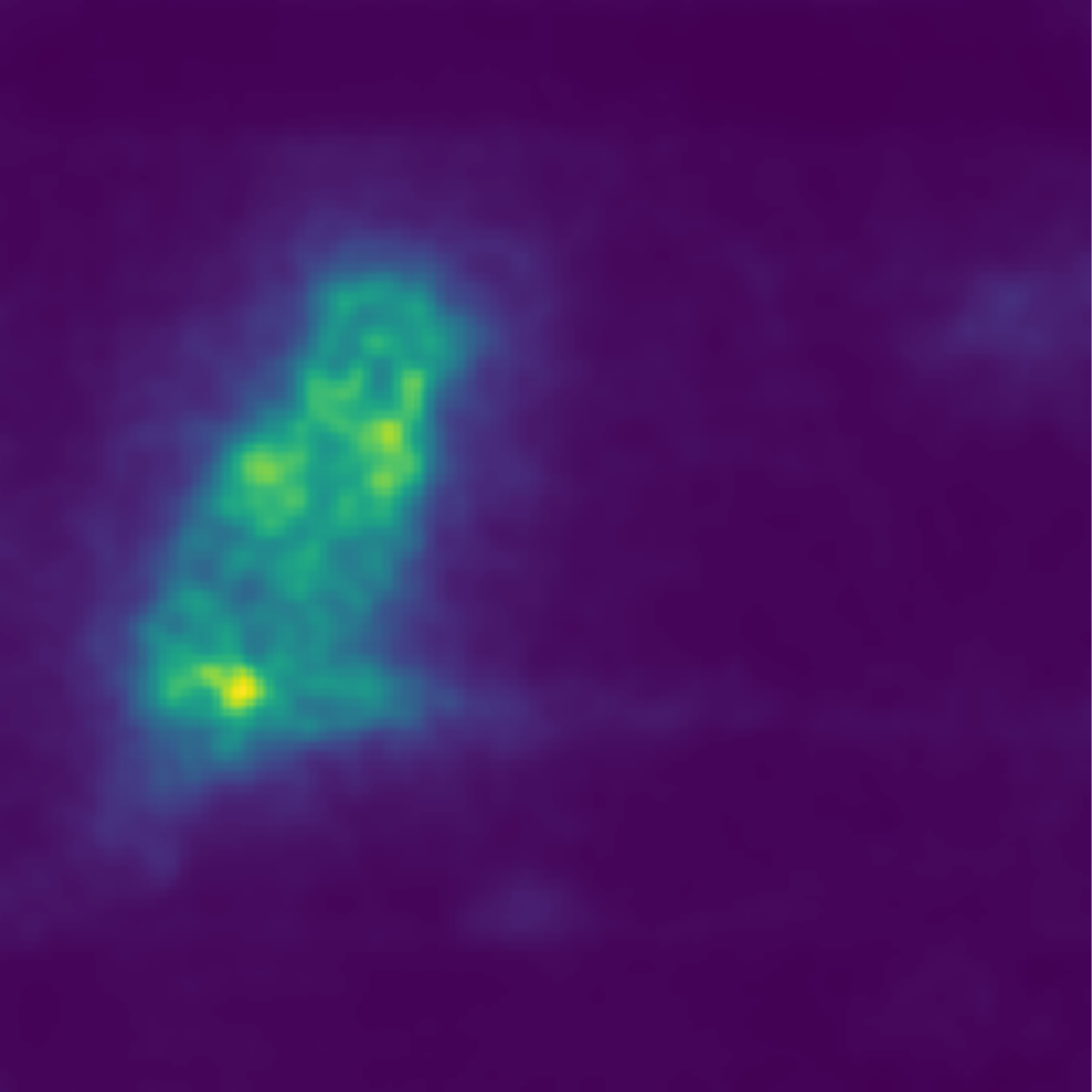} \\
\end{tabular}
\endgroup
 \caption{Perceptual Perturbations on ImageNet as explanations. Illustration of perceptual perturbations on typical images taken from ImageNet \cite{ILSVRC15}.  See discussion in Section~\ref{sec:visual-explanations}, and for further examples see Fig.~S1 in supplementary materials section A. \label{fig:ppe}}
\end{figure*}

\section{Methodology}
\label{sec:methodology}
We consider a classifier $C(\cdot)$ that takes an image $x$ as input, and returns a $k$ dimensional confidence vector. 

For classifiers that assign a single class to each image, we assume the classifier $C(\cdot)$ assigns the label $i=\arg\max_j C_j(x)$ to the image $x$.
Given image $x$ classified as label $i$ we consider the scalar multi-class margin:
\begin{equation}
 M_i(x')=C_i(x') -\max_{j\neq i}C_j(x') \label{margin}
\end{equation}
and note that $M_i(x')\le 0$ if and only if $C(\cdot)$ does not assign label $i$
to image $x'$.

For classifiers $C(\cdot)$ that assign multiple classes to a single image 
(e.g. pointing game (Sec.~\ref{sec:pointing})), 
we assume 
that the classifier $C(\cdot)$ assigns the labels $I=\{\forall j:  C_j(x)>0\}$ to
the image $x$. 
For each $i\in I$, we are interested in the per label classifier response, and instead define the margin as: 
\begin{equation}
 M_i(x')=C_i(x') \label{eq:multiobject}
\end{equation}
Again, $M_i(x')\le 0$ if and only if $C(\cdot)$ does not assign label $i$
to image $x$.
In both cases, an adversarial perturbation $x'$ can be found by minimizing: 
\begin{equation}
    \left(M_i(x')-T\right)^2 \label{obj} 
\end{equation}
where $T$ is a target value smaller than zero. It is well-known~\cite{sorensen1982newton} that minimizing a loss of the form:
\begin{equation}
    \left(M_i(x')-T\right)^2 +\lambda ||x'-x||_2^2 \label{LSE}
\end{equation}
is equivalent to finding a minimizer of Eq.~\eqref{obj} that lies in the ball defined by \mbox{$||x-x'||_2^2 \leq \rho$} for some $\rho$. As such, minimizing this objective for an appropriate value of $\lambda$ and $T$ is a good strategy for finding adversarial perturbations of image $x$ with small $\ell_2$ norm.

Writing $C^{(l)}(x)$ for the classifier response of the $l^\text{th}$ layer of the neural net, we consider the related loss:  
\begin{equation}
    \left(M_i(x')-T\right)^2 +\lambda' \sum_{l\in \cal L}||C^{(l)}(x')-C^{(l)}(x)||_2^2 +\lambda ||x'-x||_2^2 \label{eq:multiclass}
\end{equation}
defined over a set of layers of the neural network $\cal L$. 

The second term is the perceptual loss of \cite{johnson2016perceptual}, and minimizing this objective is equivalent to finding a minimizer of Eq.~\eqref{LSE}  subject to the requirement that $x'$ lies in the ball defined by \mbox{$\sum_{l\in\cal L}||C^{(l)}(x')-C^{(l)}(x)||_2^2 \leq \rho'$} for some $\rho'$.

To convert the adversarial perturbation to a 
saliency map, we first calculate the size of the adversarial perturbation in each pixel by computing the average squared difference over the channels. 
Second, in order to highlight 
areas with large changes, we apply a Gaussian blur with parameter $\sigma$ to the differences to give our resultant saliency map.

We systematically evaluate the effect of altering the 
regularized layers
for 
a range of 
tasks. We find that the method is relatively stable and 
Eq.~\eqref{eq:multiclass} performs better than the unregularized 
Eq.~\eqref{LSE} for 
weak localization, insertion and deletion and the pointing game. As shown  in Fig.~\ref{diff_layers}, as more layers are regularized the perturbation becomes more localized.

\section{Perceptual Perturbations as Explanations}

\label{sec:visual-explanations}
Before describing our experimental overview, 
we give a qualitative analysis of the perceptual perturbations
(Fig.~\ref{fig:ppe}). The perturbations do a good job of localizing on a single object class, even in the
presence of highly textured images (dragonfly on fern), and in images with multiple classes (baseball and people). Some error in localization seems to arise from supporting classes adjacent to the object - 
e.g.
human legs behind the lawnmower are found to be salient.

Furthermore, a qualitative evaluation can be seen in Fig.~\ref{qualitative}.
These images were selected to be challenging -- we visualize a subset
of those images where unregularized adversarial perturbations did not align with the object. Compared to other visual explanation techniques, our method highlights the interior textures of the target object in the image. This differs from gradient-based methods which capture finer edge details such as SmoothGrad~\cite{smilkov2017smoothgrad} and to activation-based methods which highlight the entire object coarsely such as Grad-CAM~\cite{GradCAM2016}. This is perhaps clearest in the first image where we capture the interior texture of the socks rather than just its hard contours. 
\label{sec:experimental-results}

\begin{table}
\begin{tabular}{|c|ccc|c|}

\hline
Method & Value & Per. & Mean & Best\\
\hline\hline
  GuidedBP \cite{springenberg2014striving}      &  0.48              & 0.52             & 0.49              & 0.48\\
  Grad-CAM \cite{GradCAM2016}                    &  0.49              & 0.51             & 0.48              & 0.48\\
  Guided-Grad-CAM \cite{GradCAM2016}             &  \emph{0.46}       & 0.49             & \emph{0.45}              & 0.45 \\
  Excitation \cite{Zhang_2017}                  &  0.48              & 0.50             & \underline{0.44}  & \emph{0.44}\\
  SmoothG \cite{smilkov2017smoothgrad}          &  0.47              & 0.49             & 0.47              & 0.47\\
  IntegratedG \cite{sundararajan2017axiomatic}  &  \underline{0.44}  & 0.51             & 0.48              & \emph{0.44} \\
  Extremal \cite{fong2019understanding}         &  0.55              & 0.52             & 0.54              & 0.52\\
  RISE \cite{petsiuk2018rise}                   &  0.51              & 0.52             & 0.48              & 0.48\\
  NormGrad \cite{rebuffi2020revisiting}         & 0.49               & 0.52                & 0.46                 & 0.47 \\
  sNormGrad \cite{rebuffi2020revisiting}        & 0.49              & 0.52                & 0.47                 & 0.47 \\
\hline
  Us NoPer                                     & 0.50               &\emph{0.47}        & 0.46             & 0.46\\ 
  Us Unguided                                   & \underline{0.44}   &\underline{0.44}  & \bf{0.43}          & \underline{0.43}\\ 
  Us Guided                                     &\bf 0.41            &\bf 0.43          & \bf{0.43}         &\bf 0.41\\   
\hline
\end{tabular}
\caption{Results for Weak Object Localization (lower is better, see sec.\ref{sec:visual-explanations}). We have the lowest error for each thresholding strategy.\label{tab:localisation}
} \end{table}

\begin{figure*}[!t]
\rotatebox{90}{\hspace{0.3cm} Weak Localization}
\includegraphics[width=0.495\linewidth]{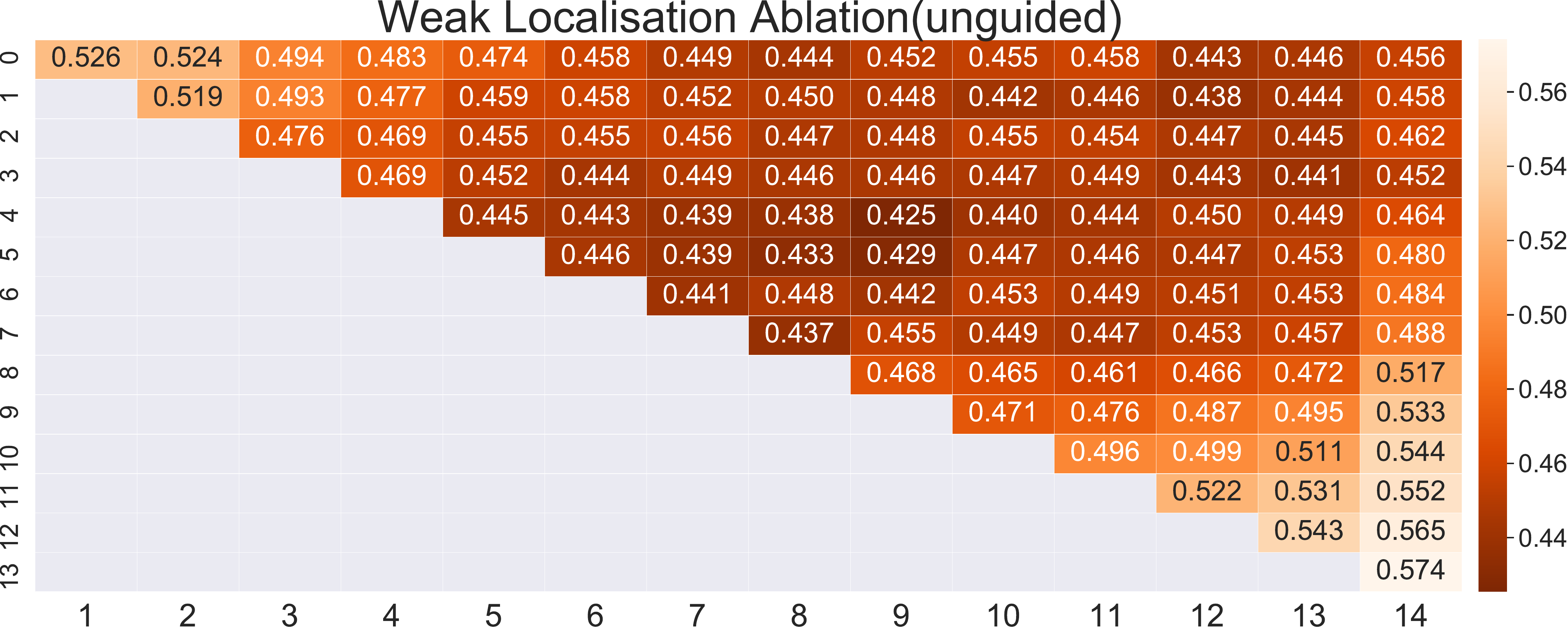}
\includegraphics[width=0.495\linewidth]{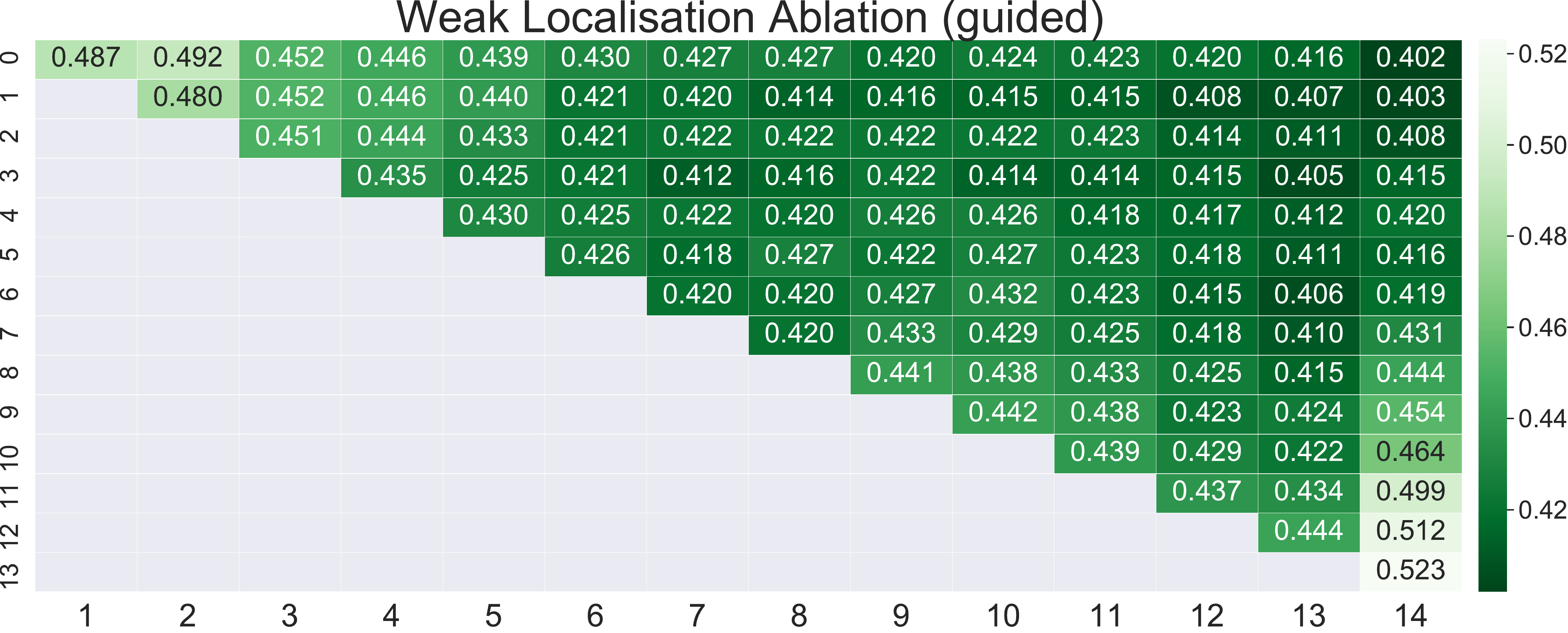}\\
\rotatebox{90}{\hspace{0.3cm} Insertion/Deletion}
\includegraphics[width=0.495\linewidth]{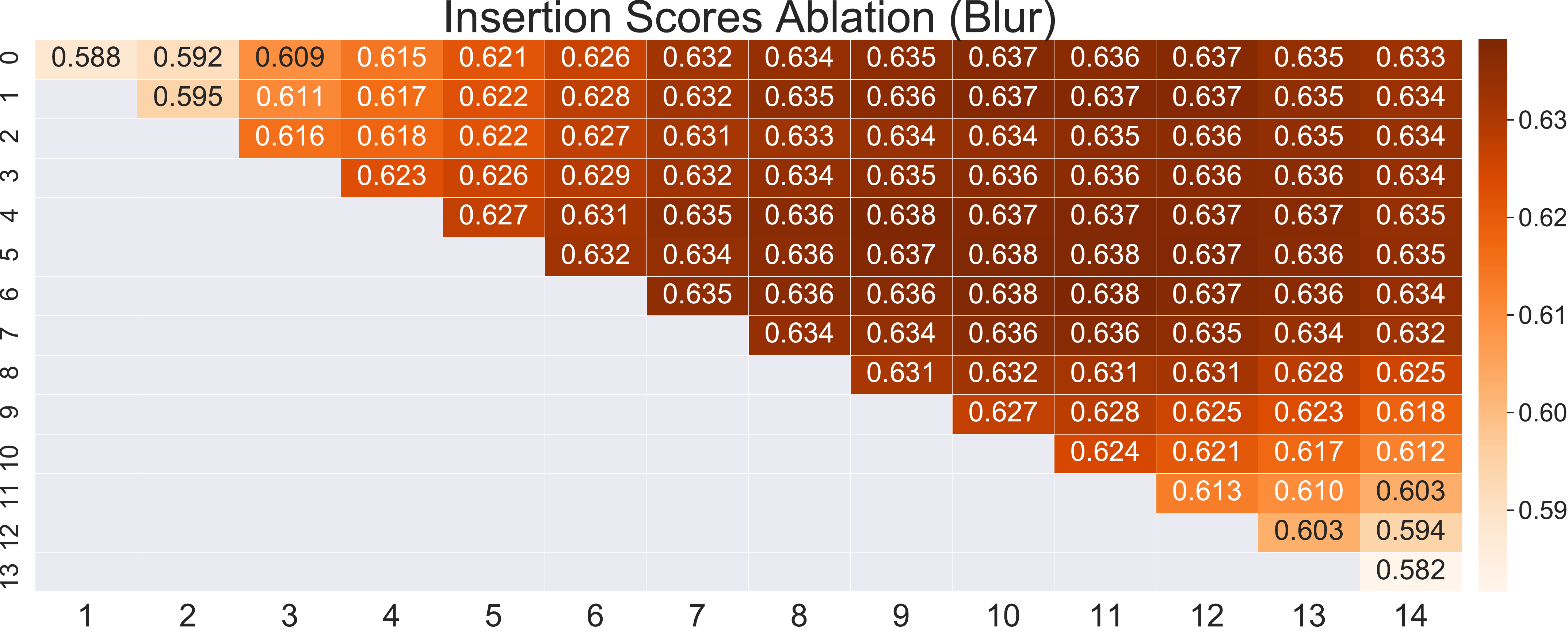}
\includegraphics[width=0.495\linewidth]{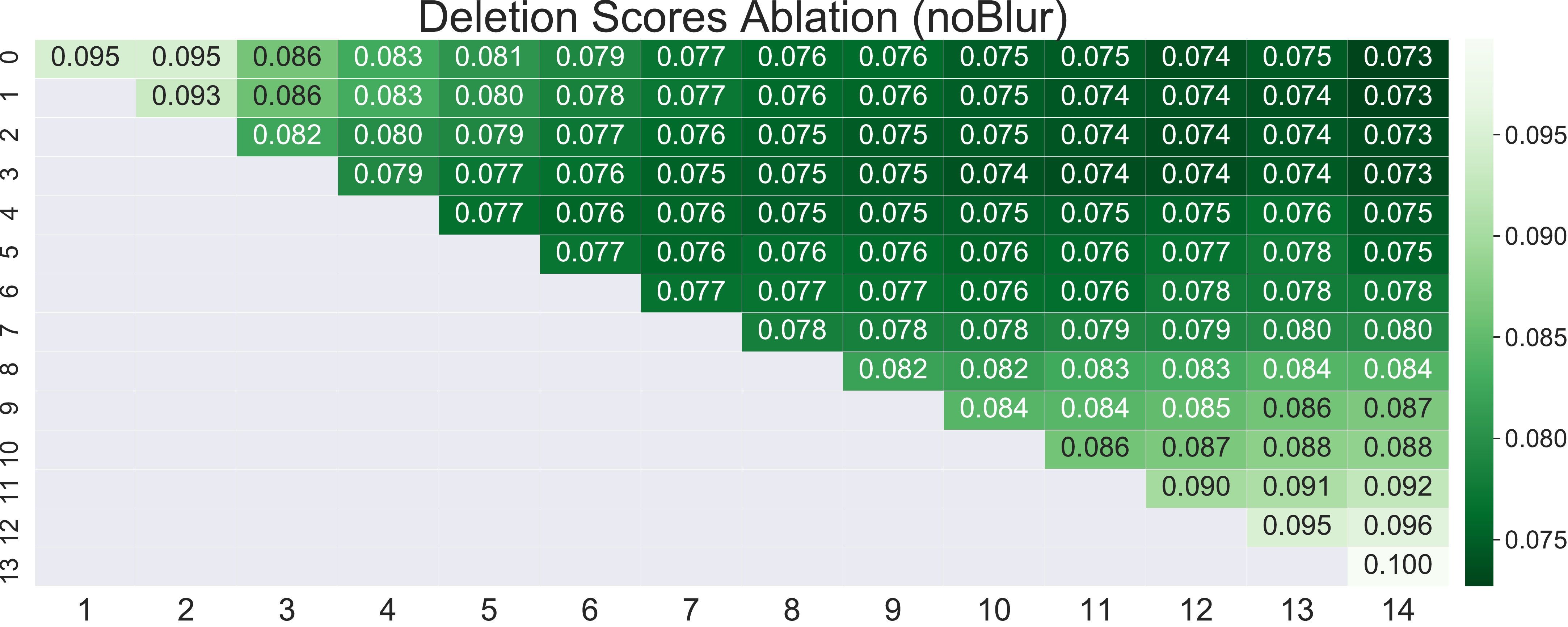}
\\
\rotatebox{90}{\hspace{0.6cm} Pointing Game}
\includegraphics[width=0.495\linewidth]{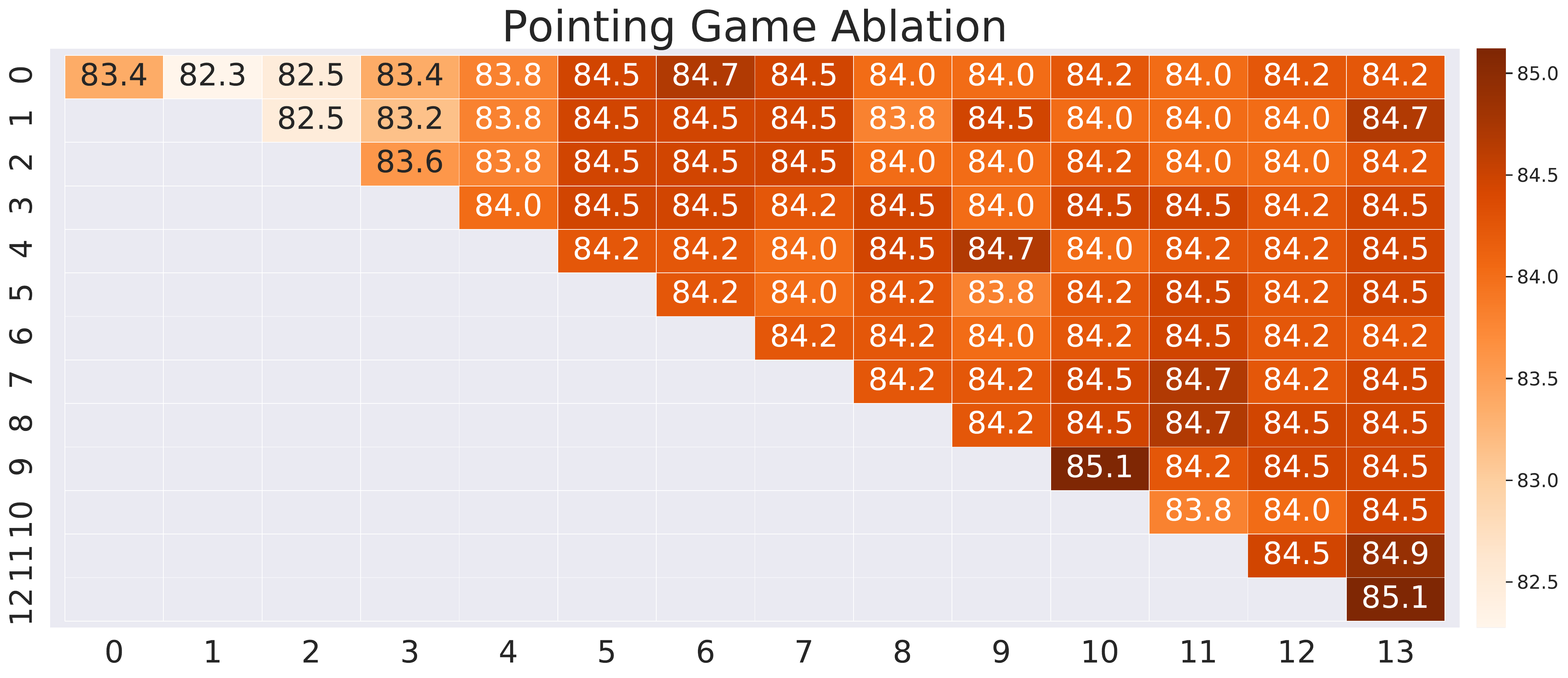}
\includegraphics[width=0.495\linewidth]{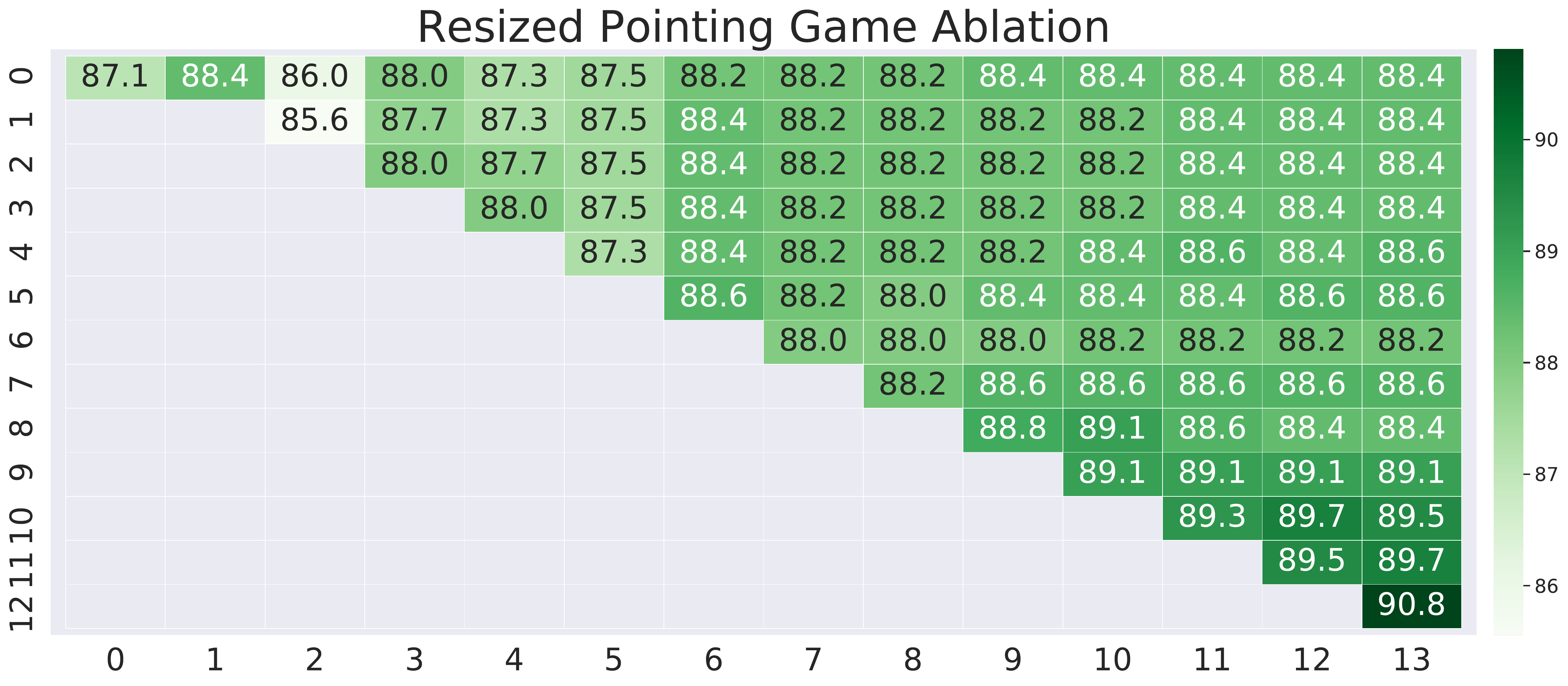}\\
\caption{
Sensitivity study on the choice of layers regularized for the three games
where the $i$th and $j$th entry displays the performance from regularizing layers $\{i,i+1,\dots,j-1\}$.
\emph{Top} shows the performance of weak localization for the unguided variant and the guided variant (Lower is better). The best performance is achieved when we regularize ReLU layers from 4 to 9 for the unguided variant and from 0 to 14 for the guided variant. \emph{Middle} shows the performance of deletion for the unBlur variant (lower is better) and insertion for the blur variant (higher is better). ReLU layers from 0 to 12 have been selected as it achieves the largest deviation between deletion score and insertion score. \emph{Lower} shows pointing game, with left panel representing orig. image size and right 
the resized image size. We regularize by each set of ReLU layers from  ReLU 0 to ReLU 12, with $i=j=0$ equivalent to the no perceptual approach, and report the average success rate (higher is better).
For ease of presentation in this game we present the best performance over $\sigma \in \{0,1,\dots 100\}$. 
\label{fig:heatmap}
}
\end{figure*}
\subsection{Sensitivity Studies}
For all experiments we performed extensive sensitivity studies to evaluate the importance of regularizing over different layers (Fig.~\ref{fig:heatmap}). Certain trends can be detected. Regularizing over most of the layers is effective for weak localization, insertion and deletion, and pointing games. However, the pointing game performs best with different layers, possibly as you only need to find a single point of an object, we find that regularizing only the top layer is optimal in our ablation 
study, even 
when 
testing
at multiple resolutions. 

\subsection{Weak Localization}
We evaluate perceptual perturbations as explanations using the weak localization protocol \cite{fong_2017},
and test our approach on the first 2000 ImageNet \cite{ILSVRC15} validation images.

We construct a set of bounding boxes for the largest connected region using three simple strategies based on: thresholding the raw values, thresholding a fixed percent of the image, and thresholding scaled by the image mean following \cite{fong_2017}. To match the previous thresholding strategies we normalize individual saliency maps to be in the range of $[0,1]$ before applying the blur.
For the first strategy, we use a value threshold where we grid search over the set of thresholds $\alpha$ where $0<\alpha\leq 1$ at intervals of size $0.05$. For the second strategy, we use a percentage threshold where we consider the $\alpha\%$ most salient pixels grid search over the same interval. For the third strategy, we use a threshold scaled by the per image mean where we grid search over the set of thresholds $\alpha$ where $0<\alpha\leq 10.50$ at intervals of size $0.05$.
We report the scores on all three strategies as well as the optimal strategy for each explanation method. For each threshold, we extract the largest connected component and draw a bounding box around it. The object is considered to be successfully localized when the Intersection over Union measure between this box and the ground truth is at least $0.5$. Following Grad-CAM's guided version \cite{GradCAM2016}, which makes use of image gradients, we consider a guided variant of our own method consisting of an element-wise multiplication between our perturbations and the normalized gradient of the $C_i(x)$ with respect to the image $x$.

We set $T=-2$, $\lambda'=10000, \lambda=1$ in Eq.~\eqref{eq:multiclass}, and we run an ablation study for the first 1000 images. We select two sequential sets of ReLU layers to regularize over in a VGG19bn network \cite{VGGnet} using the value-threshold strategy. 
For the un-guided variant of our method, we regularize from ReLU 4 to 9 
($0$ indexed) 
and for the guided variant we regularize from ReLU 0 to 14.
We report the results for each strategy using these layers. Qualitative evidence suggests that regularizing sequentially more and higher layers tends to improve object localization in the image (see Figs~\ref{diff_layers},\ref{fig:heatmap}).   

We compare our method, its guided 
variant and our method without 
the perceptual loss to other
  methods~\cite{fong2019understanding, petsiuk2018rise, rebuffi2020revisiting, GradCAM2016, smilkov2017smoothgrad, springenberg2014striving, sundararajan2017axiomatic, Zhang_2017}. For the methods in~\cite{GradCAM2016,smilkov2017smoothgrad,springenberg2014striving,sundararajan2017axiomatic}, we used the PyTorch CNN Visualizations repository~\cite{uozbulak_pytorch_vis_2019}.  
For~\cite{fong2019understanding,Zhang_2017}, we use the TorchRay implementation~\cite{TorchrayGithub} and for \cite{rebuffi2020revisiting}, we use its NormGrad branch~\cite{TorchrayNormGradGithub}.
For RISE~\cite{petsiuk2018rise} 
we used the authors code from~\cite{RISEGithub}. 
We used the provided parameters for all methods.

We varied $\sigma \in[1, 30]$ and found that results saturate at $\sigma=20$ for all three thresholds. We outperform all other methods  (see Table~\ref{tab:localisation}). Our guided variant obtains the lowest error for two thresholding strategies 
and joint best using mean-based thresholding. The non-guided version performs second-best in value thresholding, percent-based thresholding, and joint best in the mean-based thresholding strategy. When selecting the best possible threshold for each method, our method and it's unguided variant achieve the lowest and second lowest error rates respectively.

\subsection{Insertion/Deletion Game}

\begin{figure*}\centering
\begin{tabular}{c
@{\hspace{1mm}}c
@{\hspace{1mm}}c
@{\hspace{1mm}}c}
Orig. Image & Saliency  & \hspace{5mm} Deletion Curve & \hspace{4mm} Insertion Curve \\
\includegraphics[height=0.241\linewidth]{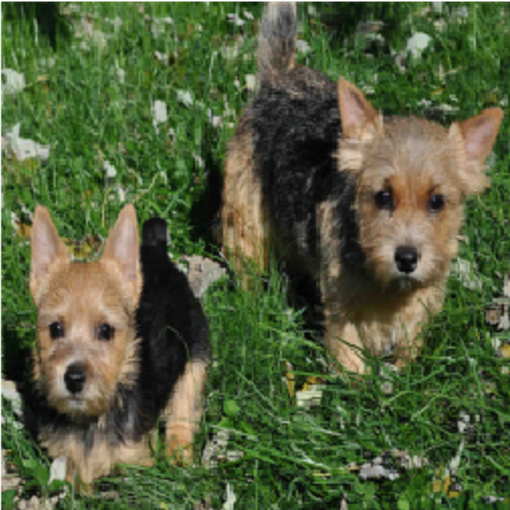}&
\includegraphics[height=0.241\linewidth]{figs/newExplanations/fig3/29_dim1_blur.pdf}&
\includegraphics[trim={0 0 0 0.5cm},clip,height=0.241\linewidth]{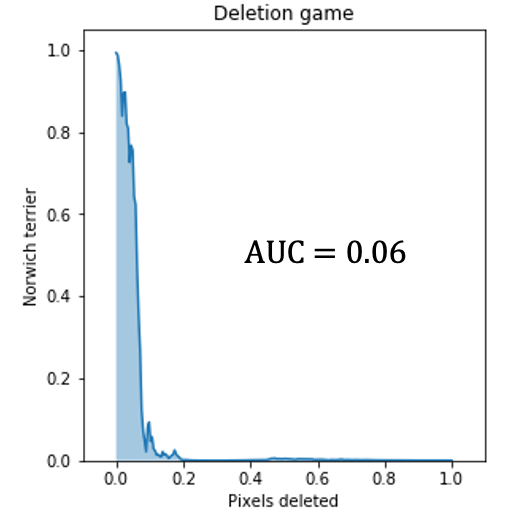} &
\includegraphics[trim={0 0 0 0.5cm},clip,height=0.241\linewidth]{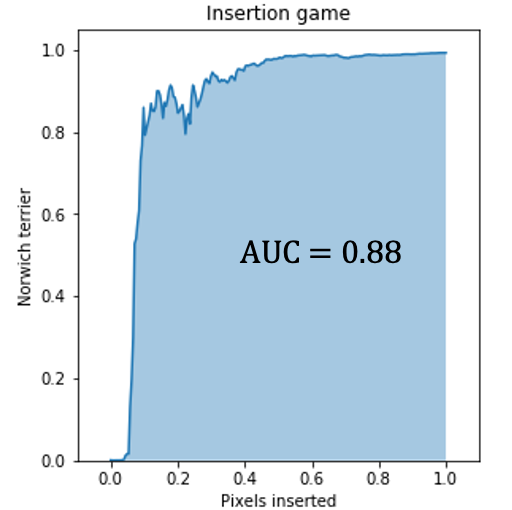}
\end{tabular}
\caption{
{Insertion (higher is better) and Deletion (lower is better) metrics \cite{petsiuk2018rise}. The average AUC over 5000 images in the ImageNet validation set as pixels are sequentially set to, or changed from, their value in the original image.}
\label{insertion}
}
\end{figure*}

We compute the insertion and deletion metrics from \cite{petsiuk2018rise}.
For the deletion metric, we construct the deletion response curve by sequentially changing the most salient pixels from their original value to mid-gray and measuring the classifier response. 
The deletion metric is defined as the AUC of the deletion curve, a smaller AUC score (i.e. a sharper drop in classifier response) is considered indicative of a better explanation.
The insertion metric is similar, however, rather than removing the pixels of largest saliency, it inserts the original values into a blurred version of the original image. The metric is again the AUC of the insertion curve, however a higher AUC score is considered better as this corresponds to a sharper increase in classifier response with the addition of the most salient pixels.

We set $T=-2$, $\lambda'=10000, \lambda=1$ in Eq.~\eqref{eq:multiclass}, and we run an ablation study for the first 500 images.
We identify the ReLU layers from 0 to 12 to regularize over in a VGG19bn network that achieve an optimal difference between having a high insertion score and a low deletion score (see Fig.~\ref{fig:heatmap}). 
We found that Gaussian blurring the saliency map improves the performance for the insertion metric, while it decreases the performance of the deletion metric. We varied the blur width between $1$ and $30$ and found the performance for insertion saturates when $\sigma \geq 4.0$. For consistency, we used the blur function in \cite{petsiuk2018rise}.

We compare our perceptual method and its blurred variants with a set of alternative methods for this experiment \cite{fong2019understanding, petsiuk2018rise, rebuffi2020revisiting, GradCAM2016, Simonyan2013, springenberg2014striving, Zeiler2014, Zhang_2017}.  We use the standard implementation of this game from the RISE repository \cite{petsiuk2018rise,RISEGithub}. We compare our method with RISE from the above package and each of the saliency methods implemented in the TorchRay package~\cite{TorchrayGithub} using the standard settings. For NormGrad we used its branch in the TorchRay Repository~\cite{TorchrayNormGradGithub}.
We include two additional baselines. The first baseline is our method without the perceptual loss and the second baseline is the $\ell_2$ 
norm between the image and a blurred version.
Our approach performs significantly better than our baseline and is noticeably better than all others for the deletion metric and joint third for the insertion metric (see Table \ref{tab:insertionDeletionGameRes}).

\begin{table}
\begin{tabular}{|c|cc|}
\hline
Method & Deletion Score & Insertion Score \\ \hline \hline
Gradient~\cite{Simonyan2013}               & 0.19             & 0.51              \\
Deconv~\cite{Zeiler2014}                   & 0.21             & 0.56              \\
GuidedBP~\cite{springenberg2014striving}   & 0.14             & 0.57              \\
Excitation~\cite{Zhang_2017}               & 0.12             & \emph{0.63}       \\
Grad-CAM~\cite{GradCAM2016}                &  0.11            & \underline{0.64}  \\ 
Extremal~\cite{fong2019understanding}      &  0.16            & 0.62              \\ 
RISE~\cite{petsiuk2018rise}                &  0.12            & \bf 0.65          \\ 
NormGrad~\cite{rebuffi2020revisiting}      & \underline{0.09} & 0.58              \\
sNormGrad~\cite{rebuffi2020revisiting}     & \emph{0.10}      & 0.59              \\ \hline  
blurDiff ($\sigma=2.5$)				        &  0.14            & 0.59             \\  
Us NoPer ($\sigma=0.0$)					   &  \emph{0.10}     & 0.42              \\   
Us NoPer ($\sigma=2.5$)					   &  0.15            & 0.54              \\ \hline  
Us ($\sigma=0.0$)                                       & \bf 0.07         & 0.54              \\ 
Us ($\sigma=1.0$)                      & \underline{0.09} & 0.61              \\ 
Us ($\sigma=2.5$)                      &  0.11            & 0.62              \\ 
Us ($\sigma=5.0$)                      &  0.12            & \emph{0.63}       \\ \hline
\end{tabular}
\caption{Results for deletion (first column) and insertion game (second column).  The result shows our method performs better than other methods for the deletion metric (without blur) and is comparable to other methods for the insertion metric (with blur).
\label{tab:insertionDeletionGameRes}}
\end{table}

\subsection{Pointing Game}
\label{sec:pointing}
Additionally, we evaluated on the pointing game introduced by~\cite{Zhang_2017} on the VOC dataset. 
In this game 
the task is to select a single point in the image which is in the object in question~\cite{Zhang_2017}. The game is subdivided into 2 subtasks, a standard set of images, and a more difficult subset 
in which the object occupies less than 25\% of the image, and must also contain a distracting class (see~\cite{Zhang_2017}). 
We use the TorchRay~\cite{TorchrayGithub} implementation of the game and comparison methods 
and ~\cite{TorchrayNormGradGithub} for the implementation of NormGrad. 

The pointing game uses a modified VGG16 classifier that assigns multiple classes to a single region of the image, and thus we use Eq.~\eqref{eq:multiclass} 
with $T=-10$, $\lambda' = 1000$, $\lambda = 1$ and the formulation from Eq.~\eqref{eq:multiobject}. 
In this game, the images vary in size, and the classifier returns a detector like response 
for a set of overlapping regions of the image. 
Given a choice of class, we suppress all candidate regions by ensuring that the maximal response of any region is close to the target value.

\begin{table}
\begin{center}
\begin{tabular}{|c|cc|}
\hline
Method & Orig. Image & Scaled Image \\ \hline \hline
Center                                   &  69.6 (42.4)               &  69.6 (42.4)  \\ \hline
Gradient~\cite{Simonyan2013}             &  76.3 (56.9)               &  84.6 (70.0)  \\
Deconv~\cite{Zeiler2014}                 &  67.5 (44.2)               &  75.0 (53.5)  \\ 
GuidedBP~\cite{springenberg2014striving} &  75.9 (53.0)               &  83.7 (67.1)  \\ 
Excitation~\cite{Zhang_2017}             &  77.1 (56.6)               &  84.0 (67.5)  \\ 
Grad-CAM~\cite{GradCAM2016}              &  \emph{86.6 (74.0)}        &  \underline{89.1 (77.7)}  \\ 
Extremal~\cite{fong2019understanding}    &  \textbf{88.0 (77.3)}      &  86.4 (71.0)  \\  
RISE~\cite{petsiuk2018rise}              &  \underline{86.7 (75.4)}   &  NA (NA)       \\ 
NormGrad~\cite{rebuffi2020revisiting}    &  81.9 (64.9)               &  \emph{88.6} (75.6)  \\ 
sNormGrad~\cite{rebuffi2020revisiting}   &  86.0 (72.7)               &  \textbf{90.1 (80.8)}  \\ \hline
Us NoPer                                 &  81.2 (62.9)               &  86.0 (71.6)  \\ 
Us                                       &  85.1 (69.0)               &  88.2 (\emph{76.5})  \\ \hline 
\end{tabular}
\end{center}
\caption{Results for the Pointing Game. We present two sets of results, the
performance on the images in the original size (first column),
and second the performance on images which are scaled to 1.5x using bilinear
scaling (second column). For each
result, the first number shows the percentage on the standard set,
with the number in brackets the performance on the difficult set. 
\label{tab:pointingGameRes}}
\end{table}

\begin{figure*}\begingroup
\setlength{\tabcolsep}{0pt} \renewcommand{\arraystretch}{0.0} \begin{tabular}{@{}c@{}c@{}c@{}c@{}c@{}c@{}c@{}c@{}c@{}c@{}}
Orig. & None & FC3 & FC2 & FC1 & 
Conv5\_4
& Conv4\_4
& Conv3\_4
& Conv2\_2
& Conv1\_1
\\
\includegraphics[width=0.1\linewidth]{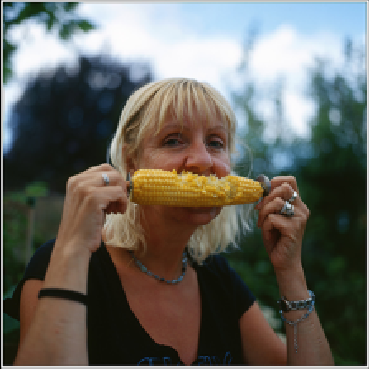}
& \includegraphics[width=0.1\linewidth]{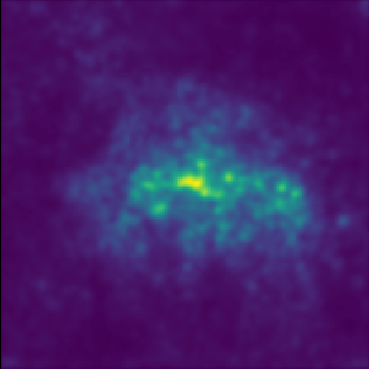}
& \includegraphics[width=0.1\linewidth]{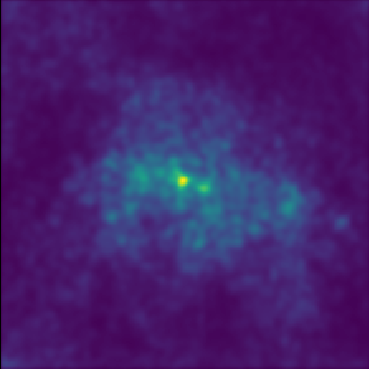}
& \includegraphics[width=0.1\linewidth]{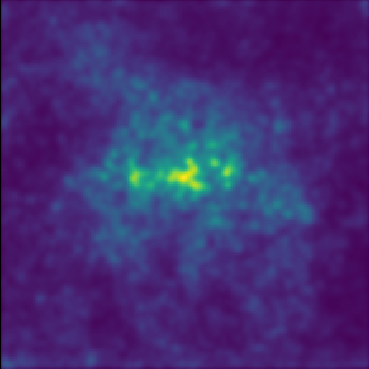}
& \includegraphics[width=0.1\linewidth]{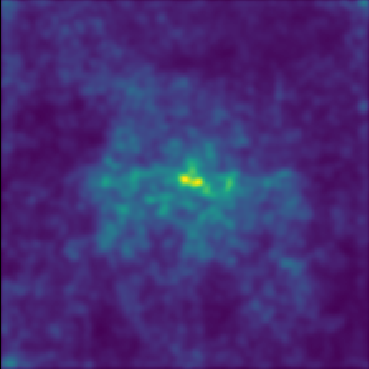}
& \includegraphics[width=0.1\linewidth]{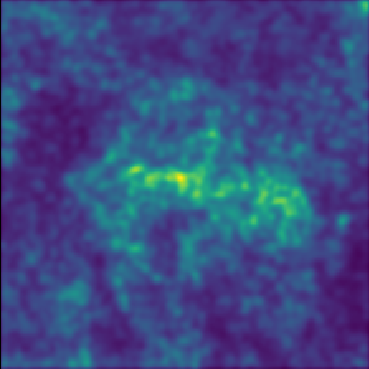}
& \includegraphics[width=0.1\linewidth]{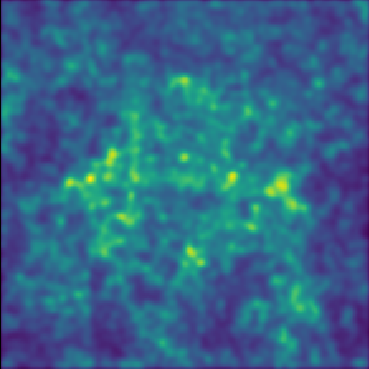}
& \includegraphics[width=0.1\linewidth]{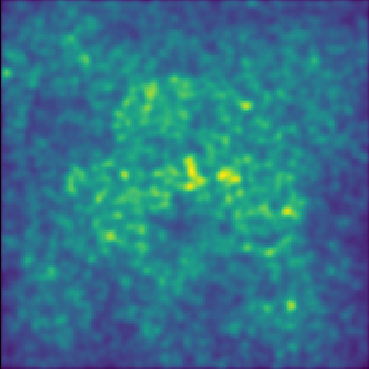}
& \includegraphics[width=0.1\linewidth]{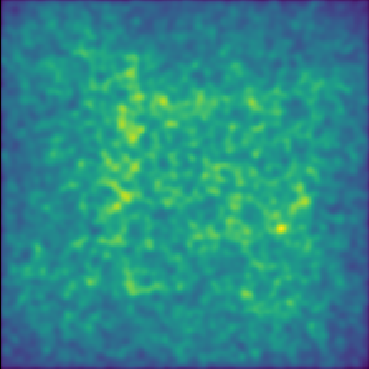}
& \includegraphics[width=0.1\linewidth]{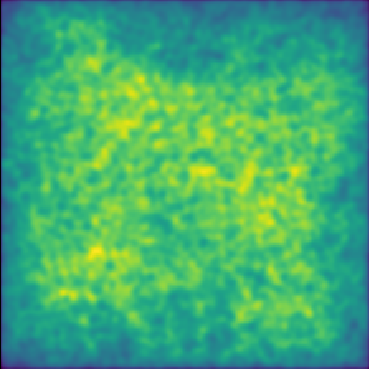}
\\
  \includegraphics[width=0.1\linewidth]{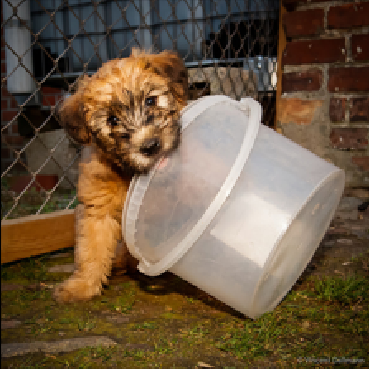}
& \includegraphics[width=0.1\linewidth]{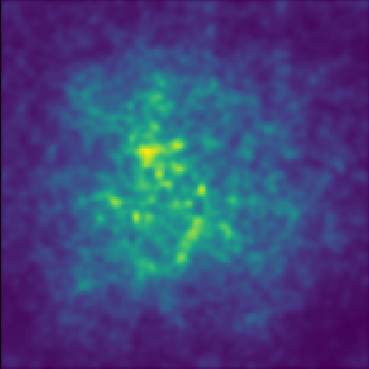}
& \includegraphics[width=0.1\linewidth]{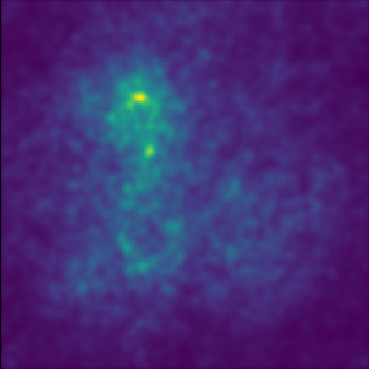}
& \includegraphics[width=0.1\linewidth]{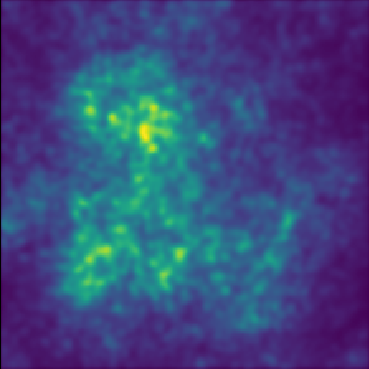}
& \includegraphics[width=0.1\linewidth]{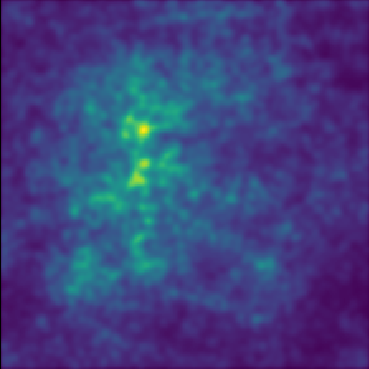}
& \includegraphics[width=0.1\linewidth]{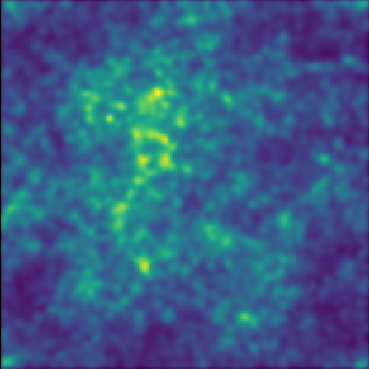}
& \includegraphics[width=0.1\linewidth]{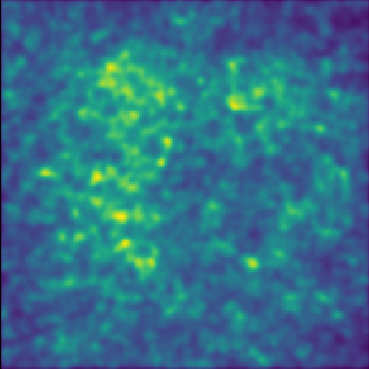}
& \includegraphics[width=0.1\linewidth]{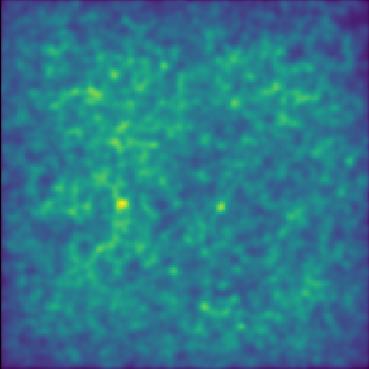}
& \includegraphics[width=0.1\linewidth]{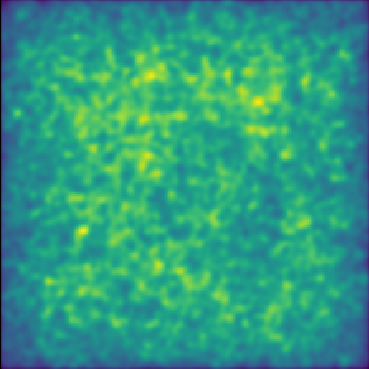}
& \includegraphics[width=0.1\linewidth]{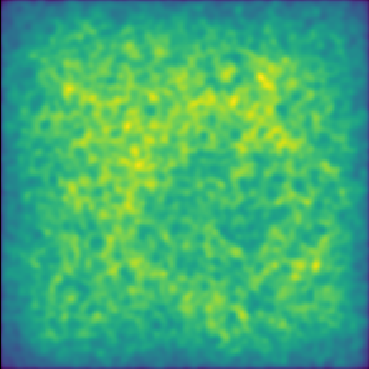}
\\
  \includegraphics[width=0.1\linewidth]{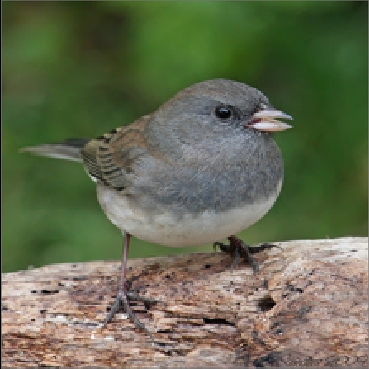}
& \includegraphics[width=0.1\linewidth]{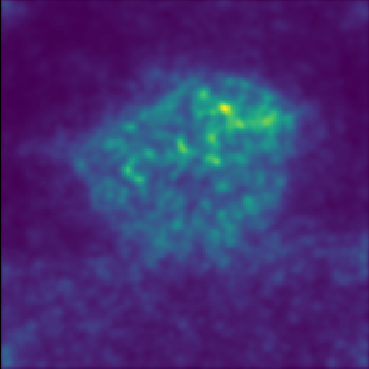}
& \includegraphics[width=0.1\linewidth]{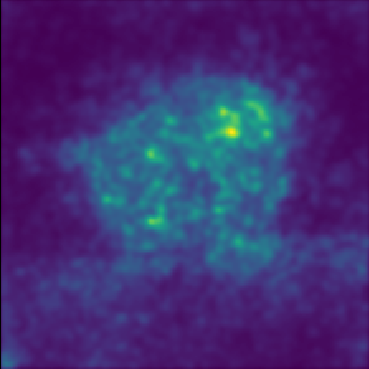}
& \includegraphics[width=0.1\linewidth]{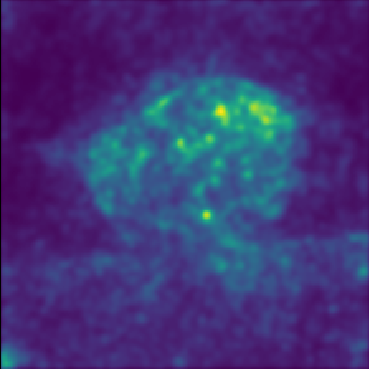}
& \includegraphics[width=0.1\linewidth]{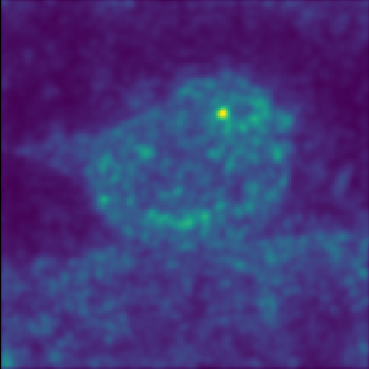}
& \includegraphics[width=0.1\linewidth]{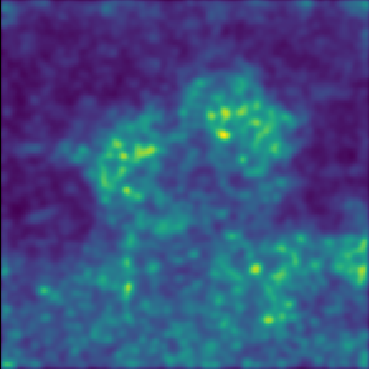}
& \includegraphics[width=0.1\linewidth]{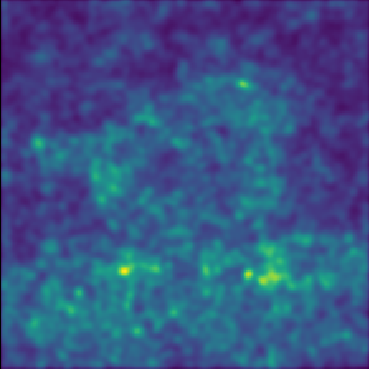}
& \includegraphics[width=0.1\linewidth]{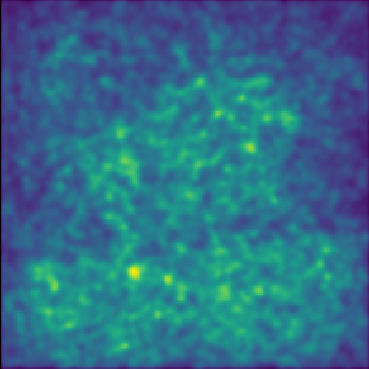}
& \includegraphics[width=0.1\linewidth]{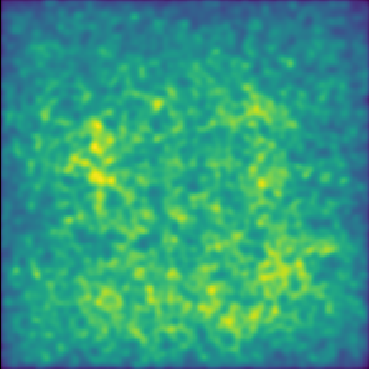}
& \includegraphics[width=0.1\linewidth]{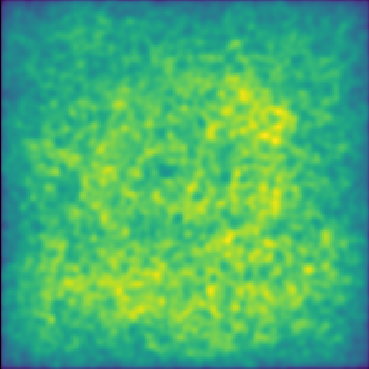}
\end{tabular}
\endgroup
\caption{Sanity checks plots. These plots demonstrate how our saliency maps change when we randomise the underlying classifier. }
\label{fig:sanityChecks}
\end{figure*}

As with the previous games, we select the layers and blur using an ablation study. 
We test each range of contiguous layers from the first ReLU layer, 
to final ReLU layer, and we vary the blur between $0$ and $100$\footnote{For 
efficiency, we create one perturbation per set of layers and vary 
$\sigma$}. We perform the study on the first 500 bounding boxes, which corresponds to 
457 standard bounding boxes of which 160 are in the difficult subset 
and $43$ boxes excluded by the benchmark.

Results can be seen in Fig.~\ref{fig:heatmap} lower left.
Due to the reduced dataset ($457$),
we report the success rate over the standard set, rather than the average class success reported on the full dataset. 
For compactness, we display the best result over the $\sigma$s, and 
to measure consistency 
we present the percentage of $\sigma$s which yield a performance above $82\%$ (Appendix B.2). Two layer sets achieve the largest 
score, ReLU $9$ (layer $22$) 
with $\sigma \in \{47,48\}$
or ReLU $12$, with $\sigma \in \{41,42\}$. 
We 
select ReLU $9$ with $\sigma=48$. The first layer is less sensitive to the choice of blur with $\approx 70\%$ giving 
a result above $82\%$ in contrast to the $\approx 44.6\%$ for the final layer 
(this setting also gives a lower performance overall with a score of $83.8\%$ 
($66.4\%$)).

The full results, conducted on all images/tests are presented in the 
left column of Table~\ref{tab:pointingGameRes}. 
We are competitive with the best methods, with a $2.9\%$ performance 
difference with the best method in the standard setting. 
In Appendix B, we present a study with a limited set of $\sigma$s where we achieve qualitatively similar results, but only a $0.3\%$ drop in standard performance and no drop in the difficult setting. 
We additionally compare to our framework with the same parameters but without the perceptual loss (with the blur which maximizes the ablation score - see Appendix B.1).
We refer to as this `no perceptual'. 
As we see in Table~\ref{tab:pointingGameRes} our version with the perceptual loss outperforms it. 
We also display this approach in the ablation study 
(top left corner). 
The perceptual loss outperforms in almost all cases,
with exceptions in the low layers. Further, 
the results with the perceptual are more consistently above 82\% (Appendix B).

Finally, we also propose an additional pointing game experiment. As this game is defined on non-standard sized image, we consider the case where we increase the size of the images via a simple resizing, and test the performance on this set. This helps performance as the fully convolutions network then gives predictions for more areas that are then maximized over. 
We implement this as a pre-and post- processing step, where we resize the image, construct the saliency map, and resize the saliency map to the correct size. 
We perform a similar ablation study for this case, (Fig.~\ref{fig:heatmap} bottom right panel). The optimal layer in this case is ReLU $12$ with $\sigma \in \{28.0,30.0\}$, (selected 30). 

The results can be seen in the right column of Table~\ref{tab:pointingGameRes}, where most methods work on this new dataset (the TorchRay RISE implementation requires a perfectly sized image and is NA).
The performance of many of the methods is increased in this new approach, with many weaker methods substantially increasing (except for center - a baseline which chooses the center point). Further, the performance characteristics of the best performing methods also improves with resizing, albeit by lower margins, indicating that this resizing generally helps almost all methods.The best performing method is Selective NormGrad with a score of $90.1\%$, $2.1\%$ above the previously highest score. Our Perceptual method is competitive with the best performing methods, achieving a score of $88.2\%$.

\subsection{Sanity Checks}
Finally, we apply the Sanity Checks   
proposed in \cite{adebayo2018sanity} to our method. 
We randomize the weights on the final $k$ layers of the 
network, and observe how the saliency of our approach varies visually. 
We perform this experiment on VGG19bn, as most of the experiments in this 
paper were performed on this network, 
and we match the remaining parameters/layers to the 
insertion deletion study.  
When randomizing the layers we set 
the parameters to the value that would have been in an untrained network. 
We select the layers in the VGG19bn architecture that correspond to the  
VGG16 layers used by~\cite{rebuffi2020revisiting},  namely each 
fully connected layer and the final convolution in each set with the 
exception of the first set for which we use the first convolution. 

Fig.~\ref{fig:sanityChecks} shows the results on three common sanity check images.
As the layers are progressively randomized, salie\-ncy spreads from the relevant objects towards other objects and highly textured regions in the image, and is  mostly non-existent after the second convolution qualitatively matching the behavior 
in Selective NormGrad seen in \cite{rebuffi2020revisiting}.

\section{Conclusion}
\label{sec:conclusion}
We explored a novel regularization for 
 adversarial perturbations based on the perceptual loss. This regularization is designed to
block the exploitation of exploding gradients when generating adversarial
perturbations forcing larger and more meaningful perturbations to be generated.
The fact that they remain imperceptible to humans is another piece of the puzzle in understanding
the interrelationship between adversarial perturbations, neural networks, and
human vision.
We believe that the imperceptible nature of our adversarial perturbations is due to both explanations discussed in the introduction for the existence of adversarial perturbations being  partially correct. Even when regularizing over the layers of our network and preventing adversarial perturbations from exploiting exploding gradients and some of the inherent instability in deep networks, Szegedy et al.'s~\cite{szegedy2013intriguing} argument still holds and one can obtain a new class label by slightly altering a large number of pixels.

We have shown how these perturbations can be interpreted as explanations and
obtained state-of-the-art results on several standard explainability
benchmarks.

{\bf Acknowledgments}
 This work was supported by the Omidya Group and The Alan Turing Institute under the UK Engineering and Physical Sciences Research Council (EPSRC) grant no. EP/N510129/1
 and Accenture Plc. Moreover, we acknowledge Pearl for computing resources and in particular the help of Tomas Lazauskas and Suleman Tariq.

{\small
\bibliographystyle{ieee_fullname}
\bibliography{egbibWorkshop2}

\begin{thebibliography}{10}\itemsep=-1pt

\bibitem{adebayo2018sanity}
Julius Adebayo, Justin Gilmer, Michael Muelly, Ian Goodfellow, Moritz Hardt,
  and Been Kim.
\newblock Sanity checks for saliency maps.
\newblock In S. Bengio, H. Wallach, H. Larochelle, K. Grauman, N. Cesa-Bianchi,
  and R. Garnett, editors, {\em Advances in Neural Information Processing
  Systems}, volume~31, pages 9505--9515. Curran Associates, Inc., 2018.

\bibitem{athalye2017synthesizing}
Anish Athalye, Logan Engstrom, Andrew Ilyas, and Kevin Kwok.
\newblock Synthesizing robust adversarial examples.
\newblock volume~80 of {\em Proceedings of Machine Learning Research}, pages
  284--293, Stockholmsmässan, Stockholm Sweden, 10--15 Jul 2018. PMLR.

\bibitem{carlini2017towards}
Nicholas Carlini and David Wagner.
\newblock Towards evaluating the robustness of neural networks.
\newblock In {\em 2017 IEEE Symposium on Security and Privacy (SP)}, pages
  39--57. IEEE, 2017.

\bibitem{chang2018explaining}
Chun-Hao Chang, Elliot Creager, Anna Goldenberg, and David Duvenaud.
\newblock Explaining image classifiers by counterfactual generation.
\newblock In {\em ICLR}, 2019.

\bibitem{Chattopadhay_2018}
Aditya Chattopadhay, Anirban Sarkar, Prantik Howlader, and Vineeth~N
  Balasubramanian.
\newblock Grad-cam++: Generalized gradient-based visual explanations for deep
  convolutional networks.
\newblock {\em 2018 IEEE Winter Conference on Applications of Computer Vision
  (WACV)}, Mar 2018.

\bibitem{chen2018ead}
Pin-Yu Chen, Yash Sharma, Huan Zhang, Jinfeng Yi, and Cho-Jui Hsieh.
\newblock Ead: elastic-net attacks to deep neural networks via adversarial
  examples.
\newblock In {\em Thirty-second AAAI conference on artificial intelligence},
  2018.

\bibitem{dhurandhar2018explanations}
Amit Dhurandhar, Pin-Yu Chen, Ronny Luss, Chun-Chen Tu, Paishun Ting,
  Karthikeyan Shanmugam, and Payel Das.
\newblock Explanations based on the missing: Towards contrastive explanations
  with pertinent negatives.
\newblock In {\em NeurIPS}, pages 592--603, 2018.

\bibitem{elsayed2018adversarial}
Gamaleldin Elsayed, Shreya Shankar, Brian Cheung, Nicolas Papernot, Alexey
  Kurakin, Ian Goodfellow, and Jascha Sohl-Dickstein.
\newblock Adversarial examples that fool both computer vision and time-limited
  humans.
\newblock In {\em Advances in Neural Information Processing Systems}, pages
  3910--3920, 2018.

\bibitem{TorchrayGithub}
Ruth Fong, Mandela Patrick, and Andrea Vedaldi.
\newblock Torchray.
\newblock
  {\footnotesize\UrlFont{https://github.com/facebookresearch/TorchRay}}, 2019.

\bibitem{fong2019understanding}
Ruth Fong, Mandela Patrick, and Andrea Vedaldi.
\newblock Understanding deep networks via extremal perturbations and smooth
  masks.
\newblock In {\em Proceedings of the IEEE/CVF International Conference on
  Computer Vision (ICCV)}, October 2019.

\bibitem{TorchrayNormGradGithub}
Ruth Fong, Sylvestre-Alvise Rebuffi, Xu Ji, and Andrea Vedaldi.
\newblock Torchray normgrad.
\newblock {\footnotesize\UrlFont{github.com/ruthcfong/TorchRay/tree/normgrad}},
  2019.

\bibitem{fong_2017}
Ruth~C. Fong and Andrea Vedaldi.
\newblock Interpretable explanations of black boxes by meaningful perturbation.
\newblock {\em ICCV}, Oct 2017.

\bibitem{gardner2015deep}
Jacob~R Gardner, Paul Upchurch, Matt~J Kusner, Yixuan Li, Kilian~Q Weinberger,
  Kavita Bala, and John~E Hopcroft.
\newblock Deep manifold traversal: Changing labels with convolutional features.
\newblock {\em arXiv:1511.06421}, 2015.

\bibitem{geirhos2019imagenettrained}
Robert Geirhos, Patricia Rubisch, Claudio Michaelis, Matthias Bethge, Felix~A.
  Wichmann, and Wieland Brendel.
\newblock Imagenet-trained cnns are biased towards texture; increasing shape
  bias improves accuracy and robustness, 2019.

\bibitem{gilmer2018adversarial}
Justin Gilmer, Luke Metz, Fartash Faghri, Samuel~S Schoenholz, Maithra Raghu,
  Martin Wattenberg, and Ian Goodfellow.
\newblock Adversarial spheres.
\newblock {\em arXiv:1801.02774}, 2018.

\bibitem{goodfellow2014explaining}
Ian Goodfellow, Jonathon Shlens, and Christian Szegedy.
\newblock Explaining and harnessing adversarial examples.
\newblock In {\em International Conference on Learning Representations}, 2015.

\bibitem{goyal2019counterfactual}
Yash Goyal, Ziyan Wu, Jan Ernst, Dhruv Batra, Devi Parikh, and Stefan Lee.
\newblock Counterfactual visual explanations.
\newblock volume~97 of {\em Proceedings of Machine Learning Research}, pages
  2376--2384, Long Beach, California, USA, 09--15 Jun 2019. PMLR.

\bibitem{hendricks2018generating}
Lisa~Anne Hendricks, Ronghang Hu, Trevor Darrell, and Zeynep Akata.
\newblock Generating counterfactual explanations with natural language.
\newblock {\em Proceedings of the 2018 ICML Workshop on Human Interpretability
  in Machine Learning}, pages 95--98, 2018.

\bibitem{ioffe2015batch}
Sergey Ioffe and Christian Szegedy.
\newblock Batch normalization: Accelerating deep network training by reducing
  internal covariate shift.
\newblock volume~37 of {\em Proceedings of Machine Learning Research}, pages
  448--456, Lille, France, 07--09 Jul 2015. PMLR.

\bibitem{johnson2016perceptual}
Justin Johnson, Alexandre Alahi, and Li Fei-Fei.
\newblock Perceptual losses for real-time style transfer and super-resolution.
\newblock In {\em ECCV}, pages 694--711. Springer, 2016.

\bibitem{kaur2019perceptuallyaligned}
Simran Kaur, Jeremy Cohen, and Zachary~C. Lipton.
\newblock Are perceptually-aligned gradients a general property of robust
  classifiers?
\newblock {\em Science Meets Engineering of Deep Learning" Workshop at NeurIPS
  2019}, 2019.

\bibitem{leeNonLinear2003}
Ann~B. Lee, Kim~S. Pedersen, and David Mumford.
\newblock The nonlinear statistics of high-contrast patches in natural images.
\newblock {\em International Journal of Computer Vision}, 54(1):83--103, 2003.

\bibitem{lewis2013counterfactuals}
David Lewis.
\newblock {\em Counterfactuals}.
\newblock John Wiley \& Sons, 1973.

\bibitem{liu2016delving}
Yanpei Liu, Xinyun Chen, Chang Liu, and Dawn Song.
\newblock Delving into transferable adversarial examples and black-box attacks.
\newblock {\em ICLR}, 2017.

\bibitem{modas2019sparsefool}
Apostolos Modas, Seyed-Mohsen Moosavi-Dezfooli, and Pascal Frossard.
\newblock Sparsefool: a few pixels make a big difference.
\newblock In {\em CVPR}, pages 9087--9096, 2019.

\bibitem{Moosavi_Dezfooli_2016}
Seyed-Mohsen Moosavi-Dezfooli, Alhussein Fawzi, and Pascal Frossard.
\newblock Deepfool: A simple and accurate method to fool deep neural networks.
\newblock {\em CVPR}, Jun 2016.

\bibitem{hiddenSpace}
A. {Mustafa}, S. {Khan}, M. {Hayat}, R. {Goecke}, J. {Shen}, and L. {Shao}.
\newblock Adversarial defense by restricting the hidden space of deep neural
  networks.
\newblock In {\em 2019 IEEE/CVF International Conference on Computer Vision
  (ICCV)}, pages 3384--3393, 2019.

\bibitem{uozbulak_pytorch_vis_2019}
Utku Ozbulak.
\newblock Pytorch cnn visualizations.
\newblock
  {\scriptsize\UrlFont{https://github.com/utkuozbulak/pytorch-cnn-visualizations}},
  2019.

\bibitem{papernot2016distillation}
Nicolas Papernot, Patrick McDaniel, Xi Wu, Somesh Jha, and Ananthram Swami.
\newblock Distillation as a defense to adversarial perturbations against deep
  neural networks.
\newblock In {\em 2016 IEEE Symposium on Security and Privacy (SP)}, pages
  582--597. IEEE, 2016.

\bibitem{pascanu2012understanding}
Razvan Pascanu, Tomas Mikolov, and Yoshua Bengio.
\newblock Understanding the exploding gradient problem.
\newblock {\em CoRR, abs/1211.5063}, 2, 2012.

\bibitem{pearl2000causality}
Judea Pearl.
\newblock {\em Causality: models, reasoning and inference}, volume~29.
\newblock Springer, 2000.

\bibitem{RISEGithub}
Vitali Petsiuk.
\newblock Rise: Randomized input sampling for explanation of black-box models.
\newblock {\footnotesize\UrlFont{https://github.com/eclique/RISE}}, 2018.

\bibitem{petsiuk2018rise}
Vitali Petsiuk, Abir Das, and Kate Saenko.
\newblock Rise: Randomized input sampling for explanation of black-box models.
\newblock In {\em Proceedings of the British Machine Vision Conference (BMVC)},
  2018.

\bibitem{rebuffi2020revisiting}
Sylvestre-Alvise Rebuffi, Ruth Fong, Xu Ji, and Andrea Vedaldi.
\newblock There and back again: Revisiting backpropagation saliency methods.
\newblock In {\em Proceedings of the IEEE/CVF Conference on Computer Vision and
  Pattern Recognition (CVPR)}, June 2020.

\bibitem{Ribiero2016}
M Ribeiro, S Singh, and C Guestrin.
\newblock “why should i trust you?”explaining the predictions of any
  classifier.
\newblock {\em SigKDD}, 2016.

\bibitem{roth2019odds}
Kevin Roth, Yannic Kilcher, and Thomas Hofmann.
\newblock The odds are odd: A statistical test for detecting adversarial
  examples.
\newblock In {\em ICML}, pages 5498--5507, 2019.

\bibitem{ILSVRC15}
Olga Russakovsky, Jia Deng, Hao Su, Jonathan Krause, Sanjeev Satheesh, Sean Ma,
  Zhiheng Huang, Andrej Karpathy, Aditya Khosla, Michael Bernstein,
  Alexander~C. Berg, and Li Fei-Fei.
\newblock {ImageNet Large Scale Visual Recognition Challenge}.
\newblock {\em International Journal of Computer Vision (IJCV)},
  115(3):211--252, 2015.

\bibitem{GradCAM2016}
Ramprasaath~R. Selvaraju, Michael Cogswell, Abhishek Das, Ramakrishna Vedantam,
  Devi Parikh, and Dhruv Batra.
\newblock Grad-cam: Visual explanations from deep networks via gradient-based
  localization.
\newblock In {\em Proceedings of the IEEE International Conference on Computer
  Vision (ICCV)}, Oct 2017.

\bibitem{simon2018adversarial}
Carl-Johann Simon-Gabriel, Yann Ollivier, Leon Bottou, Bernhard Sch{\"o}lkopf,
  and David Lopez-Paz.
\newblock First-order adversarial vulnerability of neural networks and input
  dimension.
\newblock volume~97 of {\em Proceedings of Machine Learning Research}, pages
  5809--5817, Long Beach, California, USA, 09--15 Jun 2019. PMLR.

\bibitem{Simonyan2013}
K Simonyan, A Vedaldi, and A Zisserman.
\newblock Deep inside convolutional networks.
\newblock {\em ICLR}, 2013.

\bibitem{VGGnet}
Karen Simonyan and Andrew Zisserman.
\newblock Very deep convolutional networks for large-scale image recognition.
\newblock In {\em International Conference on Learning Representations}, 2015.

\bibitem{smilkov2017smoothgrad}
Daniel Smilkov, Nikhil Thorat, Been Kim, Fernanda Viégas, and Martin
  Wattenberg.
\newblock Smoothgrad: removing noise by adding noise.
\newblock {\em arXiv:1706.03825}, 2017.

\bibitem{song2017pixeldefend}
Yang Song, Taesup Kim, Sebastian Nowozin, Stefano Ermon, and Nate Kushman.
\newblock Pixeldefend: Leveraging generative models to understand and defend
  against adversarial examples.
\newblock In {\em International Conference on Learning Representations}, 2018.

\bibitem{sorensen1982newton}
Danny~C Sorensen.
\newblock Newton’s method with a model trust region modification.
\newblock {\em SIAM Journal on Numerical Analysis}, 19(2):409--426, 1982.

\bibitem{springenberg2014striving}
Jost~Tobias Springenberg, Alexey Dosovitskiy, Thomas Brox, and Martin
  Riedmiller.
\newblock Striving for simplicity: The all convolutional net.
\newblock {\em arXiv:1412.6806}, 2014.

\bibitem{stutz2019disentangling}
David Stutz, Matthias Hein, and Bernt Schiele.
\newblock Disentangling adversarial robustness and generalization.
\newblock In {\em Proceedings of the IEEE Conference on Computer Vision and
  Pattern Recognition}, pages 6976--6987, 2019.

\bibitem{sundararajan2017axiomatic}
Mukund Sundararajan, Ankur Taly, and Qiqi Yan.
\newblock Axiomatic attribution for deep networks.
\newblock {\em arXiv:1703.01365}, 2017.

\bibitem{szegedy2013intriguing}
Christian Szegedy, Wojciech Zaremba, Ilya Sutskever, Joan Bruna, Dumitru Erhan,
  Ian Goodfellow, and Rob Fergus.
\newblock Intriguing properties of neural networks.
\newblock In {\em International Conference on Learning Representations}, 2014.

\bibitem{carlini2020}
Florian Tramer, Nicholas Carlini, Wieland Brendel, and Aleksander Madry.
\newblock On adaptive attacks to adversarial example defenses.
\newblock In H. Larochelle, M. Ranzato, R. Hadsell, M.~F. Balcan, and H. Lin,
  editors, {\em Advances in Neural Information Processing Systems}, volume~33,
  pages 1633--1645. Curran Associates, Inc., 2020.

\bibitem{tsipras2018robustness}
Dimitris Tsipras, Shibani Santurkar, Logan Engstrom, Alexander Turner, and
  Aleksander Madry.
\newblock Robustness may be at odds with accuracy.
\newblock In {\em International Conference on Learning Representations}, 2019.

\bibitem{van2019interpretable}
Arnaud Van~Looveren and Janis Klaise.
\newblock Interpretable counterfactual explanations guided by prototypes.
\newblock {\em arXiv:1907.02584}, 2019.

\bibitem{verma2020counterfactual}
Sahil Verma, John Dickerson, and Keegan Hines.
\newblock Counterfactual explanations for machine learning: A review.
\newblock {\em arXiv preprint arXiv:2010.10596}, 2020.

\bibitem{wachter2017counterfactual}
Sandra Wachter, Brent Mittelstadt, and Chris Russell.
\newblock Counterfactual explanations without opening the black box: Automated
  decisions and the gpdr.
\newblock {\em Harv. JL \& Tech.}, 31:841, 2017.

\bibitem{wagner2019interpretable}
Jorg Wagner, Jan~Mathias Kohler, Tobias Gindele, Leon Hetzel, Jakob~Thaddaus
  Wiedemer, and Sven Behnke.
\newblock Interpretable and fine-grained visual explanations for convolutional
  neural networks.
\newblock In {\em Proceedings of the IEEE/CVF Conference on Computer Vision and
  Pattern Recognition (CVPR)}, June 2019.

\bibitem{Zeiler2014}
M Zeiler and R Fergus.
\newblock Visualizing and understanding convolutional networks.
\newblock {\em ECCV}, 2014.

\bibitem{Zhang_2017}
Jianming Zhang, Sarah~Adel Bargal, Zhe Lin, Jonathan Brandt, Xiaohui Shen, and
  Stan Sclaroff.
\newblock Top-down neural attention by excitation backprop.
\newblock {\em Int. J. Comput. Vis.}, 126(10):1084–1102, Dec 2017.

\bibitem{Zhou_2016CAM}
Bolei Zhou, Aditya Khosla, Agata Lapedriza, Aude Oliva, and Antonio Torralba.
\newblock Learning deep features for discriminative localization.
\newblock {\em CVPR}, Jun 2016.

\end{thebibliography}
}

 \newpage

 \clearpage
 \appendix

\onecolumn

\pagebreak

\setcounter{equation}{0}
\setcounter{figure}{0}
\setcounter{table}{0}
\makeatletter
\renewcommand{\theequation}{S\arabic{equation}}
\renewcommand{\thefigure}{S\arabic{figure}}
\renewcommand{\thetable}{S\arabic{table}}
\newcommand{\bibnumfmt}[1]{[S#1]}
\newcommand{\citenumfont}[1]{S#1}

\begin{center}
\textbf{\Large Supplementary Material: Explaining Classifiers using Adversarial Perturbations on the Perceptual Ball}
\end{center}

\section{Additional ImageNet Examples}
\label{app:additionalImageNet}

\begin{figure}[h]
\begingroup
\setlength{\tabcolsep}{0pt} 
\renewcommand{\arraystretch}{0.0} 
\begin{center}
\begin{tabular}{p{0.16\linewidth}p{0.16\linewidth}p{0.16\linewidth}p{0.16\linewidth}p{0.16\linewidth}p{0.16\linewidth}}
         \includegraphics[width=\linewidth]{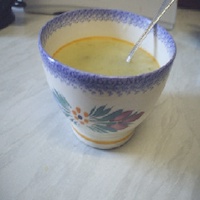}& \includegraphics[width=\linewidth]{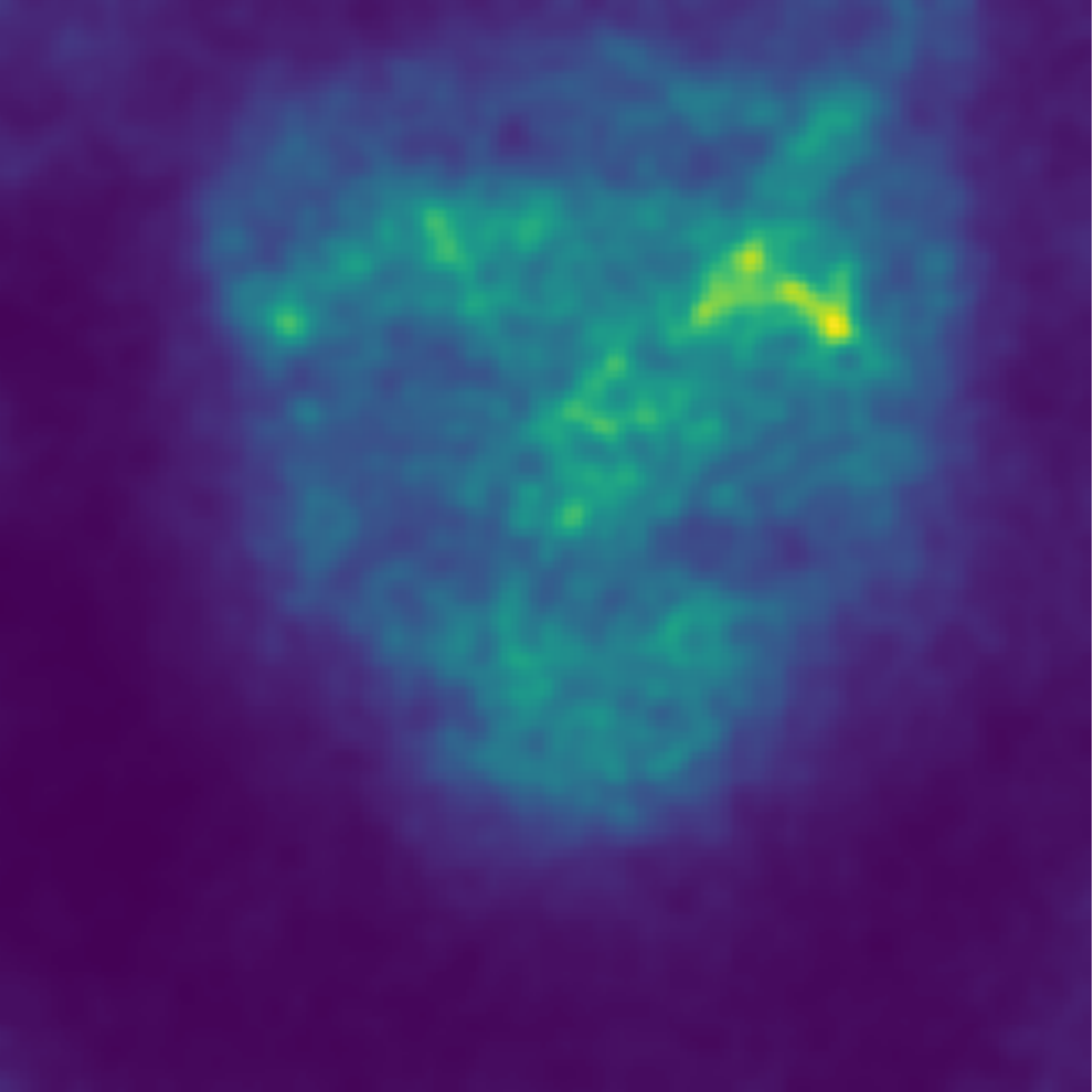}&
    \includegraphics[width=\linewidth]{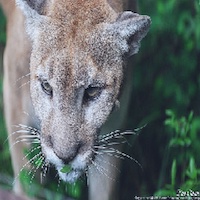}& \includegraphics[width=\linewidth]{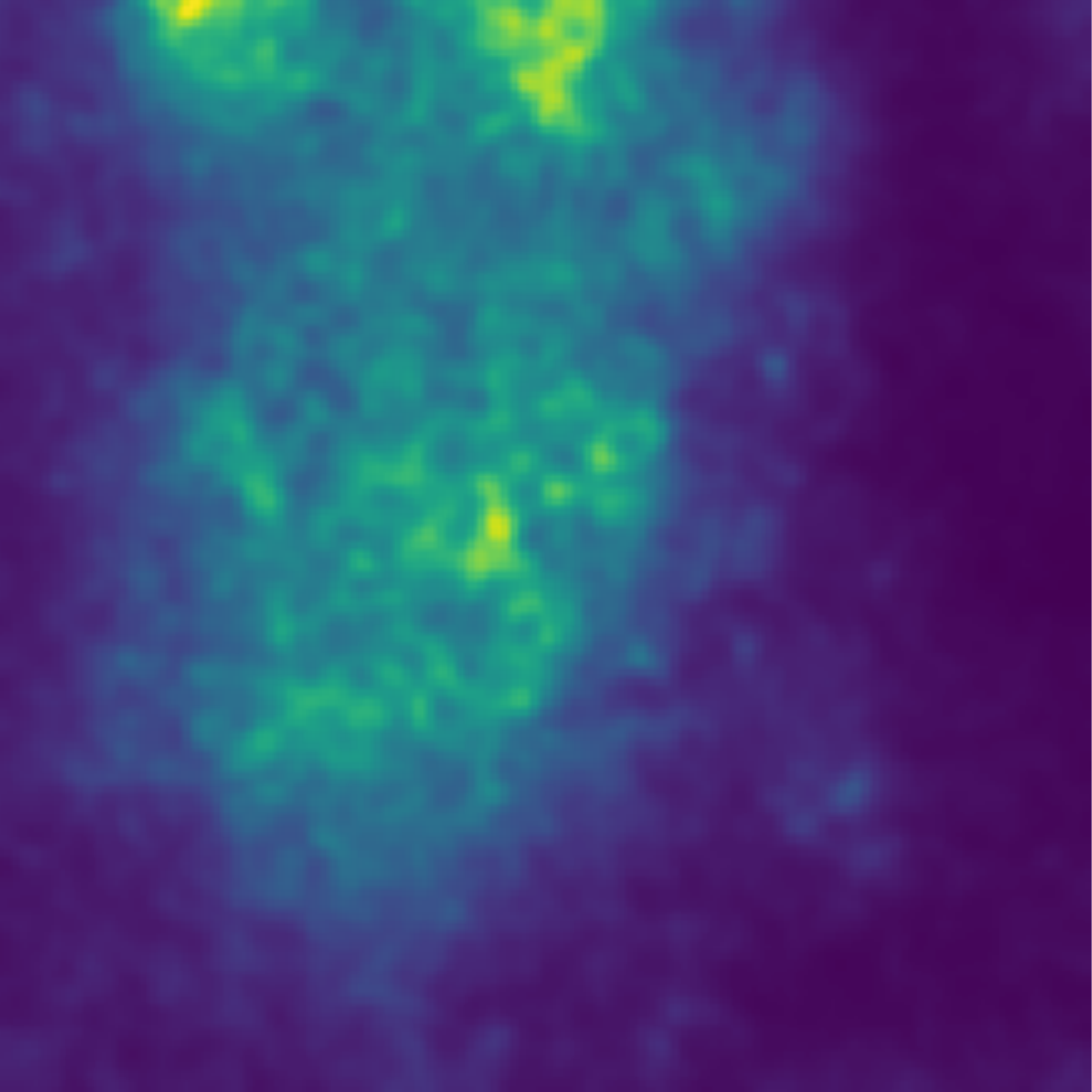}
&
 \includegraphics[width=\linewidth]{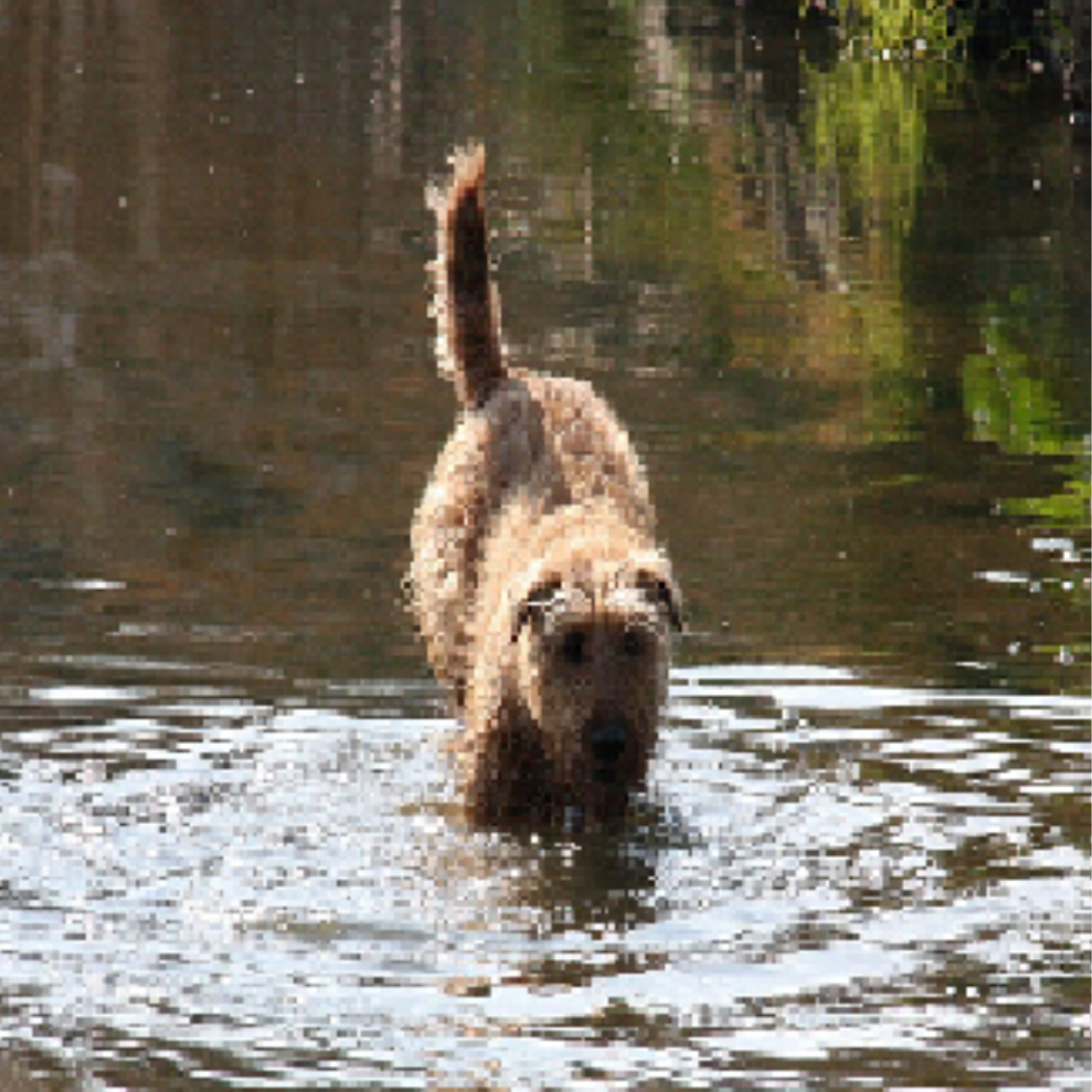}& \includegraphics[width=\linewidth]{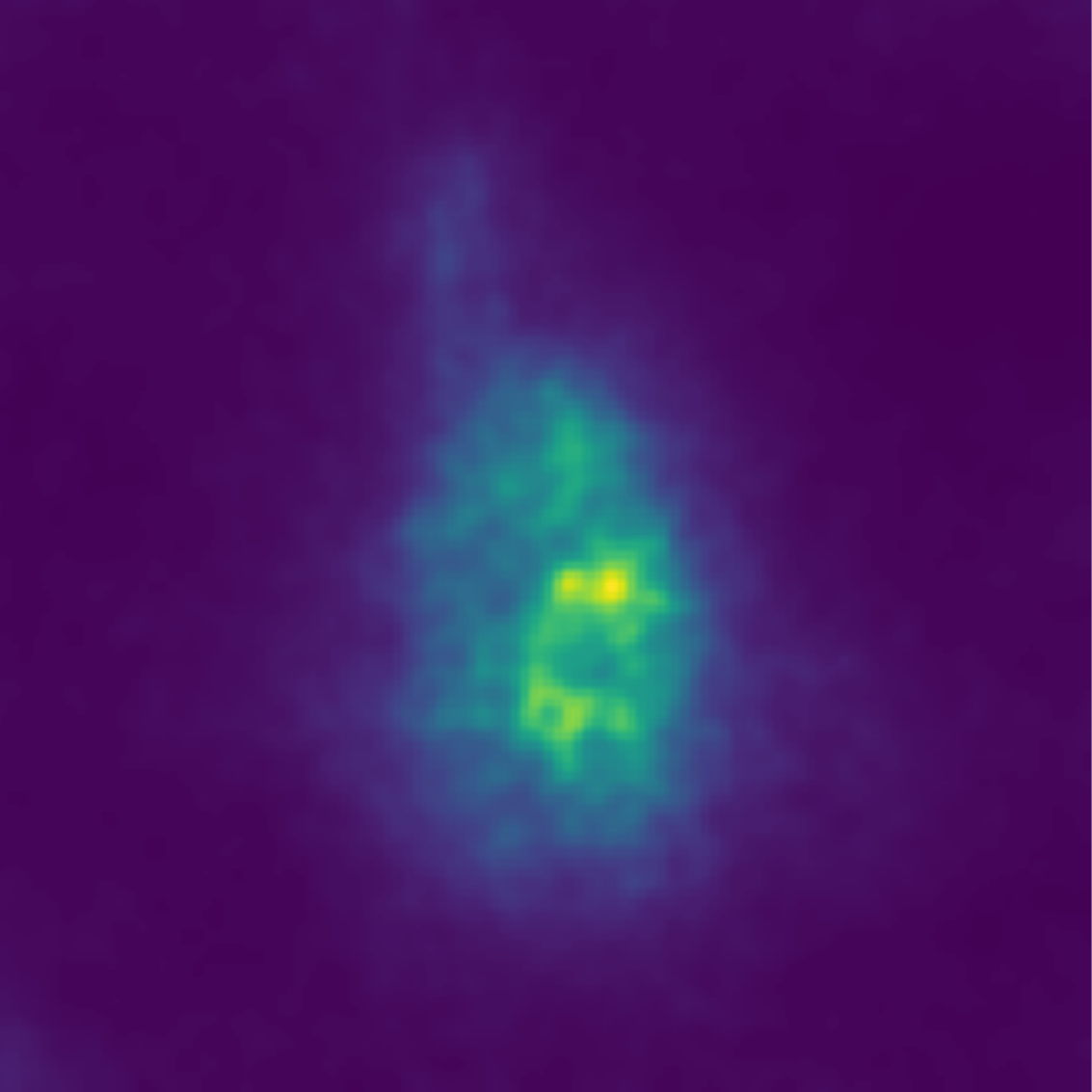} 
 \\
\includegraphics[width=\linewidth]{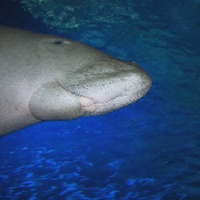}& \includegraphics[width=\linewidth]{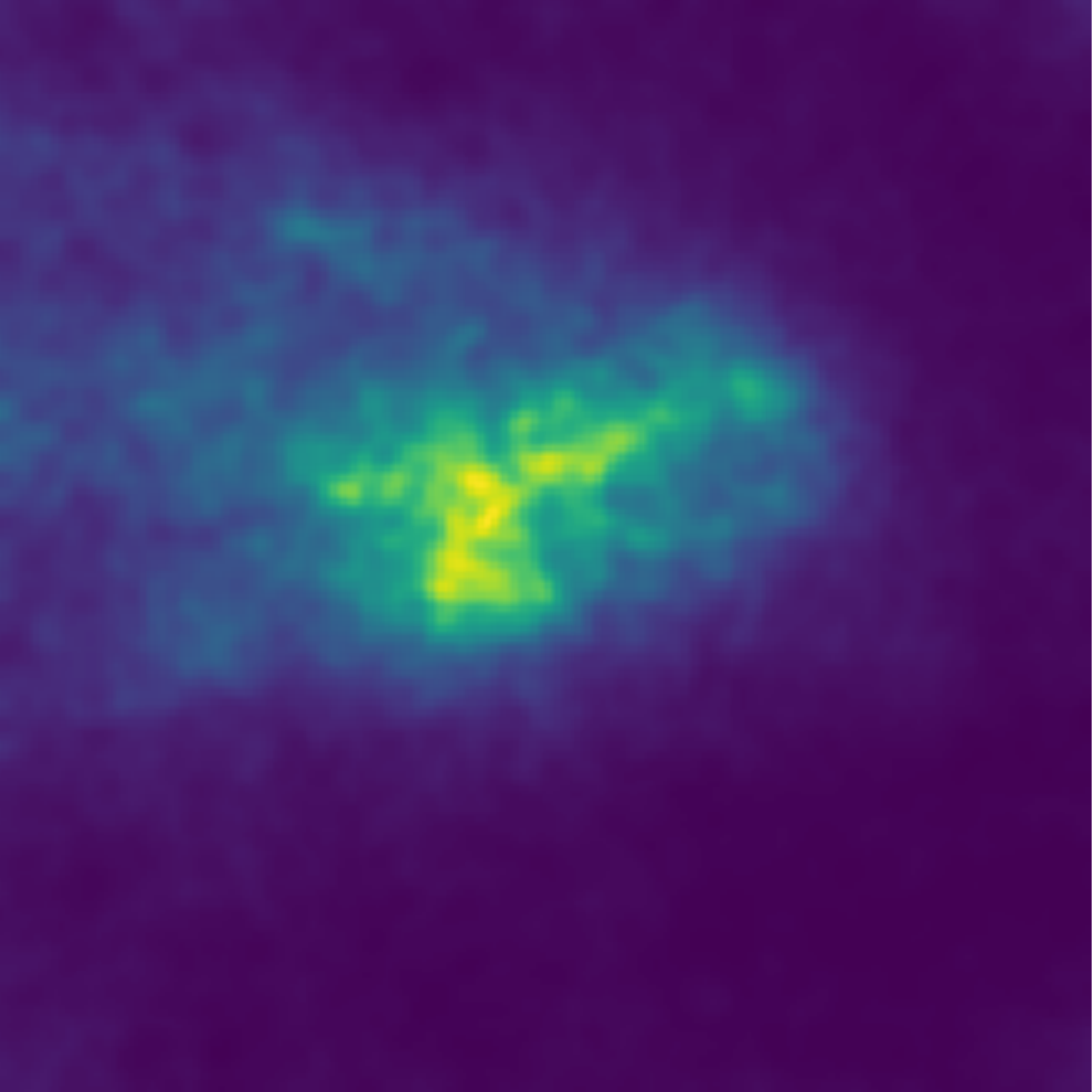}
&
  \includegraphics[width=\linewidth]{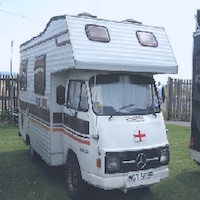}& \includegraphics[width=\linewidth]{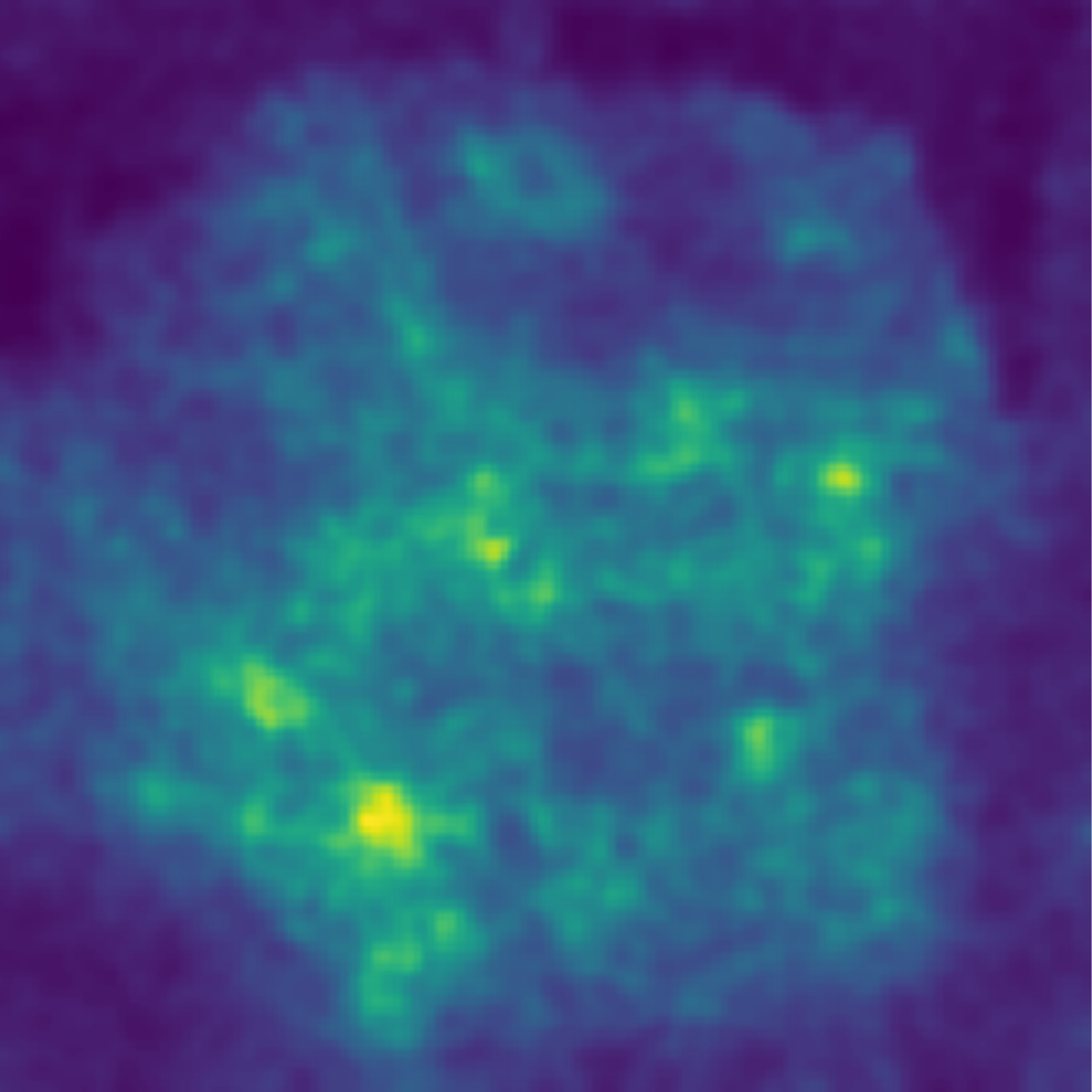}&
  \includegraphics[width=\linewidth]{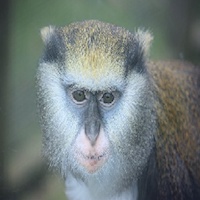}& \includegraphics[width=\linewidth]{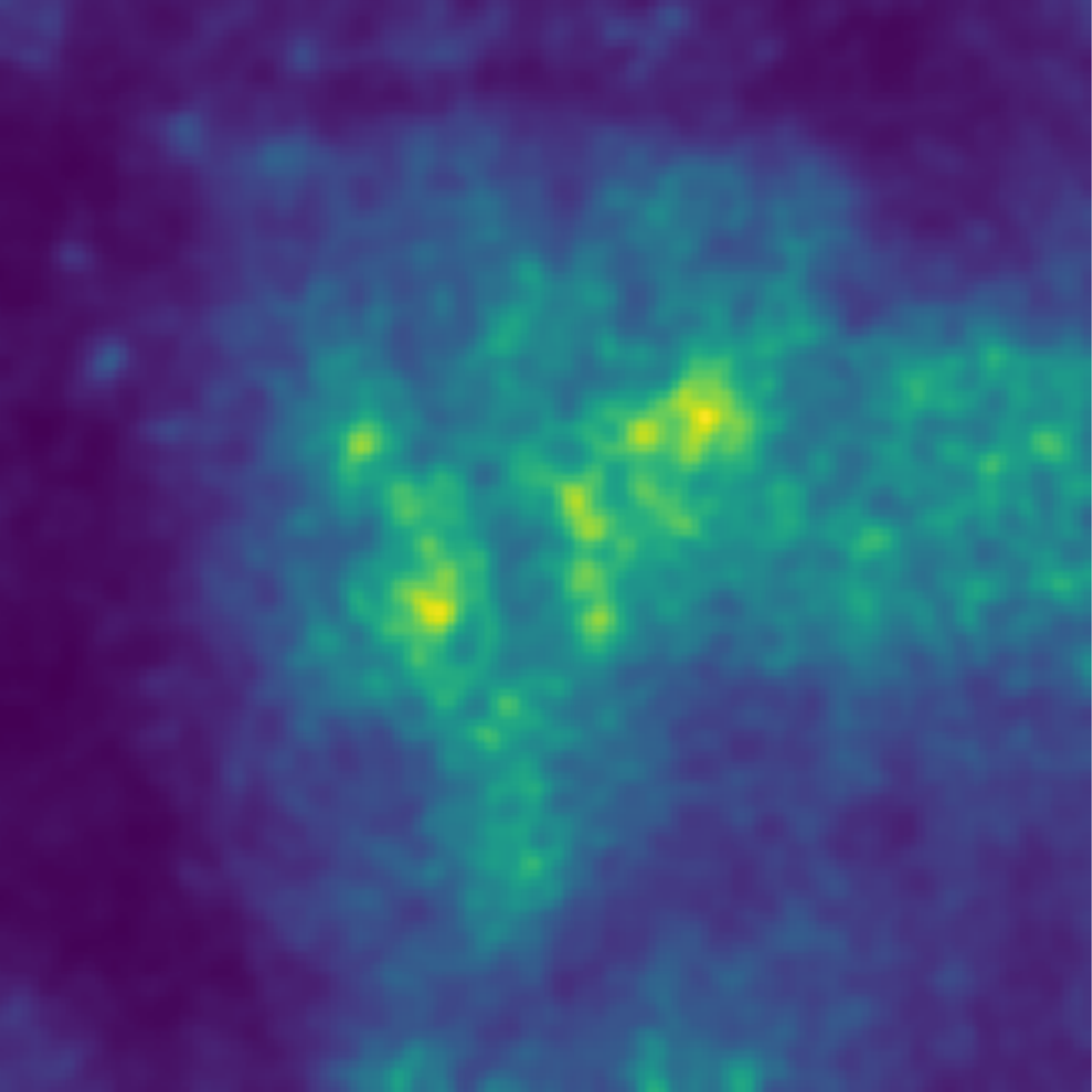}
\\
  \includegraphics[width=\linewidth]{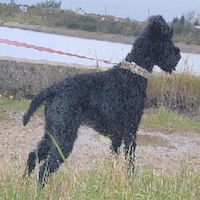}& \includegraphics[width=\linewidth]{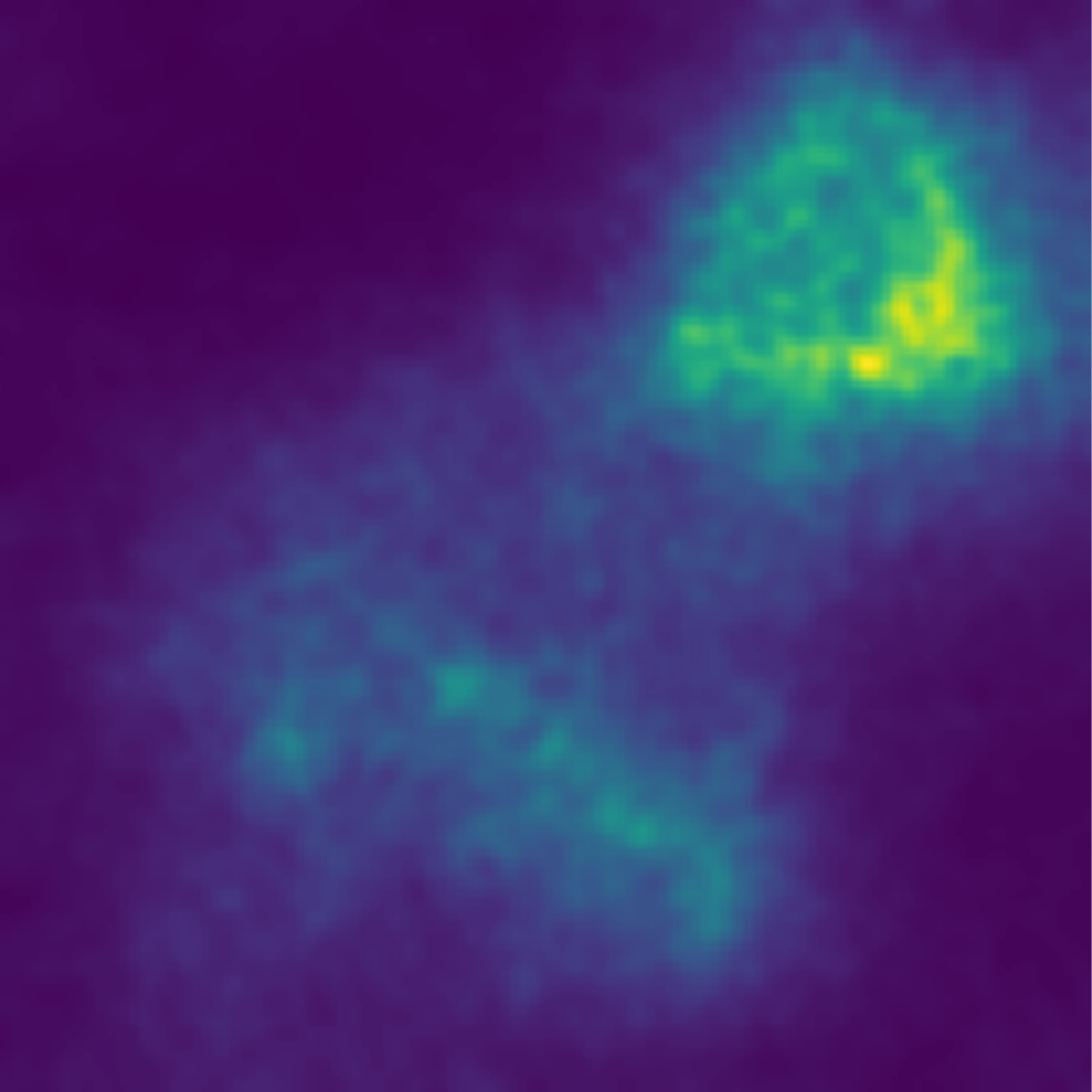} & 
\includegraphics[width=\linewidth]{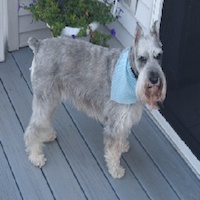}& \includegraphics[width=\linewidth]{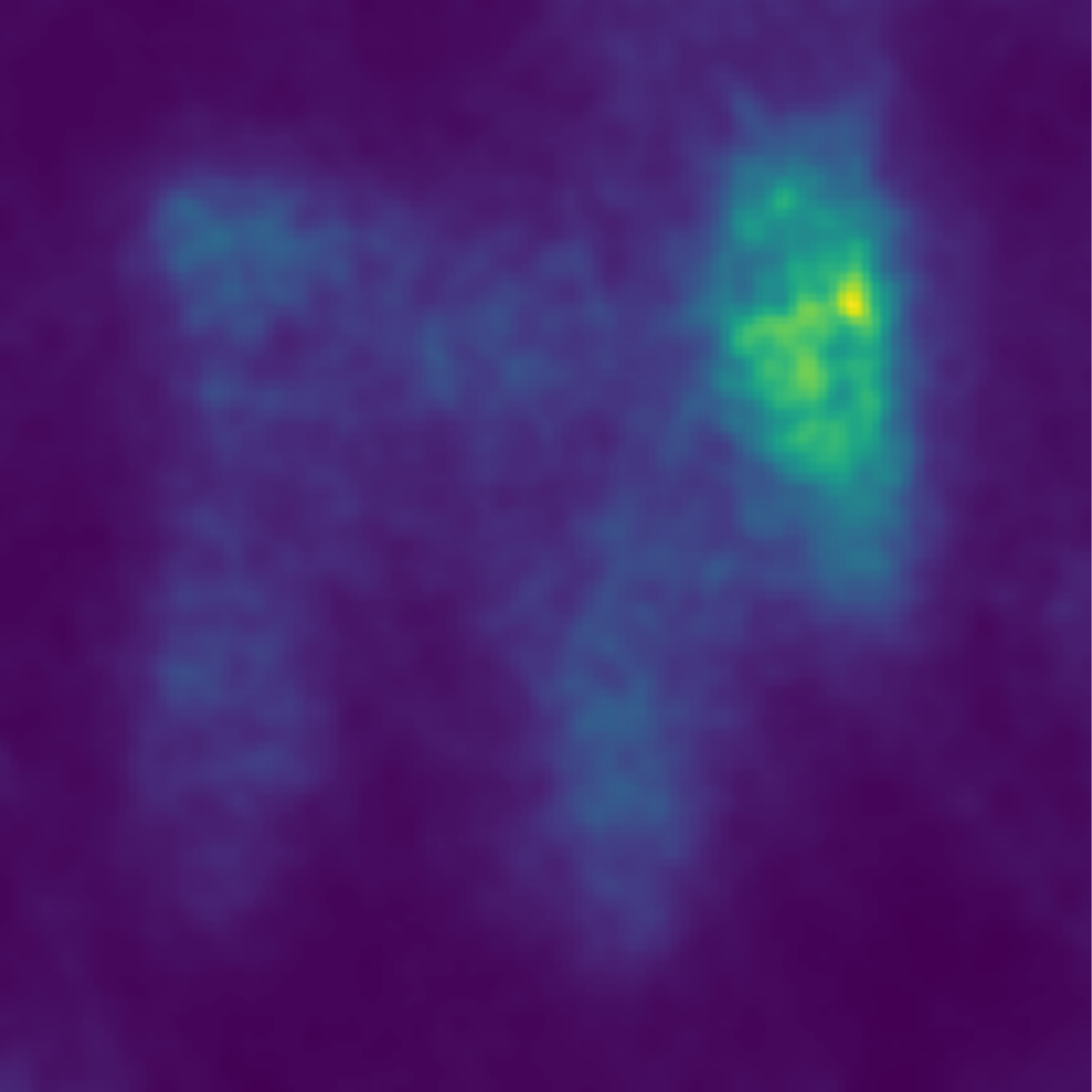} 
    &
    \includegraphics[width=\linewidth]{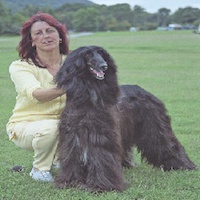}& \includegraphics[width=\linewidth]{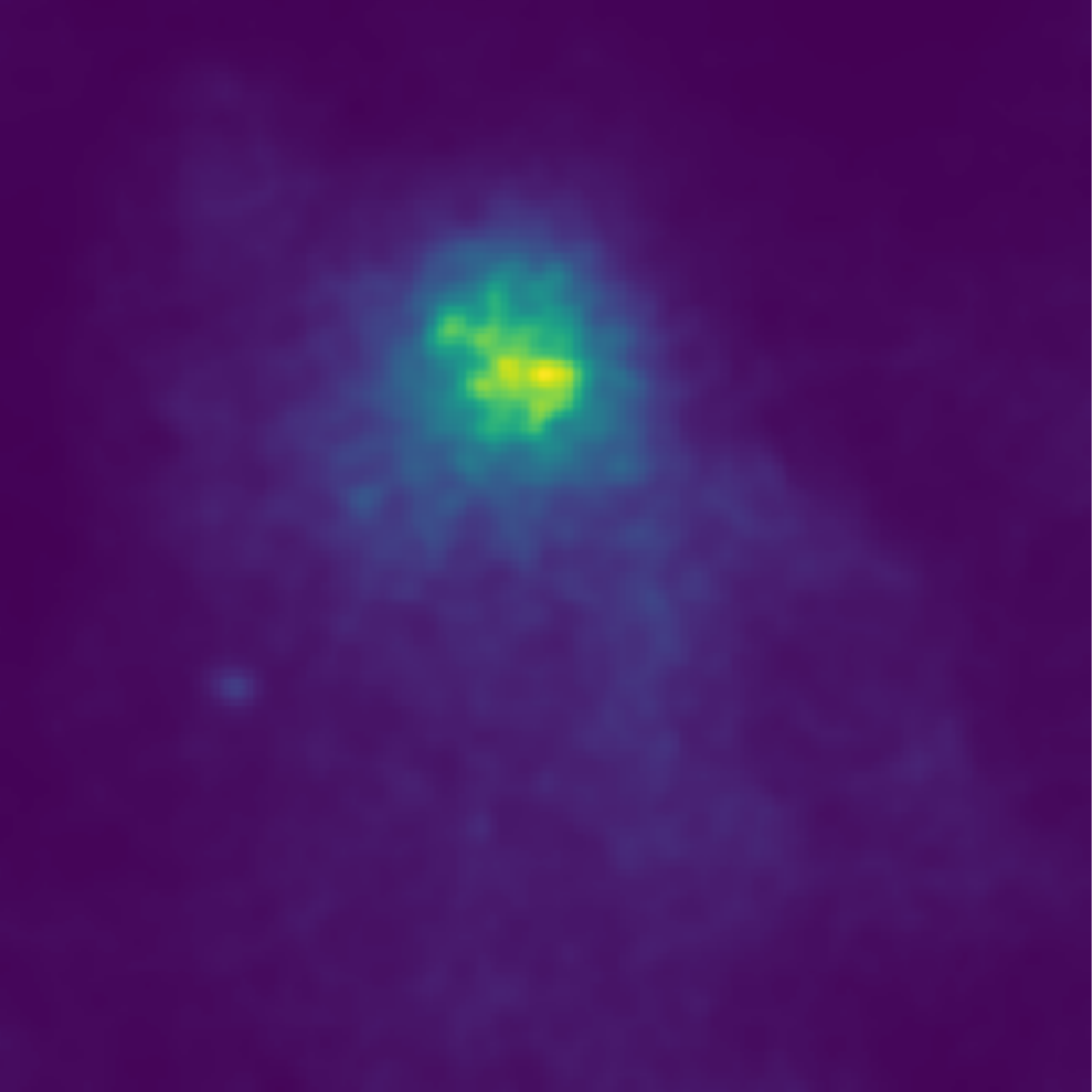}
    \\
   \includegraphics[width=\linewidth]{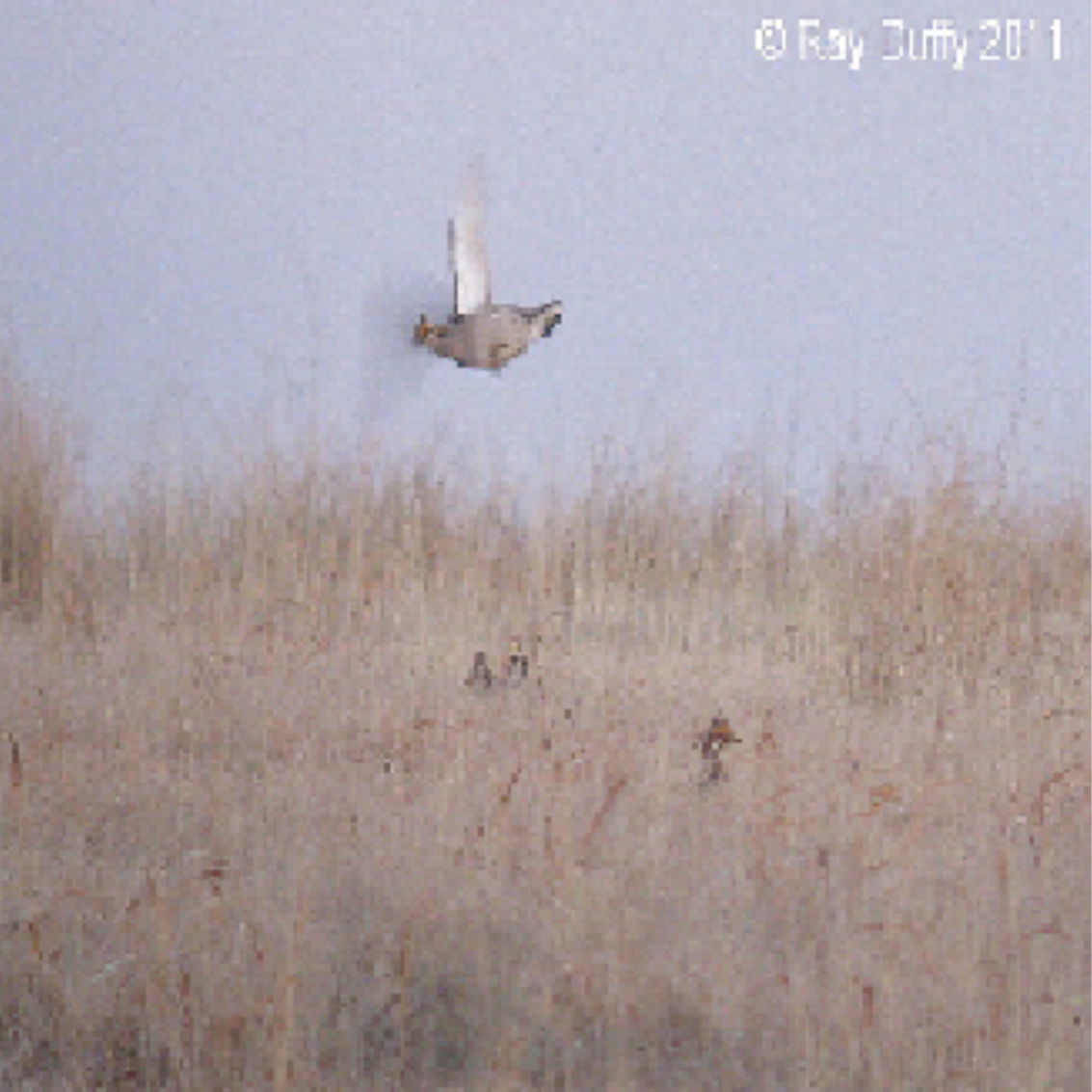}& \includegraphics[width=\linewidth]{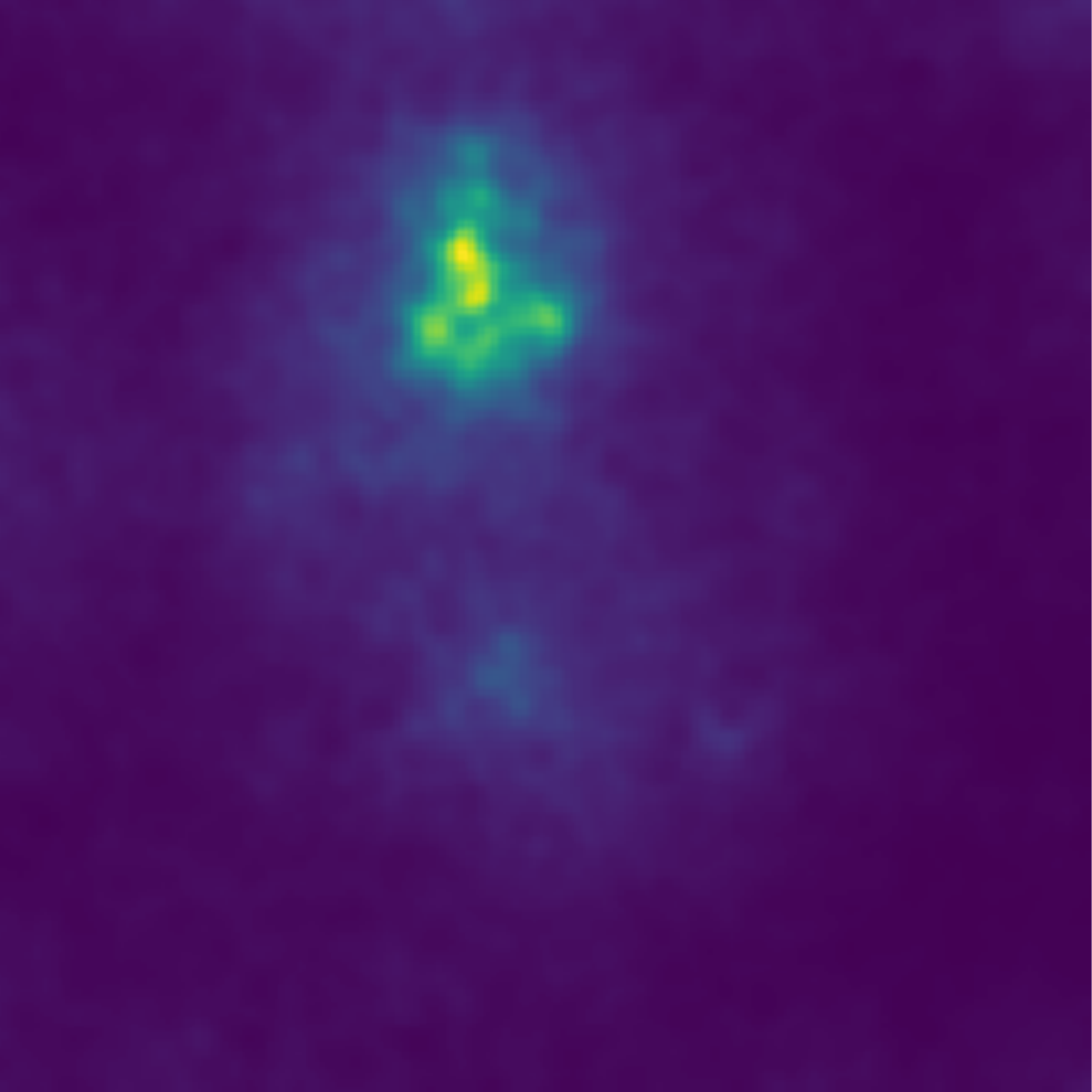}
   &
  \includegraphics[width=\linewidth]{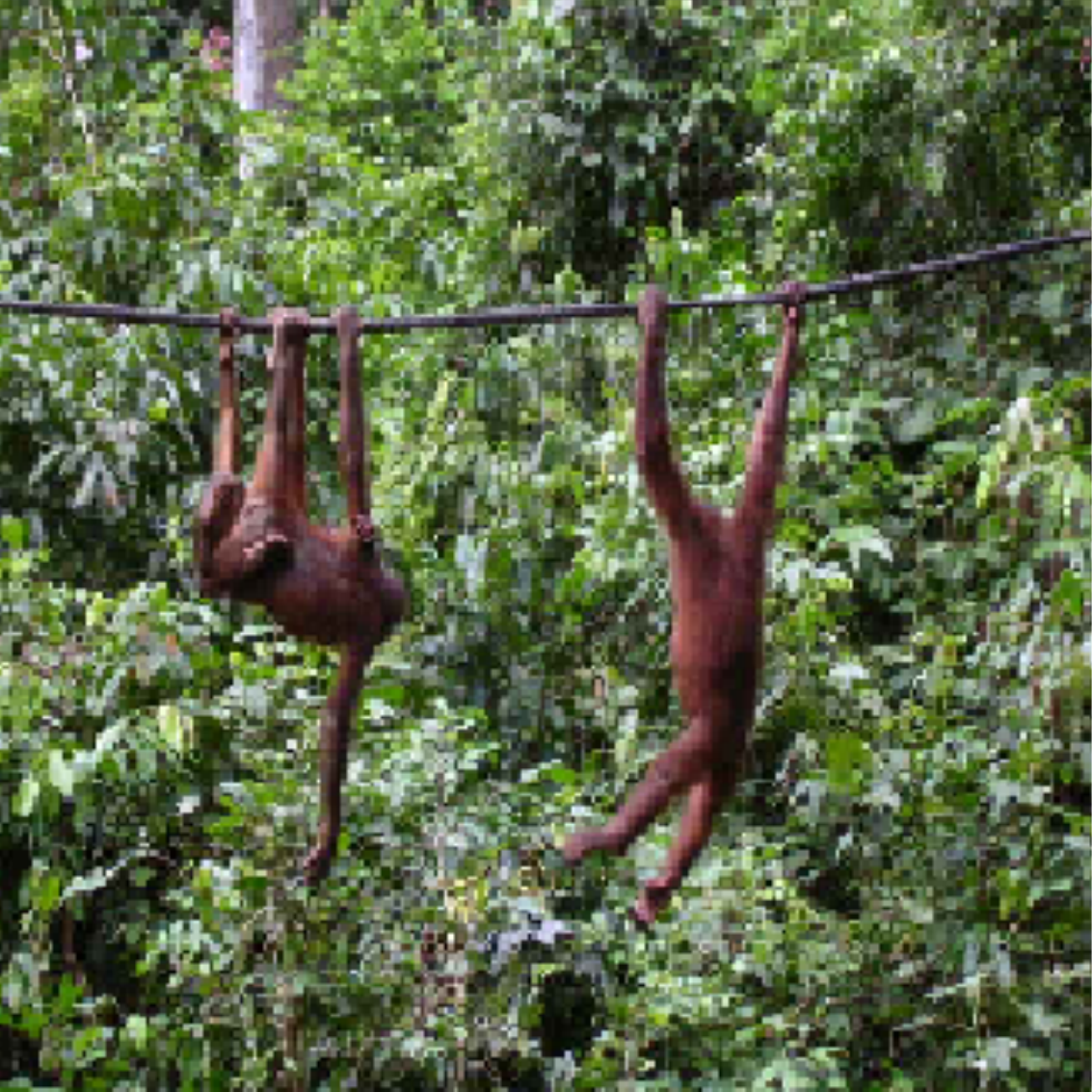}& 
  \includegraphics[width=\linewidth]{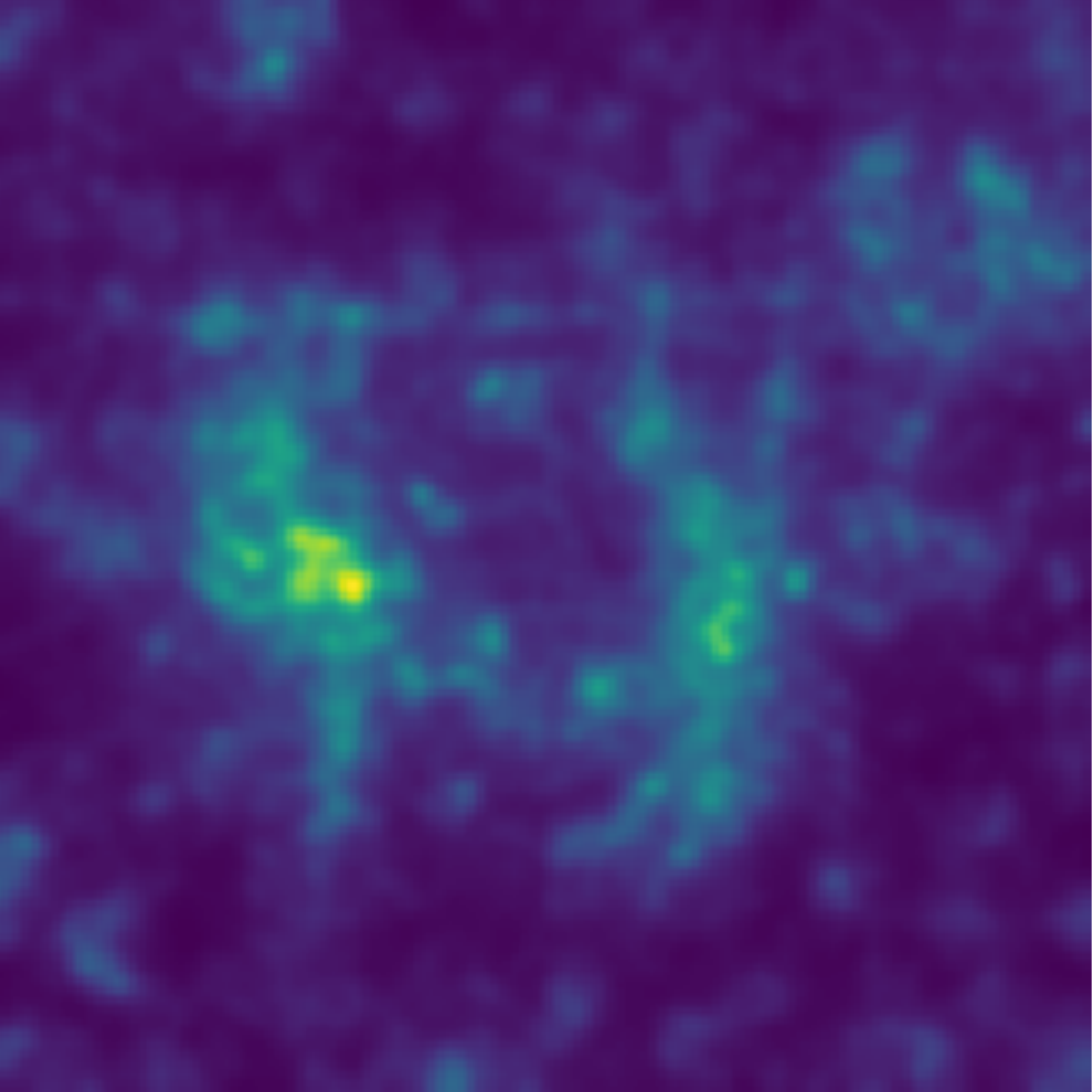}&
  \includegraphics[width=\linewidth]{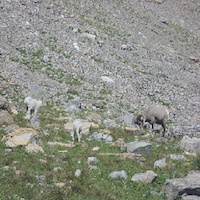}& \includegraphics[width=\linewidth]{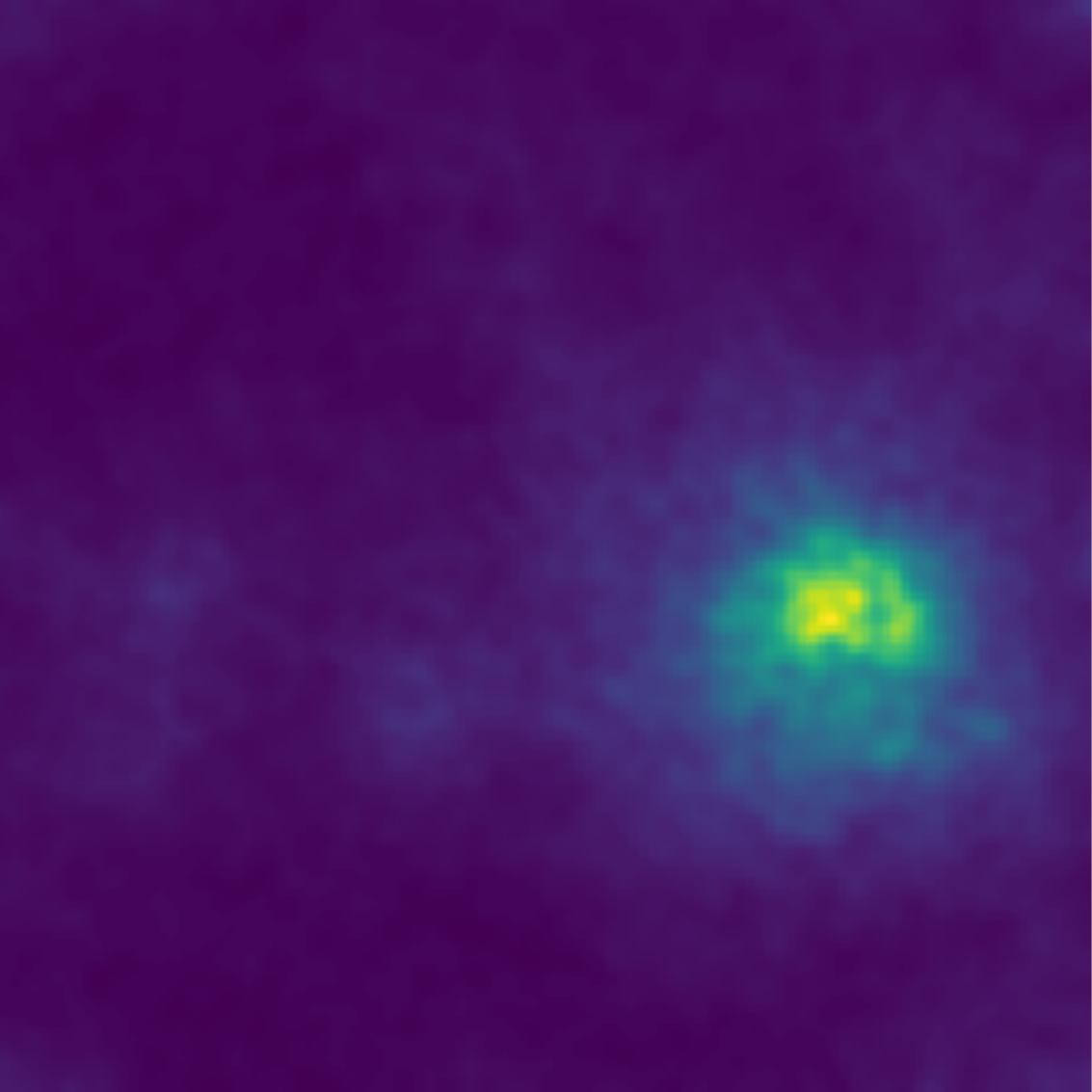}
\\
     \includegraphics[width=\linewidth]{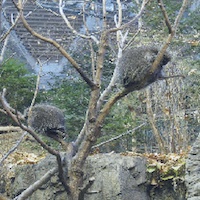}& \includegraphics[width=\linewidth]{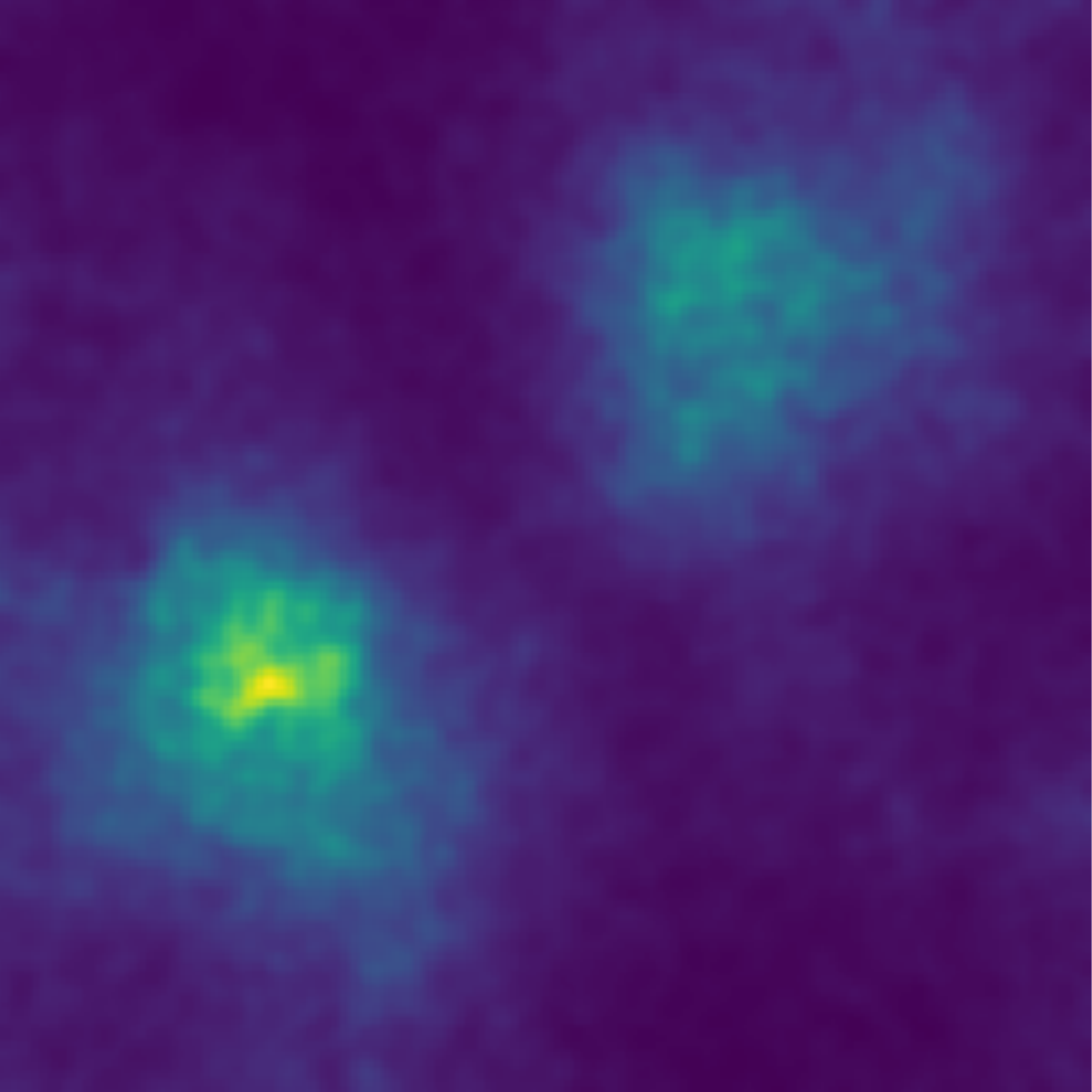}&
     \includegraphics[width=\linewidth]{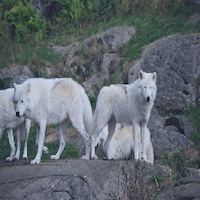}& \includegraphics[width=\linewidth]{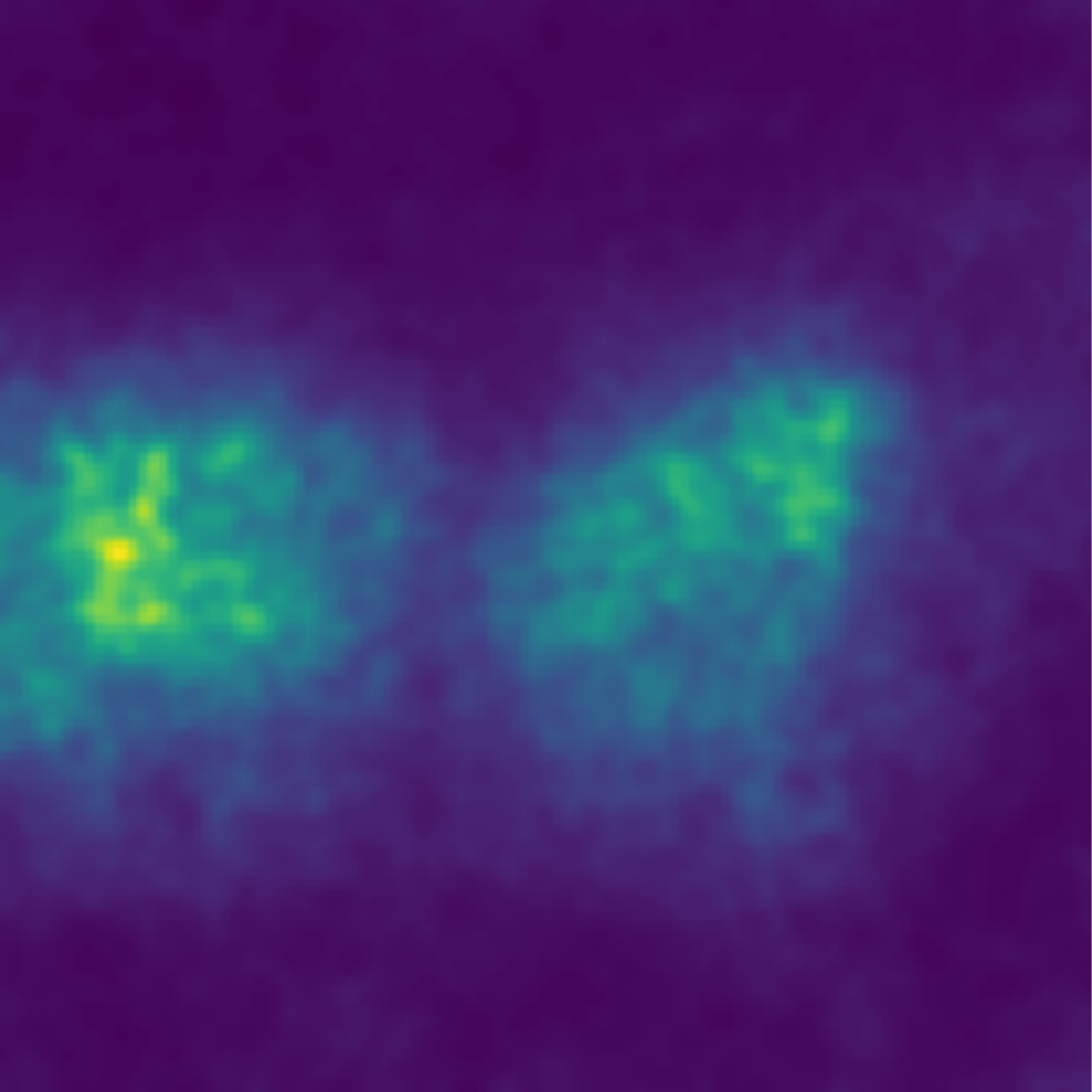}
     &
 \includegraphics[width=\linewidth]{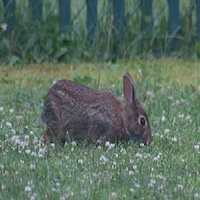}& \includegraphics[width=\linewidth]{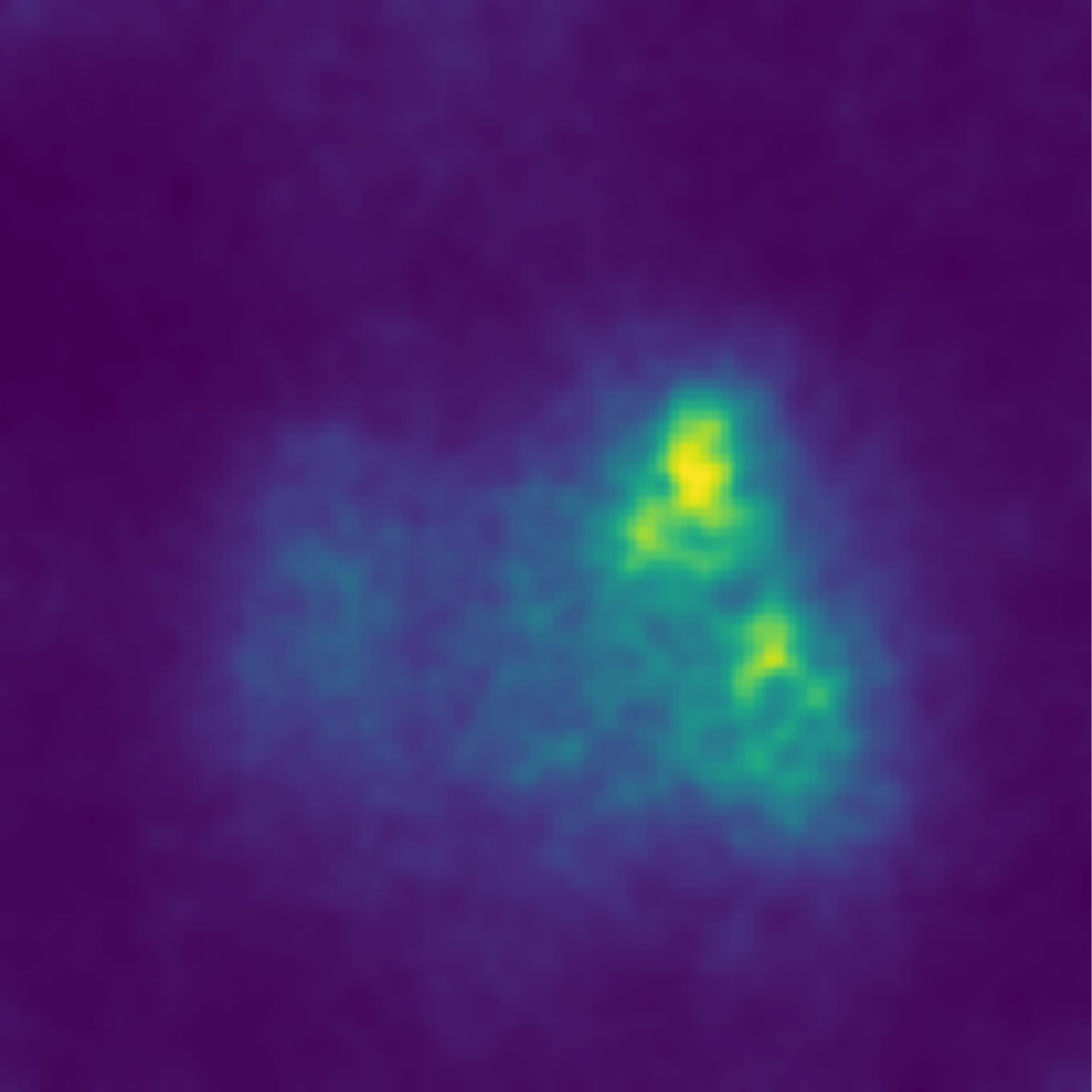}
 \\
   \includegraphics[width=\linewidth]{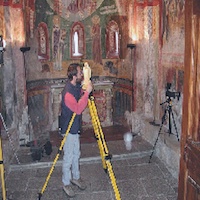}& \includegraphics[width=\linewidth]{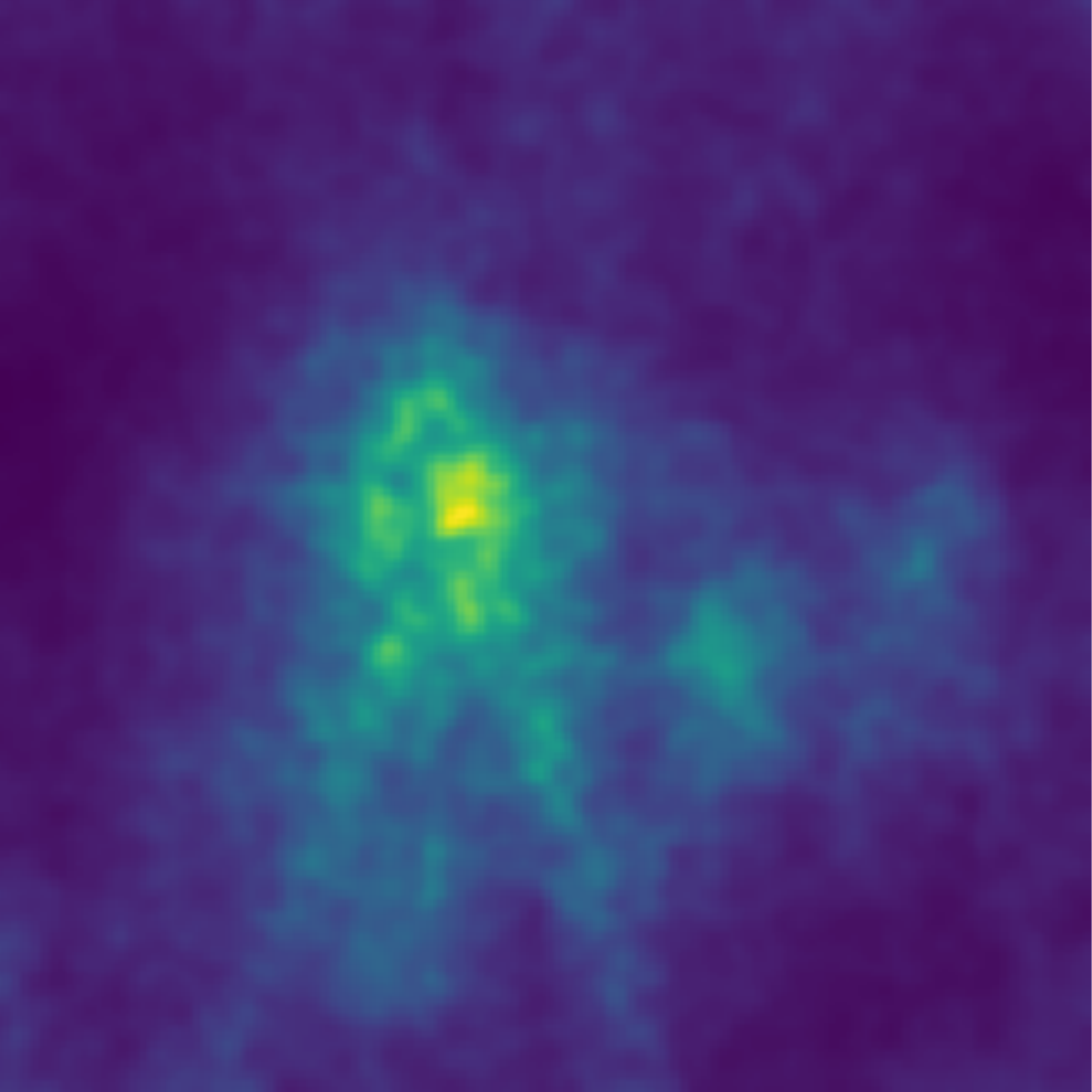}&
       \includegraphics[width=\linewidth]{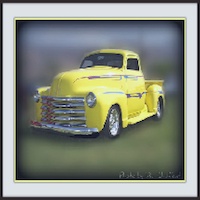}& \includegraphics[width=\linewidth]{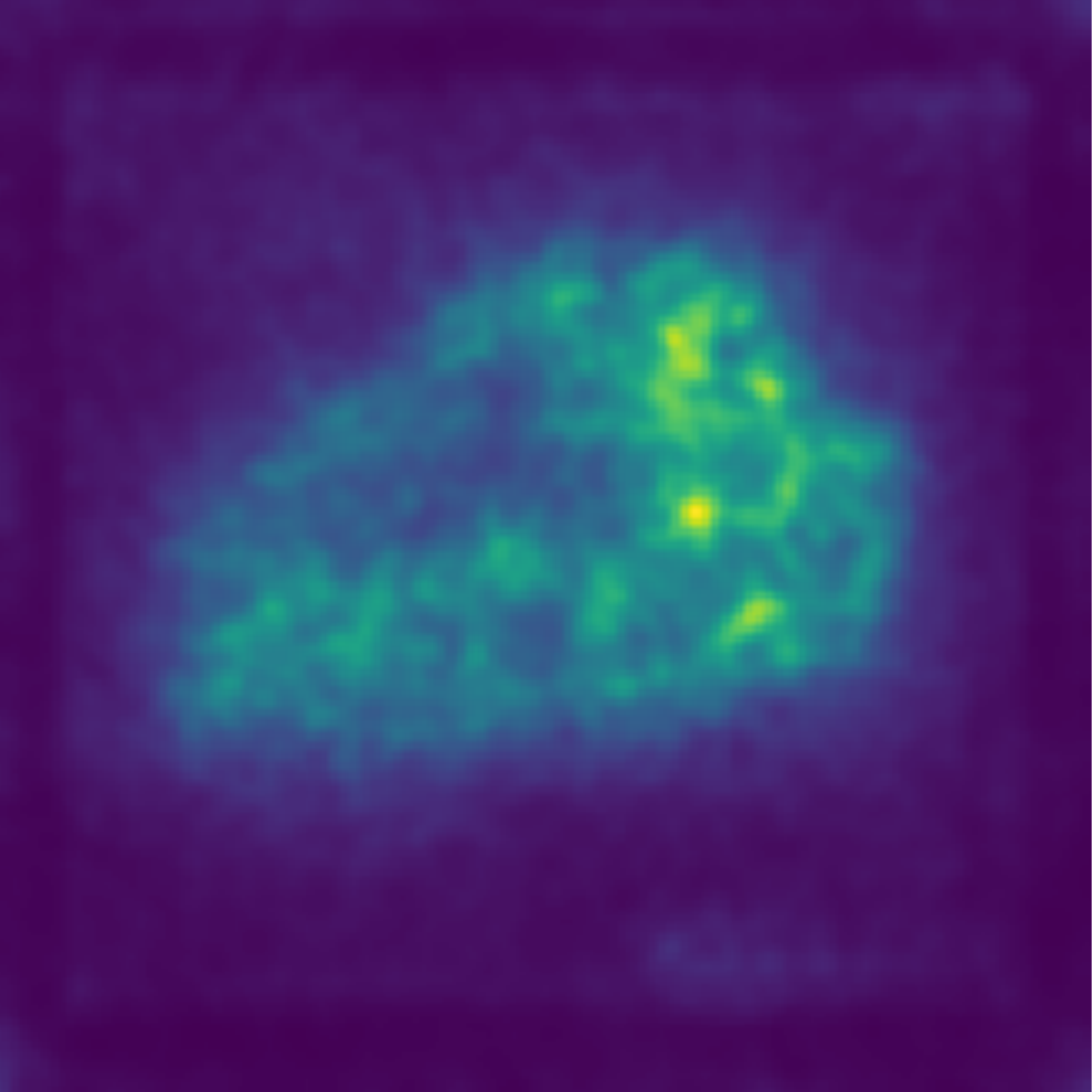}
       &
       \includegraphics[width=\linewidth]{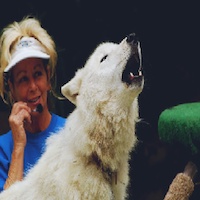}& \includegraphics[width=\linewidth]{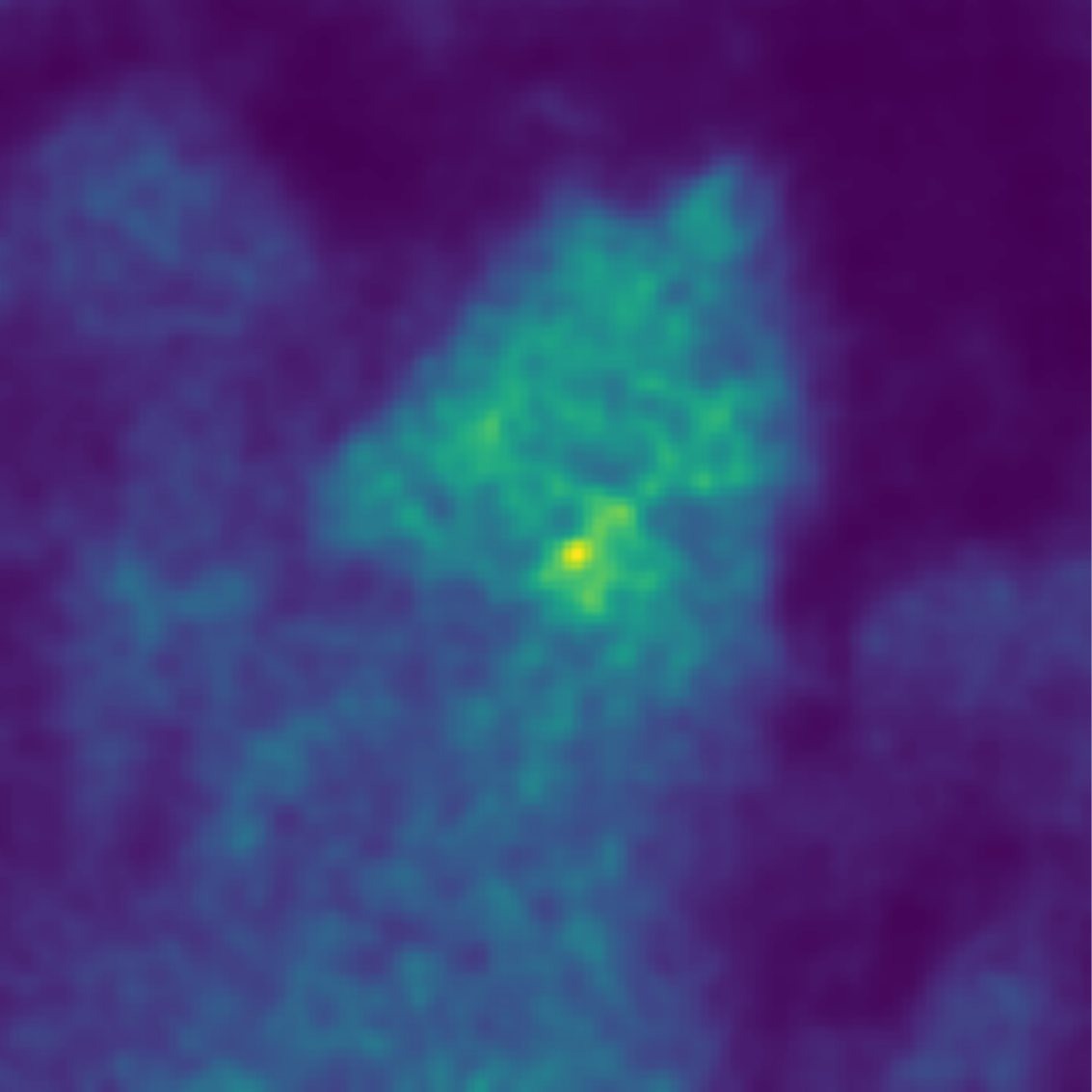}       
 \end{tabular}
\end{center}
\endgroup
 \caption{Additional Perceptual Perturbations on ImageNet as explanations. Illustration of perceptual perturbations on typical images taken from ImageNet. 
 \label{fig:SIppe}}
\end{figure}

\newpage

  \begin{figure}[H]
  \hspace{-2.5mm}
  \mbox{
  \includegraphics[width=0.17\linewidth]{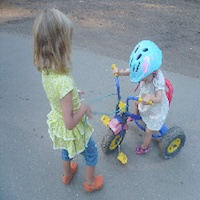}  
  \hspace{-1.5mm}
  \includegraphics[width=0.17\linewidth]{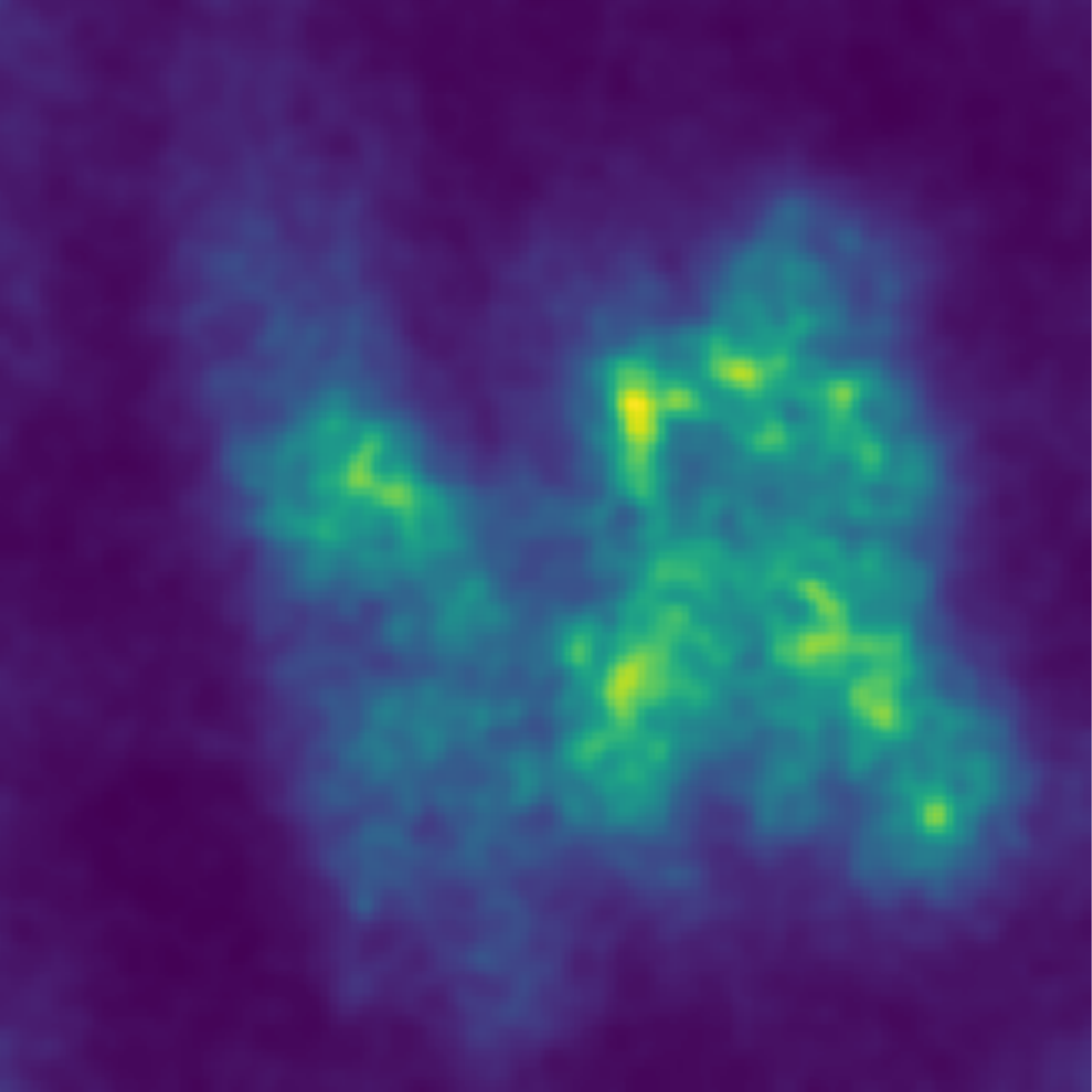}  
  \hspace{-1.5mm}
  \includegraphics[width=0.17\linewidth]{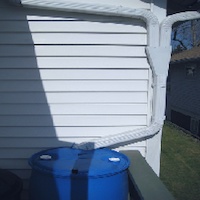}   
  \hspace{-1.5mm}
  \includegraphics[width=0.17\linewidth]{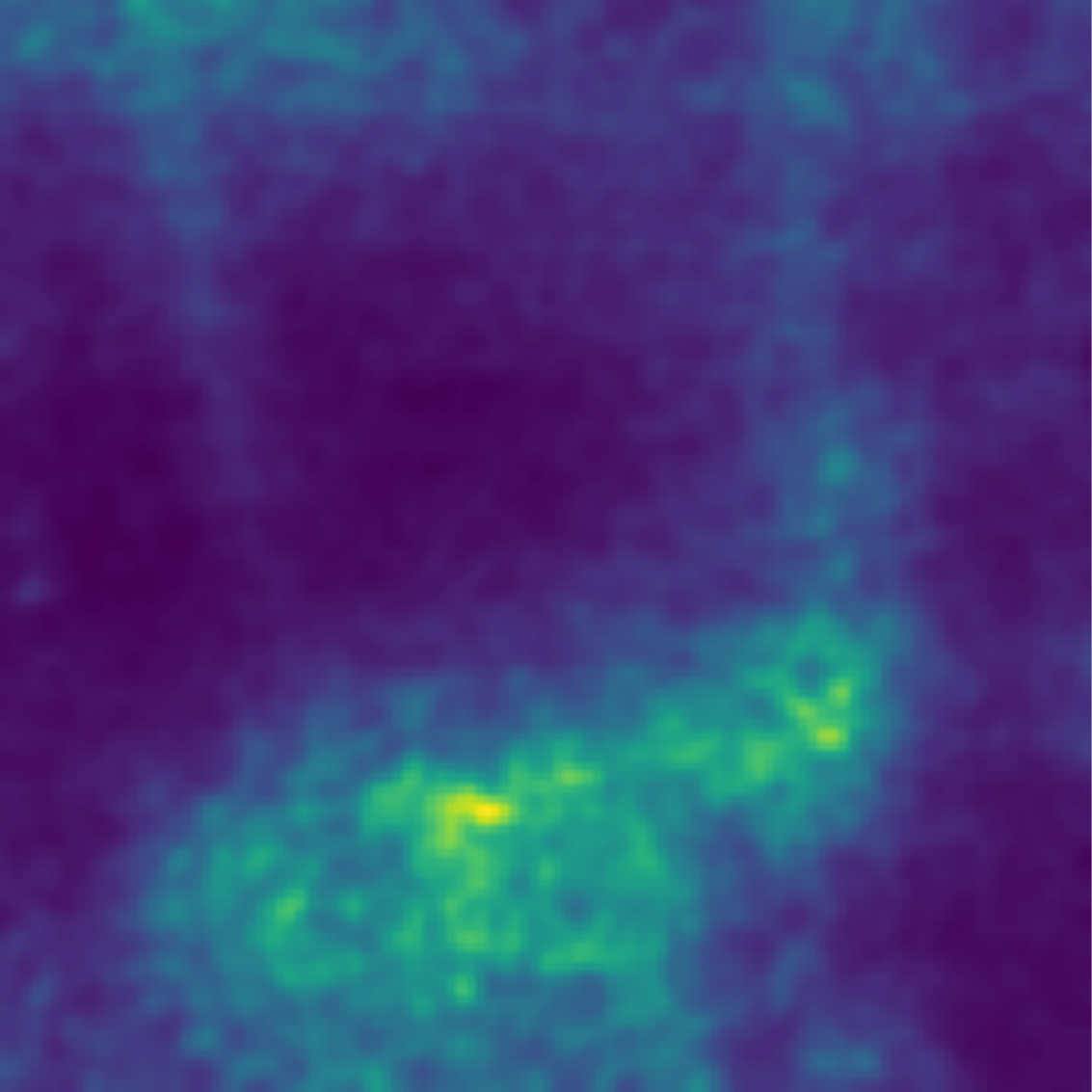}  
  \hspace{-1.5mm}
  \includegraphics[width=0.17\linewidth]{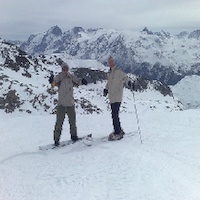}  
  \hspace{-1.5mm}
  \includegraphics[width=0.17\linewidth]{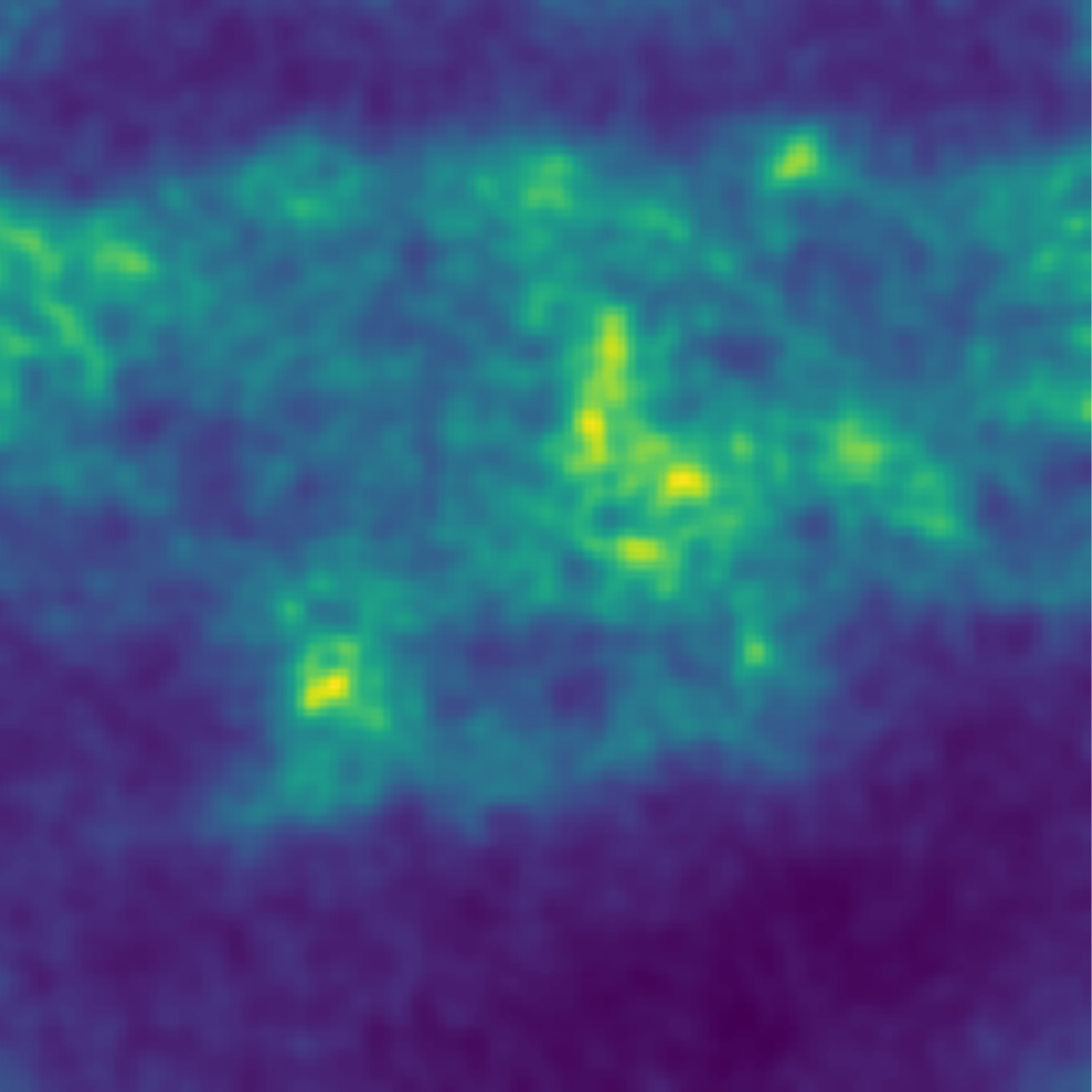}} 

  \caption{Failed explanation examples. From left to right.  
  (i) the children near the tricycle;
  (ii) the pipes of the rain barrel; 
  (iii) people in front of the alp;  are found to be salient.
    \label{failure_eg}
  }
  \end{figure}

\section{Additional Results}
\label{additionalAblation}

\begin{figure}[h]
\includegraphics[width=0.495\linewidth]{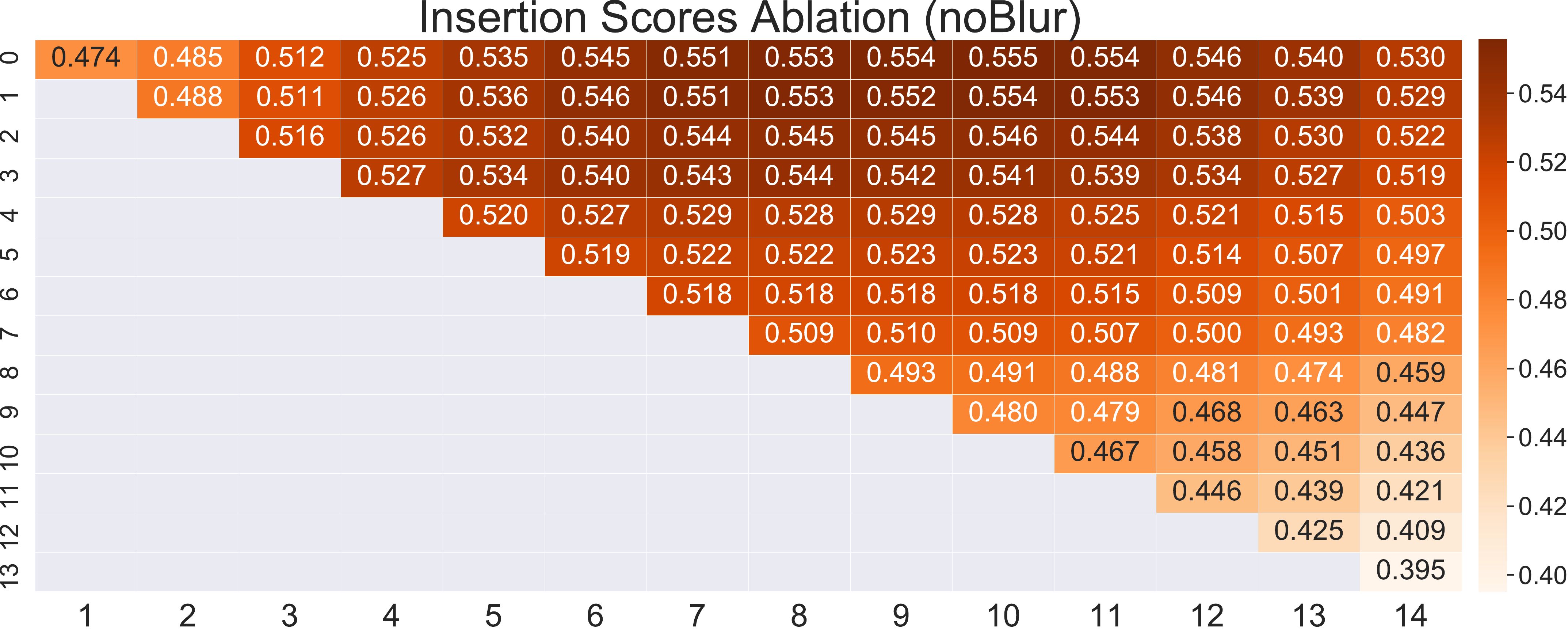}
\includegraphics[width=0.495\linewidth]{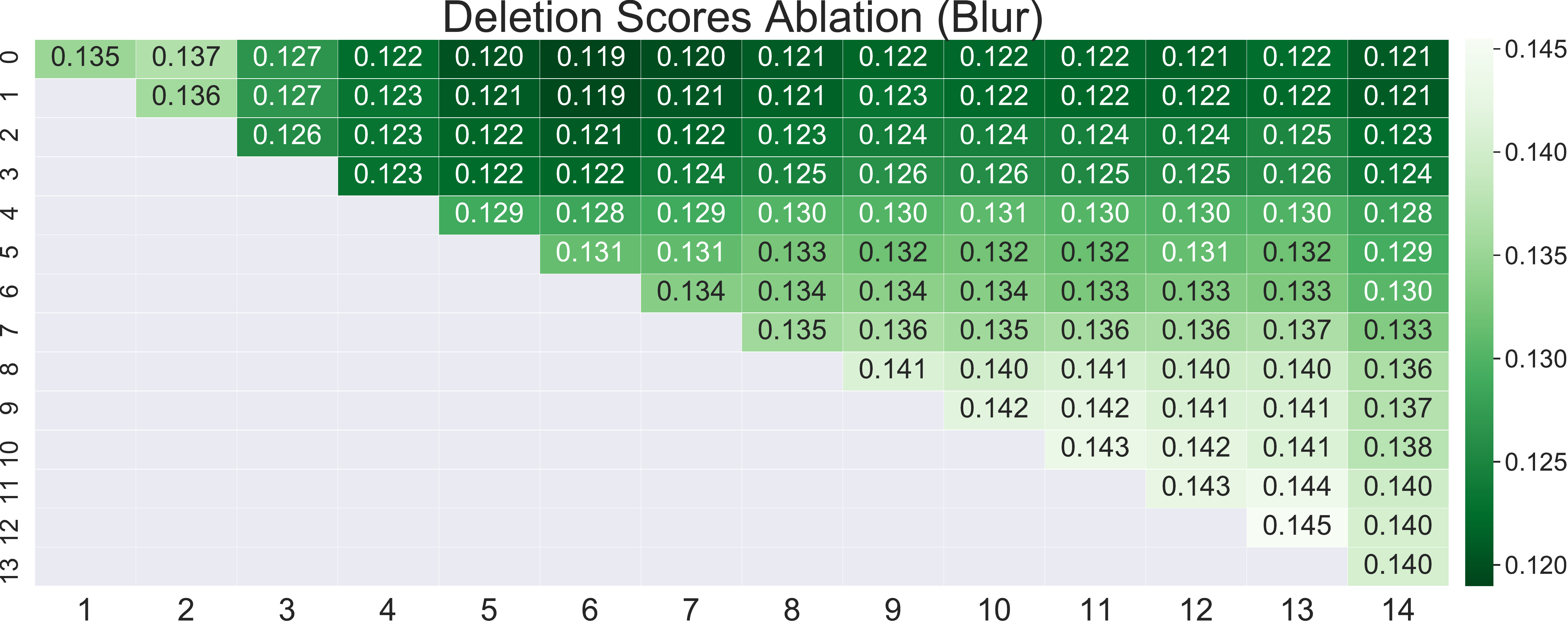}
\caption{Additional Sensitivity study on the choice of layers 
for the insertion deletion game.
On left we display the Insertion results for the no Blur variant (higher is better) and on the right we display the Deletion results for the blur variant (lower is better). 
\label{fig:heatmapAdditionalInsertion}
}
\end{figure}

\begin{figure}[b]
\includegraphics[width=0.495\linewidth]{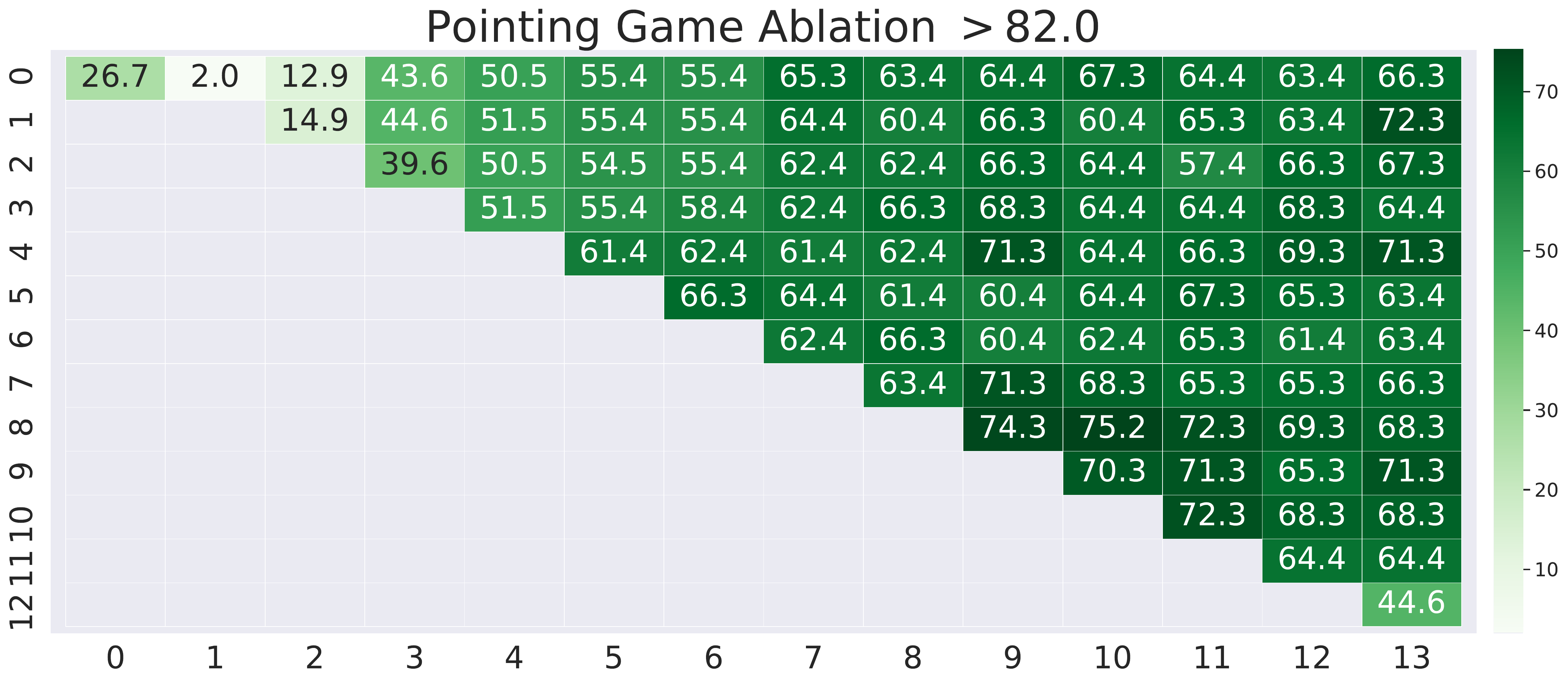}
\includegraphics[width=0.495\linewidth]{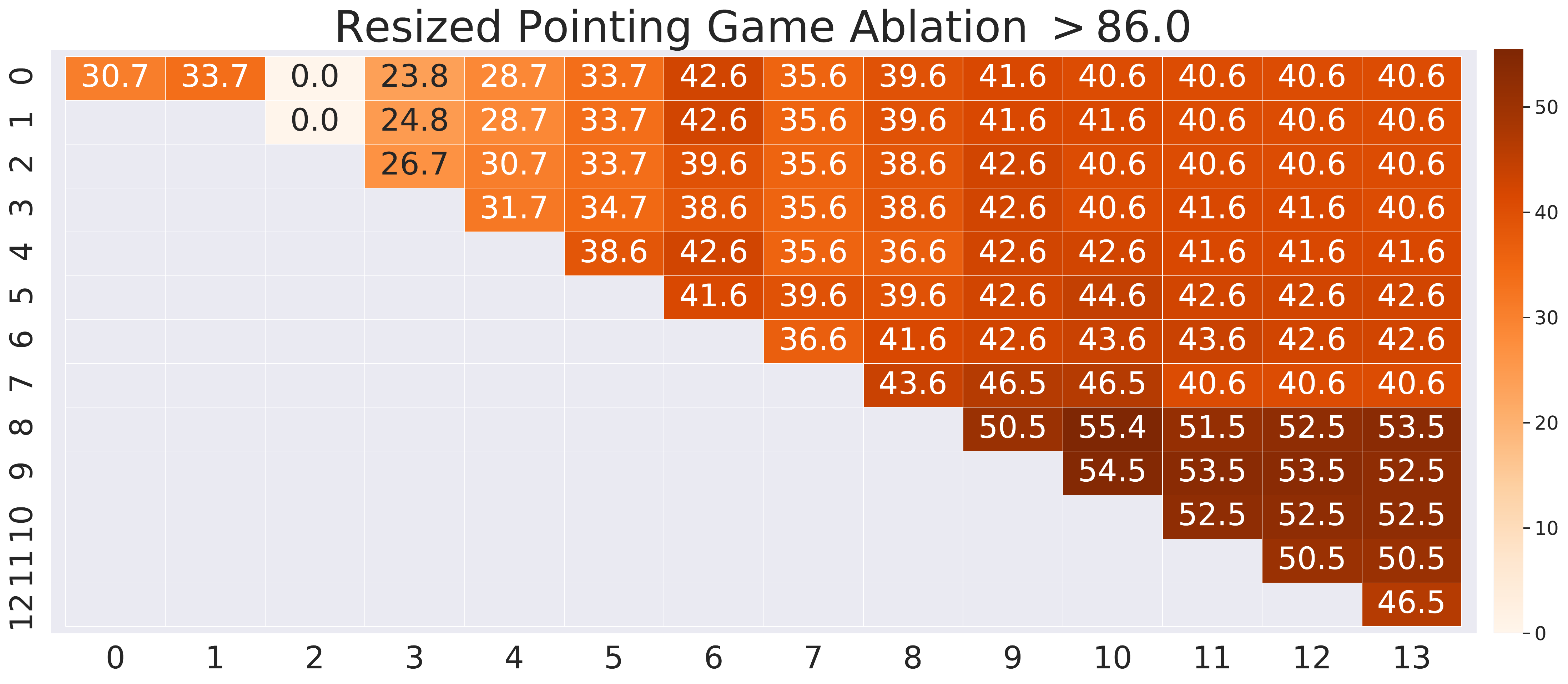}
\caption{Sensitivity study on the choice of layers regularized for the pointing game. To test how robust each result we display the number of $\sigma$s which have performance above $82\%$ for the regular image size (left) and $86\%$ for the resized image (right). 
\label{fig:heatmapPointingGame}
}
\end{figure}

In this section, we display a small number of additional results to complement the results in the paper. This consists of additional ablation results, for the insertion deletion game (Fig.~\ref{fig:heatmapAdditionalInsertion}), and the pointing game (Fig.~\ref{fig:heatmapPointingGame}). This section also contains a discussion of the parameters in the pointing game with  
Sec.~\ref{app:additionalAblationSigma} discussing the parameter choice for our  
no perceptual baseline, and 
Sec.~\ref{sec:limitedSigma} presenting results on the pointing game with a more limited set of $\sigma$s.

\subsection{Sigma Selection for the No Perceptual baseline in the pointing game} 
\label{app:additionalAblationSigma}
To compare against the no perceptual baseline in the pointing game, we select 
the best sigma found in our ablation study. 
In the non resized case $\sigma \in \{61.0,62.0\}$, and 
we select the largest value $62.0$. For the resized case, 
$\sigma \in \{31.0,32.0,33.0,36.0,37.0,49.0,52.0\}$ have 
equal performance and we select $37.0$, as it is the largest 
value in the mostly continuous sequence from $31$ to $37$.

\subsection{Pointing Game with Limited  $\mathbf{\sigma}$}
\label{sec:limitedSigma}
The range of $\sigma$ used in the pointing game is relatively large with $101$ possible values. Furthermore, as we are using $0$ padded blurs, the larger values of $\sigma$ will concentrate more on the center of the image. 

\begin{figure}
\includegraphics[width=0.495\linewidth]{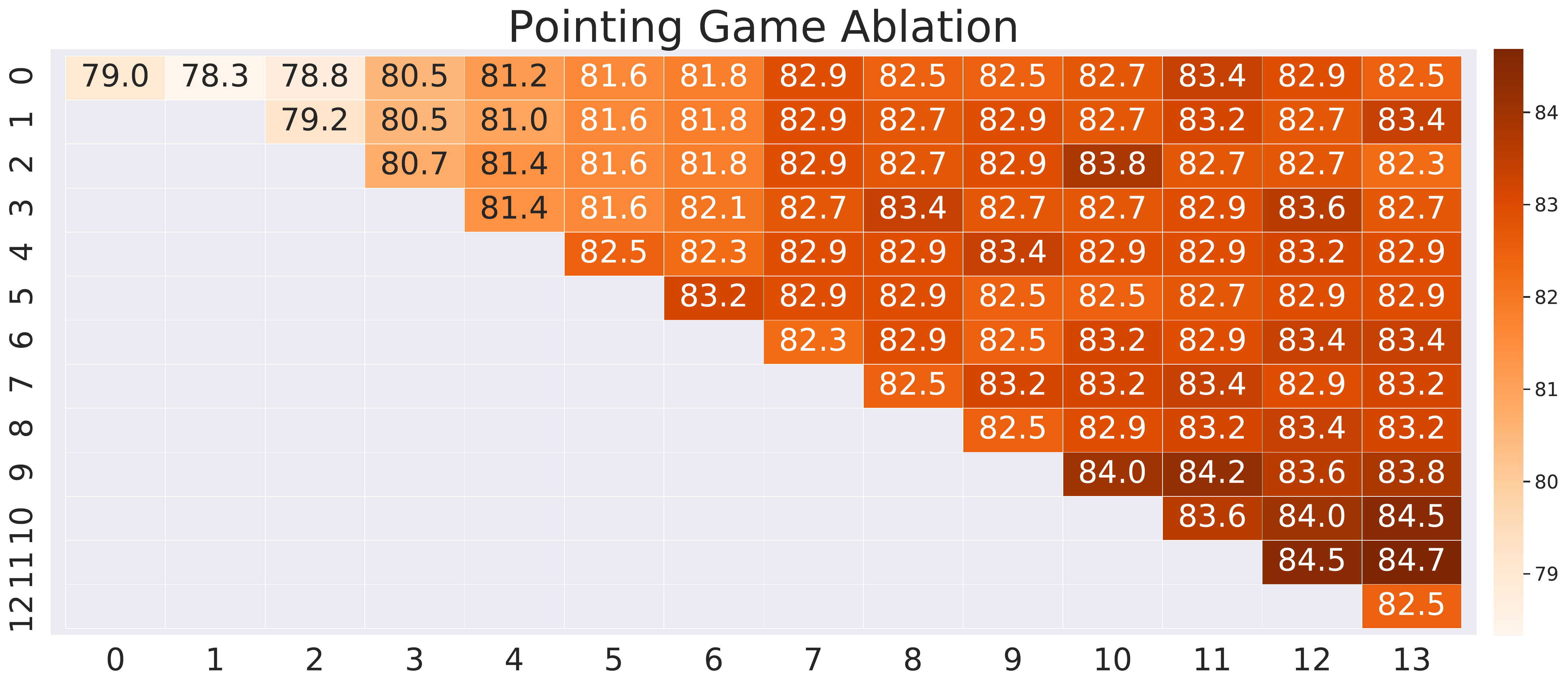}
\includegraphics[width=0.495\linewidth]{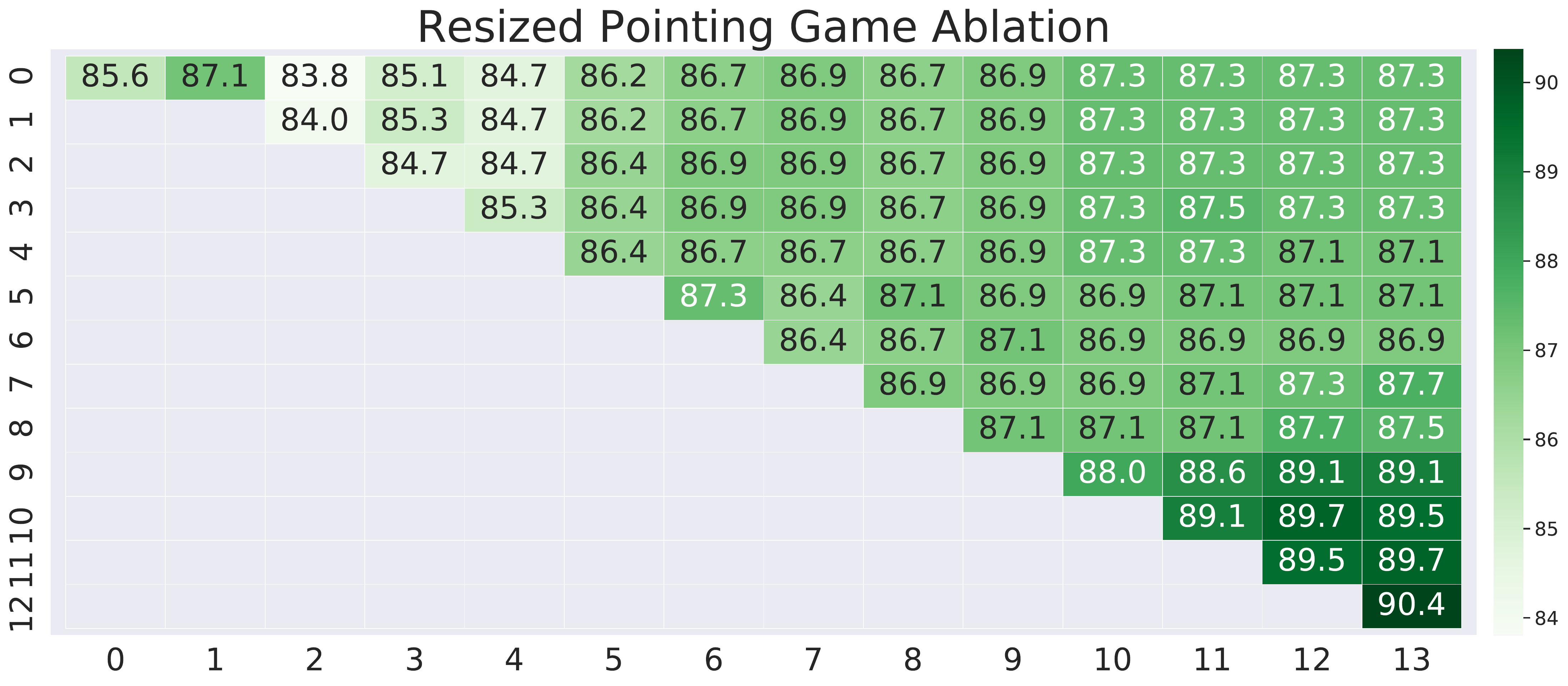}
\\
\includegraphics[width=0.495\linewidth]{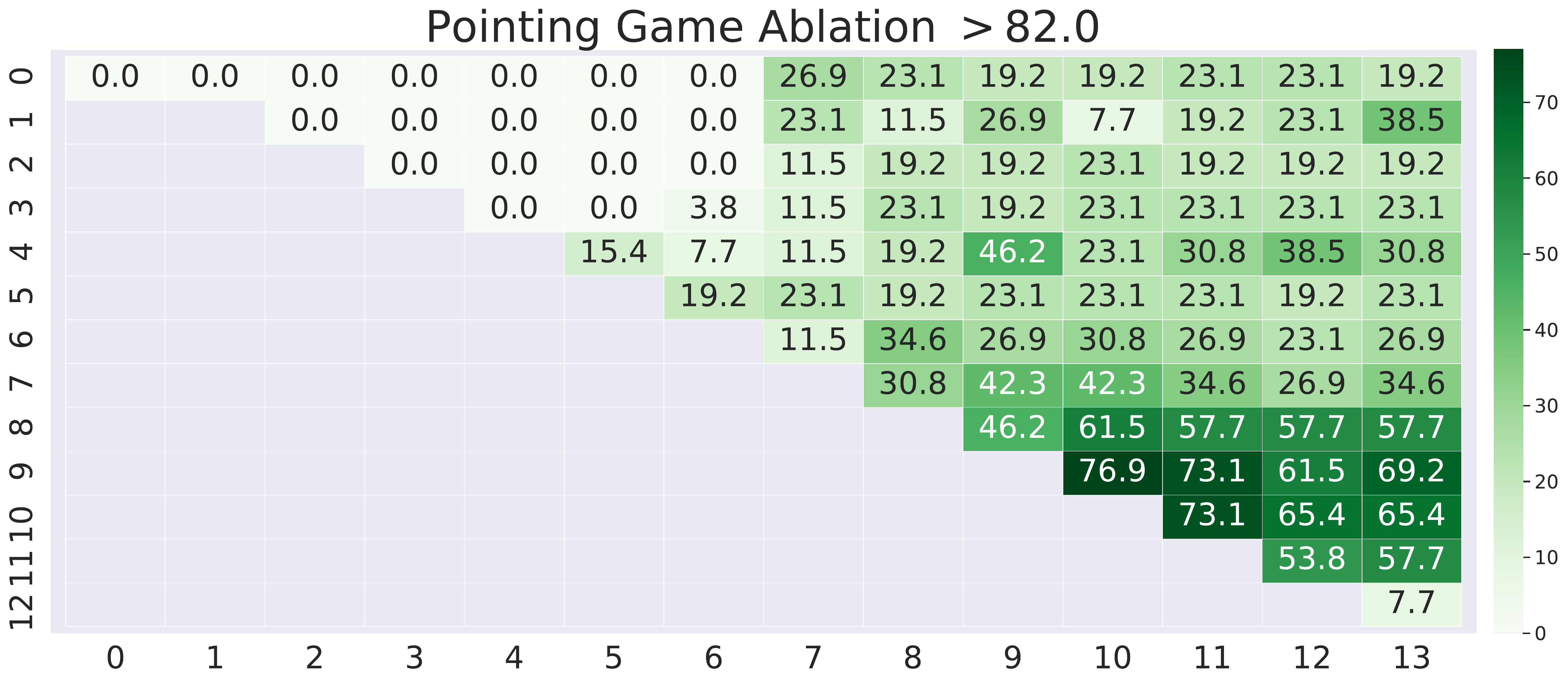}
\includegraphics[width=0.495\linewidth]{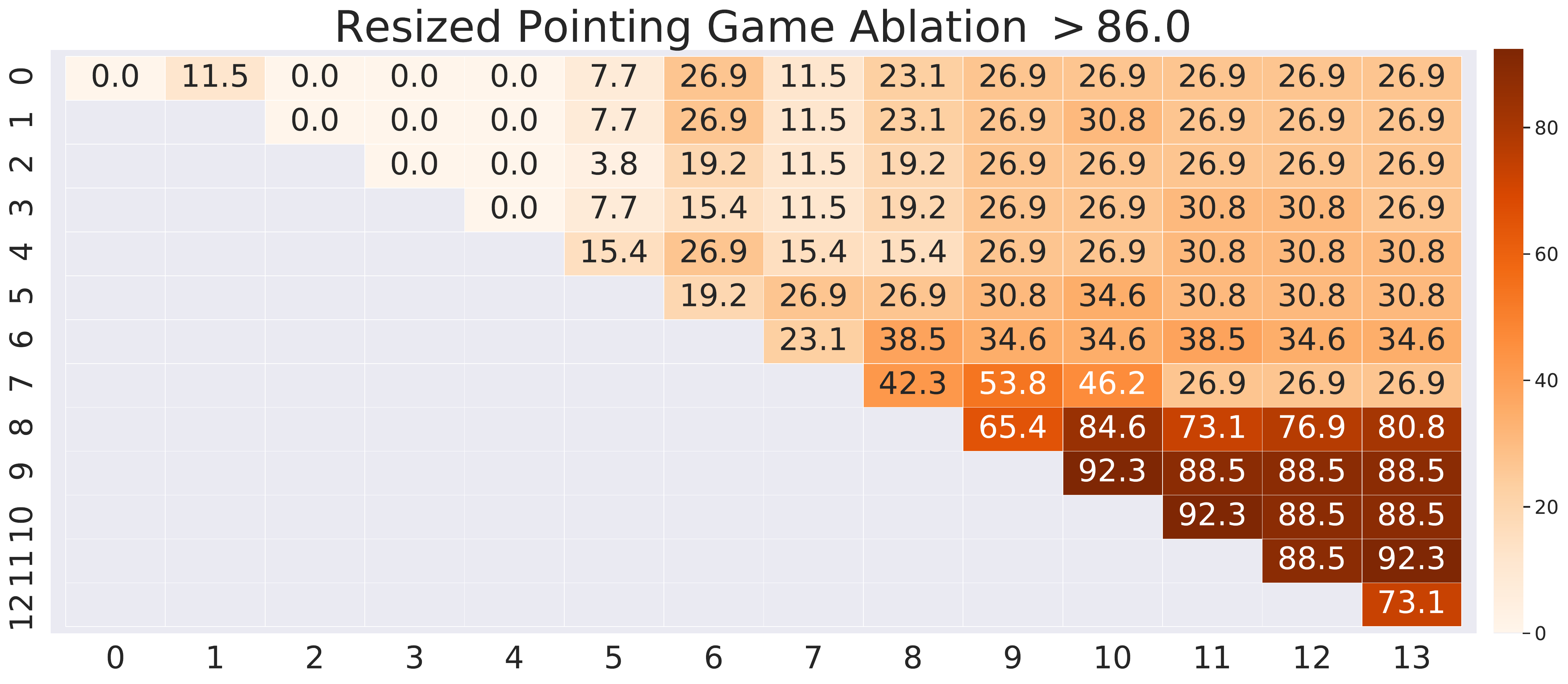}
\caption{Sensitivity study on the choice of layers regularized for the pointing game with limited $\sigma$. To test how robust each result we display the number of $\sigma$s which have performance above $82\%$ for the regular image size (left) and $86\%$ for the resized image (right). 
\label{fig:heatmapPointingGameSmallSigma}
}
\end{figure}

In this supplementary, we restrict the maximum value of $\sigma$ to $25$, which is in the same range as the value used in Extremal Perturbation ($\approx 20$) and rerun the experiment. 
The ablation shown in Fig.~\ref{fig:heatmapPointingGameSmallSigma}, shows similar results, with a slight decrease in performance ($0.4\%$ in the standard game), which is to be expected as we are maximizing over a smaller number of values of $\sigma$. 

We select the set of layers and $\sigma \in \{0,\dots 25\}$ which 
maximizes the performance on the ablation set, which for the regular 
set is ReLU 11 and 12 with $\sigma=21$, and for the resized set it is ReLU 12 with $\sigma=23.0$. We also compare against the no perceptual baseline with the same restricted $\sigma$, in both cases a $\sigma$ of $25$ is optimal. 

\begin{table}[!h]
\begin{center}
\begin{tabular}{|c|cc|}
\hline
Method & Orig. Image & Scaled Image \\ \hline \hline
Us NoPer                                 &  78.8 (58.9)               &  85.3 (70.5)  \\ 
Us                                       &  84.8 (70.5)               &  87.7 (74.5)  \\ \hline 
\end{tabular}
\end{center}
\caption{Results for the reduced $\sigma$ in the pointing game. This table
differs from the version in the main text, as for this experiment 
we further limit $\sigma \in \{0,\dots 25\}$. 
We present two sets of results, the
performance on the images in the original size (first column),
and the performance on images which are scaled to 1.5x using bilinear
scaling (second column). For each
result, the first number shows the percentage on the standard set,
with the number in brackets the performance on the difficult set. 
\label{tab:pointingGameResReducedSigma}}
\end{table}

The results can be seen in Fig.~\ref{fig:heatmapPointingGameSmallSigma} and Table~\ref{tab:pointingGameResReducedSigma}. 
The results are broadly comparable with the 
larger $\sigma$ study.
The main change is that the range of layers that 
obtain over $82\%$ 
(and over $86\%$ in the resized study)
is reduced, although the range of layers that perform well is maintained. 
For regular sized images there is a $0.3\%$ drop in the standard study and an improvement in the difficult study. The performance in the resized study is comparable with a $0.5\%$ drop in the resized study and a $2.0\%$ drop in the resized difficult study. The no perceptual baseline results have a more substantial drop with the original image size, potentially indicating that the blur is more important for the no perceptual method, in the 
standard set with a drop of 
$2.4\%$ and a drop of $4\%$ in the difficult set. 
Whereas in the resized set, there is a drop of $0.7\%$ and $1.1\%$. 
In the limited $\sigma$ the perceptual outperforms no perceptual in all cases,  
further, even comparing the limited $\sigma$ perceptual results with the general $\sigma$ no perceptual results, the perceptual results still outperforms in all cases. 
Furthermore, regardless of the $\sigma$ set, the order and the qualitative performance of the alternative methods in the standard set is maintained.

\end{document}